\documentclass{article}
\usepackage{spconf,amsmath,graphicx}
\usepackage{cite}
\usepackage{amsmath,amssymb,amsfonts}
\usepackage{algorithmic}
\usepackage{algorithm}
\usepackage{graphicx}
\usepackage{textcomp}
\usepackage{bm} 
\usepackage{xcolor}
\usepackage{subfig}
\usepackage{booktabs}
\usepackage{enumitem}
\usepackage{multirow}
\usepackage{tikz}
\usepackage{amsthm}

\theoremstyle{remark}

\usepackage{enumitem}
\def\BibTeX{{\rm B\kern-.05em{\sc i\kern-.025em b}\kern-.08em
    T\kern-.1667em\lower.7ex\hbox{E}\kern-.125emX}}
\usepackage{balance}

\makeatletter

\title{WREN: Low Light Image Enhancement Using Retinex theory-based \\Double U-Net-like Structures}
\name{
Reina Kaneko$^{1}$, Junya Hara$^{1}$, Hiroshi Higashi$^{2}$, Yuichi Tanaka$^{1}$
\thanks{This work is supported in part by JSPS KAKENHI under Grants 26H02536 and JST AdCORP under Grant JPMJKB2307.}
}
\address{
$^{1}$ The University of Osaka, Osaka, Japan,
$^{2}$ Kansai University, Osaka, Japan
}

\begin{document}
\ninept
\maketitle
\begin{abstract}
This paper proposes a neural network for low light image enhancement (LLIE) based on retinex theory to make LLIE robust for various dynamic range scenes.
The retinex theory is an image formulation model inspired by a human color perception hypothesis, where a low light image is decomposed into intrinsic color context (i.e., reflectance map) and scene-dependent illumination (i.e., illumination map). 
Due to non-uniqueness of its decomposition, existing retinex-based LLIE methods often fail to achieve stable decomposition, which
lead to over-enhancement.
Typically, they are sensitive to the dynamic ranges that vary in different lighting conditions.
To tackle this issue, we propose \textit{WREN}: An LLIE neural network with double U-Net-like structures.
WREN consists of two U-Net-like sub-networks.
The first network has one encoder and two decoders that decompose an input image into the reflectance and illumination maps.
The second network with a customized Transformer block between an encoder and a decoder \textit{only} enhances the illumination map obtained from the first network: This completely follows the assumption of the retinex theory.
Finally, the enhanced illumination map is recombined with the reflectance map.
The network is trained end-to-end with a scale-invariant loss function, which gives robustness against the illumination scaling.
Numerical results show that our method achieves the state‑of‑the‑art performance across multiple datasets. Our code is available online\footnote{https://github.com/reina0112/WREN}.
\end{abstract}
\section{Introduction}
\label{sec:intro}

Traffic accidents, disasters, and street crimes often occur in dark environments. 
In such situations, we need to monitor the situation accurately and/or recognize objects from low light images and videos. 
Low light images primarily suffer from illumination loss.
They are also degraded by sensor noise and SNR imbalances across color channels arising from differences in spectral sensitivity.
Hence, image processing and computer vision under low light conditions are crucial for night vision, search-and-rescue, and surveillance applications \cite{Chen2018Retinex}:
Low light image enhancement (LLIE) is desirable prior to any downstream tasks \cite{guoLIMELowLightImage2017c,Chen2018Retinex,xuSNRAwareLowlightImage2022,caiRetinexformerOnestageRetinexbased2023,guoZeroReferenceDeepCurve2020a,zhangBrighteningLowlightImages2021a,lanEfficientDiffusionLow2025,oginoCurveClipUtilizedReinforcement2025,yiLlieformerLowLightImage2023}.

Existing LLIE methods typically have the following two stages \cite{Chen2018Retinex,zhangBrighteningLowlightImages2021a}. 
1) Illumination compensation stage: A low light image is first decomposed into the reflectance and illumination maps. Then, the illumination map is amplified and synthesized back with the reflectance map.
2) Refinement stage: The remaining artifacts and noise are removed from the compensated normal light image. 

\begin{figure}[t]
    \centering
    \includegraphics[width=1.0\linewidth]{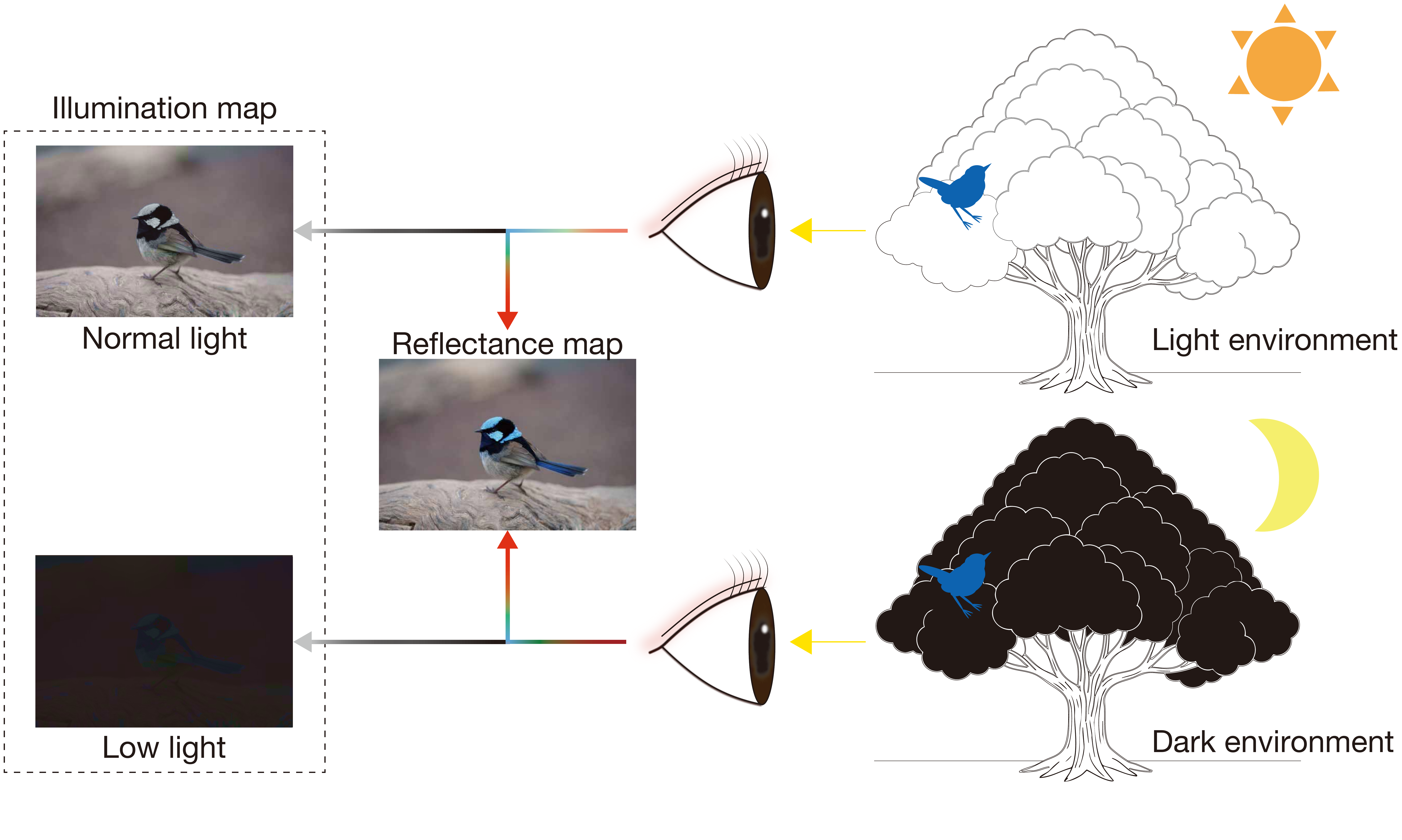}
    \caption{Illustration of the retinex theory. It assumes normal and low light images have the identical reflectance map and have different illumination maps.}
    \label{fig:retinex}
    \end{figure}

In the two-stage approach, \textit{retinex theory} \cite{landLightnessRetinexTheory1971} plays a crucial role.
It is a hypothesis in human color perception where
the reflectance map is assumed to be invariant to light conditions, while the illumination map is faded by dark environments, as illustrated in Fig.~\ref{fig:retinex}.

Most state-of-the-art LLIE methods partially employ assumptions based on the retinex theory for designing their building blocks \cite{guoLIMELowLightImage2017c,Chen2018Retinex,caiRetinexformerOnestageRetinexbased2023,zhangBrighteningLowlightImages2021a}.
In spite of such efforts, their resulting images often have over-smoothing artifacts that may limit performances of downstream tasks.
This is due to the refinement stage.
As previously mentioned, the refinement is performed on the synthesized normal light image which is the element-wise product of the reflectance and illumination maps.
The refinement stage, therefore, implicitly assumes that the reflectance map contains noise as well as the illumination map.
However, in principle of the retinex theory,
the reflectance map should be noise-free even when the illumination map is noisy \cite{landLightnessRetinexTheory1971}.
As a result, the smoothing in the refinement stage causes over-smoothing due to the simultaneous smoothing of the reflectance and illumination maps.
\begin{figure*}[t]
    \centering
    \includegraphics[width=1.0\linewidth]{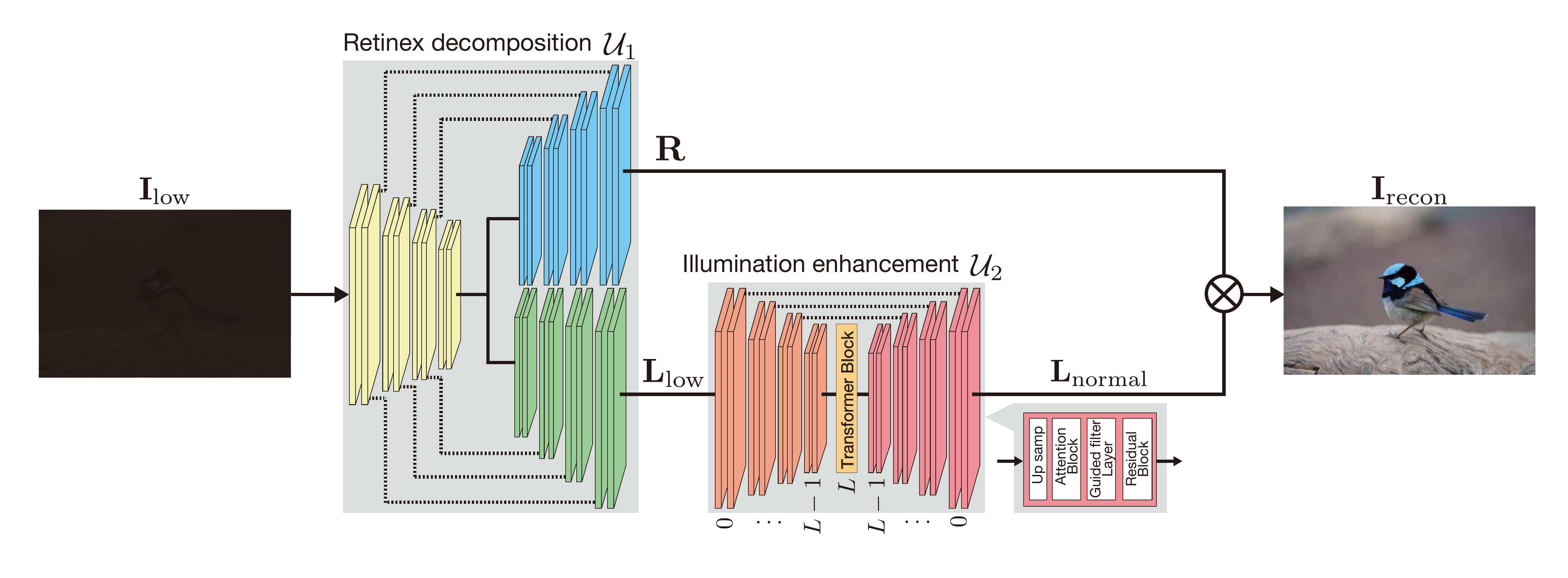}
    \caption{Illustrative overview of WREN.}
    \label{fig:network}
    \end{figure*}

In this paper, we propose \textit{WREN}, a retinex theory-guided LLIE method to overcome the over-smoothing issue in LLIE.
WREN contains two U-Net-like sub-networks.
In the first U-Net-like network, an input low light image is decomposed into reflectance and illumination maps with one encoder and two decoders.
Based on the principle of the retinex theory, the decomposed reflectance map remains unchanged in contrast to the other approaches.
The illumination map is further enhanced by the second U-Net-like network.
It has
a Transformer block \cite{vaswaniAttentionAllYou2017} between the encoder and the decoder to utilize non-local smoothness for denoising the illumination map.
We also integrate attention blocks \cite{wooCBAMConvolutionalBlock2018} and trainable guided filter blocks \cite{wuFastEndtoEndTrainable2018} in the decoder of the second sub-network for strongly denoising the low light region of the illumination map.
WREN is trained end-to-end.

In our experiment, we compare the proposed method with alternative methods across multiple LLIE datasets to evaluate it in several different illumination conditions.
Numerical results demonstrate that WREN achieves the state-of-the-art performance on both distortion- and structure-based metrics.

\section{Related Work}
\label{sec:relatedwork}
    
In this section, we review existing LLIE techniques.
First, we introduce the retinex theory. Then, 
representative model- and deep learning-based LLIE approaches are presented.

\subsection{Retinex Theory}\label{subsec:relatedretinex}
Retinex theory is an image formation model in the human visual system \cite{landLightnessRetinexTheory1971} and is widely used in LLIE. 
Let $\mathbf{I}_{\text{retina}}\in[0,1]^{H\times W\times C}$ denote a digital image corresponding to that perceived in the human retina, where $H$, $W$, and $C$ are the height, width, and number of channels, respectively.
The retinex theory assumes
\begin{equation}\label{eq:retinex}
    \mathbf{I}_{\text{retina}} = \mathbf{R}\odot\mathbf{L},
\end{equation}
where $\mathbf{R}\in[0,1]^{H\times W\times C}$ is a reflectance map that represents the intrinsic surface reflectance, $\mathbf{L}\in[0,1]^{H\times W \times C}$ is an illumination map that represents the incident light intensity, and $\odot$ is the element-wise multiplication. Note that $\mathbf{L}$ does not depend on color channels, i.e., the same matrix is replicated $C$ times across color channels. 

Note that $\mathbf{R}$ represents the \textit{true color} of the image while $\mathbf{L}$ is changed due to various illumination conditions and noises.
Therefore, $\mathbf{R}$ must be noise-free in principle of the retinex theory.

Since the decomposition of the reflectance and illumination maps is non-unique and typically ill-posed, many attempts for the retinex decomposition have been made in LLIE \cite{landLightnessRetinexTheory1971}.

\subsection{Model-based LLIE}

A seminal LLIE method, LIME \cite{guoLIMELowLightImage2017c}, estimates the illumination map via convex optimization with a piecewise smoothness prior, and the reflectance map is immediately obtained by element-wise division.
Then, it applies gamma correction \cite{rahmanAdaptiveGammaCorrection2016a} to the illumination map and synthesizes it back to the reflectance map.
Many variants of LIME have also been proposed \cite{fuWeightedVariationalModel2016,liStructureRevealingLowLightImage2018}.

When they have a mismatch of priors, which are included in optimization functions, during the estimation of the illumination map, the reflectance map is also affected due to the element-wise division.
Therefore, they often perform smoothing of the compensated normal light image: This results in over-smoothing.

\subsection{Deep learning-based LLIE}

Deep learning techniques can flexibly address the above non-unique decomposition issue as well as degradation factors as long as enough training data are available.

Many deep learning-based LLIE methods are based on the retinex decomposition in \eqref{eq:retinex} similar to the model-based approaches. For example, a convolutional neural network (CNN) proposed in \cite{Chen2018Retinex} learns the retinex decomposition by enforcing consistency of the reflectance map between paired low and normal light images. However, many deep learning-based methods design the illumination compensation and refinement stages separately because joint training across these two stages is often unstable \cite{Chen2018Retinex}.

To address the issue, Retinexformer \cite{caiRetinexformerOnestageRetinexbased2023} modifies \eqref{eq:retinex} so that it accommodates sensor noise.
It learns the illumination compensation and image refinement in an end-to-end framework. Similar to the framework of LIME, it first estimates the illumination map with a CNN. Then, low light images are enhanced by a Transformer whose attention is guided by the estimated illumination map.
However, it could be sensitive to dynamic ranges of the training data \cite{caiLearningDeepSingle2018a,zhaiPerceptualQualityAssessment2021,zhangNoReferenceEvaluationMetric2021,Liu2024NTIRE2024LLIE,bychkovskyLearningPhotographicGlobal2011}.
If lighting conditions are changed from the training data, it yields over- or under-enhanced images.
\vspace{0.2cm}

In summary, there exist three challenges in the current LLIE methods: 1) non-trivial retinex decomposition, 2) instability in end-to-end learning, and 3) sensitivity to the illumination conditions.

\section{WREN Structure}
\label{sec:proposed}
In this section, we present the proposed LLIE method, WREN.
The overview of the structure of WREN is illustrated in Fig.~\ref{fig:network}.
As seen in the figure, WREN consists of two sub-networks, both based on U-Net.
One is a retinex decomposition network $\mathcal{U}_1$ to decompose the reflectance and illumination maps from the low light image. The other is an illumination restoration network $\mathcal{U}_2$ for the decomposed illumination map.

Note that WREN intentionally keeps the relrectance map yielded by $\mathcal{U}_1$ as-is based on the principle of the retinex theory described in Section~\ref{subsec:relatedretinex}.
While WREN seems to have a surprisingly simple structure compared to the other deep learning-based LLIE methods, its enhancement power outperforms the alternative ones due to the structure strictly following the retinex assumption.
\subsection{Network Structure}
Here, we introduce the building blocks of WREN.

\subsubsection{Retinex Decomposition Network}\label{subsec:retinex_decomp_net}
For the retinex decomposition, an attention U-Net $\mathcal{U}_1$ having one shared encoder and two separated decoders is designed. The encoder have five $3\times 3$ convolution layers, downsampling layers, and ReLU activation modules \cite{nairRectifiedLinearUnits2010}.
Let $\mathbf{I}_{\text{low}}\in[0,1]^{H\times W \times C}$ be an input low light image. The two decoders respectively output the reflectance map $\mathbf{R}\in [0,1]^{H\times W\times C}$ and the illumination map $\mathbf{L}_{\text{low}}\in[0,1]^{H\times W}$ from $\mathbf{I}_{\text{low}}$ as 
\begin{equation}\label{eq:u_fork}
    (\mathbf{R},\mathbf{L}_{\text{low}})=\mathcal{U}_{1}(\mathbf{I}_\text{low}).
\end{equation}

As previously mentioned, we remain the reflectance map $\mathbf{R}$ unchanged.
The illumination map $\mathbf{L}_{\text{low}}$ is only performed the refinement by $\mathcal{U}_2$ in the second step.

\subsubsection{Illumination Compensation Network}

For the illumination compensation, we also employ a U-Net-type network. 
Let us denote the standard encoder of U-Net by $\mathcal{U}_{\text{2},\text{enc}}$ and a customized U-Net decoder $\widehat{\mathcal{U}}_{\text{2},\text{dec}}$ (its structure is shown below).
We enhance the illumination map $\mathbf{L}_{\text{low}}$ in \eqref{eq:u_fork} as follows:
\begin{equation}
        \mathbf{L}_{\text{normal}} = \widehat{\mathcal{U}}_{\text{2},\text{dec}}(\{\mathbf{F}_{\ell}\}_{\ell=0}^L) = \widehat{\mathcal{U}}_{\text{2},\text{dec}}(\mathcal{U}_{\text{2},\text{enc}}(\mathbf{L}_{\text{low}}))
        ,\label{eq:illumi_enhance_net}
\end{equation}
where
$\mathbf{F}_\ell$ is encoded features by $\mathcal{U}_{\text{2},\text{enc}}$ at the $\ell$th scale.
$\widehat{\mathcal{U}}_{\text{2},\text{dec}}$ is recursively processed as follows:
\begin{equation}
    \mathbf{L}_{\text{normal}} = \mathcal{O}(\mathbf{G}_0),\label{eq:OG_0}
\end{equation}
where $\mathcal{O}$ applies a channel-wise convolution followed by a sigmoid activation and
\begin{equation}
        \mathbf{G}_\ell = \begin{cases}
        \mathcal{T}(\mathbf{F}_L) & \ell = L,\\
            \Phi_\ell(\mathbf{G}_{\ell+1},\mathbf{F}_\ell) & \ell=L-1,\dots,0,
        \end{cases}\label{eq:Gell}
\end{equation}
in which $\mathcal{T}$ is a Transformer block, i.e., a submodule in Transformer containing multi-head attention, MLP, and layer normalization \cite{vaswaniAttentionAllYou2017}.
The building blocks in \eqref{eq:Gell} are, in particular, given by
\begin{align}
         \mathcal{T}(\mathbf{F}_L) =&\mathbf{F}_L
+\mathrm{MHSA}\!\bigl(\mathrm{LN}(\mathbf{F}_L)\bigr)\\\nonumber
&+\mathrm{MLP}\!\left(\mathrm{LN}\!\left(\mathbf{F}_L
+\mathrm{MHSA}\!\bigl(\mathrm{LN}(\mathbf{F}_L)\bigr)\right)\right),\\
     \Phi_\ell(\mathbf{G}_{\ell+1},\mathbf{F}_\ell) =& \operatorname{Res}_\ell(\operatorname{GF}_\ell(\operatorname{Attn}_\ell (\operatorname{Up}_\ell(\mathbf{G}_{\ell+1},\mathbf{F}_\ell)))).
\end{align}
Their components are listed as follows:
\begin{itemize}
    \item $\mathrm{MHSA}$: Multi-head self attention.
    \item $\mathrm{MLP}$: Multilayer perceptron.
    \item $\mathrm{LN}$: Layer normalization.
    \item $\operatorname{Up}_\ell$: Upsampling block.
    \item $\operatorname{Attn}_\ell$: Attention block \cite{wooCBAMConvolutionalBlock2018}.
    \item $\operatorname{GF}_\ell$: Trainable guided filter that uses encoder features $\{\mathbf{F}_\ell\}$ as its guide \cite{wuFastEndtoEndTrainable2018}.
    \item $\operatorname{Res}_\ell$: Residual block.
\end{itemize}
Note that $\mathcal{T}$ is inserted between the encoder and decoder to capture non-local features in the encoded illumination maps thanks to self-attention.
This promotes non-local illumination smoothness.

\begin{table}[t]
    \centering
        \caption{Training configuration.}
    \begin{tabular}{c|c}\hline
       Batch size  &  4\\
       Epochs  & 400 \\
       Optimizer & AdamW \\
       Learning rate & $3\times 10^{-4}$\\
       Scheduler & OneCycleLR \\\hline
    \end{tabular}
    \label{tab:config}
\end{table}

\begin{table*}[t]
\centering
\caption{Comparison of numerical results of LLIE. Average values of all test data are shown. Boldface and underlined numbers denote the best and second-best methods, respectively.}
\begin{tabular}{c||ccccccc|c}
\hline
                           \multirow{2}{*}{Metric}    & \multicolumn{7}{c|}{Deep learning-based methods} & Model-based method\\ \cline{2-9}
                                & \textbf{WREN} & ReDDiT & Retinexformer & RetinexNet & SNR-Net & KinD++ & Zero-DCE & LIME \\ \hline\hline

                        PSNR $\uparrow$   & \textbf{18.222} & \underline{17.315} & 17.022 & 15.105 & 16.196 & 16.076 & 14.936 & 14.521 \\ \cline{2-9}
                     SSIM $\uparrow$   & \textbf{0.814}  & 0.459 & 0.742 & \underline{0.790} & 0.709 & 0.778 & 0.708 & 0.581 \\ \cline{2-9}
                        LPIPS $\downarrow$& \textbf{0.302}  & 0.425 & 0.349 & 0.411 & 0.415 & \underline{0.329} & 0.351 & 0.447 \\ \cline{2-9}
                        MAE $\downarrow$  & \textbf{0.118}  & \underline{0.127} & 0.146 & 0.152 & 0.168 & 0.142 & 0.174 & 0.171 \\ \hline
\end{tabular}
\label{tab:results}
\end{table*}

The customized decoder $\widehat{\mathcal{U}}_{\text{2},\text{dec}}$ consists of a series of $\{\Phi_\ell\}_{\ell=0}^{L-1}$ and a final output layer $\mathcal{O}$.
They are used to preserve edges and textures in the illumination map.

\subsection{Loss Function}
WREN is trained with the following loss functions.

\subsubsection{Loss Function Corresponding to Retinex Decomposition Network}
For $\mathcal{U}_1$, we use the following loss function for\eqref{eq:u_fork}.
\begin{equation}
    \mathcal{L}_{\text{decomp}} = \alpha \mathcal{L}_{\text{reconst}} + \beta\mathcal{L}_{\text{SI}}
\label{eq:decoomploss}
\end{equation}
where $\mathcal{L}_{\text{reconst}}(\mathbf{I}_{\text{low}}, \mathbf{I}_{\text{recon}})$ is the fidelity term and
$\mathcal{L}_{\text{SI}}(\mathbf{I}_{\text{normal}}, \mathbf{I}_{\text{recon}})$ is the scale invariant loss that provides the robustness to the lighting condition change, with hyperparameters $\alpha$ and $\beta$.
$\mathcal{L}_{\text{reconst}}(\mathbf{I}_{\text{low}}, \mathbf{I}_{\text{recon}})$ and $\mathcal{L}_{\text{SI}}(\mathbf{I}_{\text{normal}}, \mathbf{I}_{\text{recon}})$ are respectively defined as follows:
\begin{align}
    \mathcal{L}_{\text{reconst}} & = \|\mathbf{I}_{\text{low}} - \mathbf{I}_{\text{recon}}\|_F^2,\label{eq:reconloss}\\
    \mathcal{L}_{\text{SI}} &= \operatorname{var}(\log(\mathbf{I}_{\text{normal}}) - \log(\mathbf{I}_{\text{recon}})),\label{eq:siloss}
\end{align}
in which $\mathbf{I_{\text{recon}}} = \mathbf{R}\odot\mathbf{L}_{\text{low}}$\footnote{Rigorously, $\textbf{L}_{\text{low}}$ is replicated across the number of channels.}, $\mathbf{I}_{\text{normal}}$ is the normal light image\footnote{While LLIE literature often uses $\cdot_{\text{high}}$ for normal light images, we use $\cdot_{\text{normal}}$ for consistency.}, and $\operatorname{var}$ represents variance.

Here, we present that \eqref{eq:siloss} is invariant to any global rescaling of $\mathbf{L}_{\text{normal}}$, i.e., dynamic range changes.
    Suppose that two normal light images sharing the same reflectance maps while having different illumination maps that only differ in their scalling: $\mathbf{I}_{\text{normal}1}=\mathbf{R}\odot\mathbf{L}_{\text{normal}}$ and $\mathbf{I}_{\text{normal}2}=\mathbf{R}\odot s\mathbf{L}_{\text{normal}}$, where $s>0$ is an arbitrary constant. Then, $\mathcal{L}_{\text{SI}}(\mathbf{I}_{\text{normal}2},\mathbf{I}_{\text{low}})$ is rewritten as follows:
    \begin{equation}\label{eq:si_loss_demo}
    \begin{split}
        \mathcal{L}_{\text{SI}}(\mathbf{I}_{\text{normal}2},\mathbf{I}_{\text{low}})&=\operatorname{var}(\log(s\mathbf{I}_{\text{normal}1})-\log(\mathbf{I}_{\text{low}}))\\
        &= \operatorname{var}(\log(\mathbf{I}_{\text{normal}1})-\log(\mathbf{I}_{\text{low}})+\log (s\mathbf{1}\mathbf{1}^\top)) \\
        &=\operatorname{var}(\log(\mathbf{I}_{\text{normal}1})-\log(\mathbf{I}_{\text{low}}))\\
        &=\mathcal{L}_{\text{SI}}(\mathbf{I}_{\text{normal}1},\mathbf{I}_{\text{low}}).
    \end{split}
    \end{equation}
    The third equality uses the invariance of the variance to an additive constant, i.e., $\operatorname{var}(\mathbf{X}+C\mathbf{1}\mathbf{1}^\top)=\operatorname{var}(\mathbf{X})$ for any constant $C$.
From \eqref{eq:si_loss_demo}, \eqref{eq:siloss} is clearly independent of the illumination scaling.

\subsubsection{Loss Function Corresponding to Illumination Compensation Network}
For $\mathcal{U}_2$, we use the following loss function for \eqref{eq:illumi_enhance_net}:
\begin{equation}
    \mathcal{L}_{\text{enh}} = \gamma\mathcal{L}_{\text{MSE}} + \zeta\mathcal{L}_{\text{MS-SSIM}} + \omega\mathcal{L}_{\text{perc}} + \delta \mathcal{L}_{\text{lab}} + \eta\mathcal{L}_{\text{bright}},
\label{eq:enhloss}
\end{equation}
where $\gamma$, $\zeta$, $\omega$, $\delta$, and $\eta$ are hyperparameters.
Its internal loss functions are represented as follows: 
\begin{align}
\mathcal{L}_{\text{MSE}} &= \|\mathbf{I}_{\text{normal}} - \mathbf{I}_{\text{recon}}\|_F^2,\\
\mathcal{L}_{\text{MS-SSIM}} &= 1 - \operatorname{MS-SSIM}(\mathbf{I}_{\text{normal}}, \mathbf{I}_{\text{recon}}),\\
\mathcal{L}_{\text{perc}} &= \|\psi(\mathbf{I}_{\text{normal}}) - \psi(\mathbf{I}_\text{recon})\|_1,\label{eq:percloss}\\
\mathcal{L}_{\text{lab}}&= \sum_{c\in\{a^\star,b^\star\}}\sum_{x,y} |\mathbf{I}_{\text{normal}}(c,x,y)-\mathbf{I}_{\text{recon}}(c,x,y)|,\\
\mathcal{L}_{\text{bright}} &= \|\max(0, \mathbf{L}_{\text{normal}}-1)\|_1.
\end{align}
With the standard MSE $\mathcal{L}_{\text{MSE}}$, we use the multi-scale structural similarity (MS-SSIM) \cite{wangMultiscaleStructuralSimilarity2003} loss $\mathcal{L}_{\text{MS-SSIM}}$, the perceptual loss $\mathcal{L}_{\text{perc}}$
\cite{johnsonPerceptualLossesRealTime2016} in which $\psi(\cdot)$ in \eqref{eq:percloss} denotes the high-level feature extractor (VGG-16 pre-trained on ImageNet) \cite{simonyanVeryDeepConvolutional2015}, the $\ell_1$ distance in CIE La$^\star$b$^\star$ space computed only for the $a^\star$ and $b^\star$ channels $\mathcal{L}_{\text{lab}}$, and the hinge penalty loss not to exceed the value above one $\mathcal{L}_{\text{bright}}$.

\subsubsection{Total Loss Function and Training}
We train \eqref{eq:u_fork} and \eqref{eq:illumi_enhance_net} end-to-end by simply combining \eqref{eq:decoomploss} and \eqref{eq:enhloss}.
\begin{equation}
    \mathcal{L} = \mathcal{L}_{\text{decomp}} + \mathcal{L}_{\text{enh}}.
\end{equation}
Finally, we obtain the normal light image \(\mathbf{I}_{\text{recon}}=\mathbf{R}\odot\mathbf{L}_{\text{normal}}\).

Our loss functions have two desirable features. First, it is robust to the dynamic range changes on training datasets thanks to \eqref{eq:si_loss_demo}.
Second, it consists of a simple pipeline derived from the retinex theory in \eqref{eq:retinex} by assuming the noiseless $\mathbf{R}$ that leads to a robust end-to-end training.
\begin{figure*}[tp]
    \centering
    \subfloat{\includegraphics[width = .10 \linewidth]{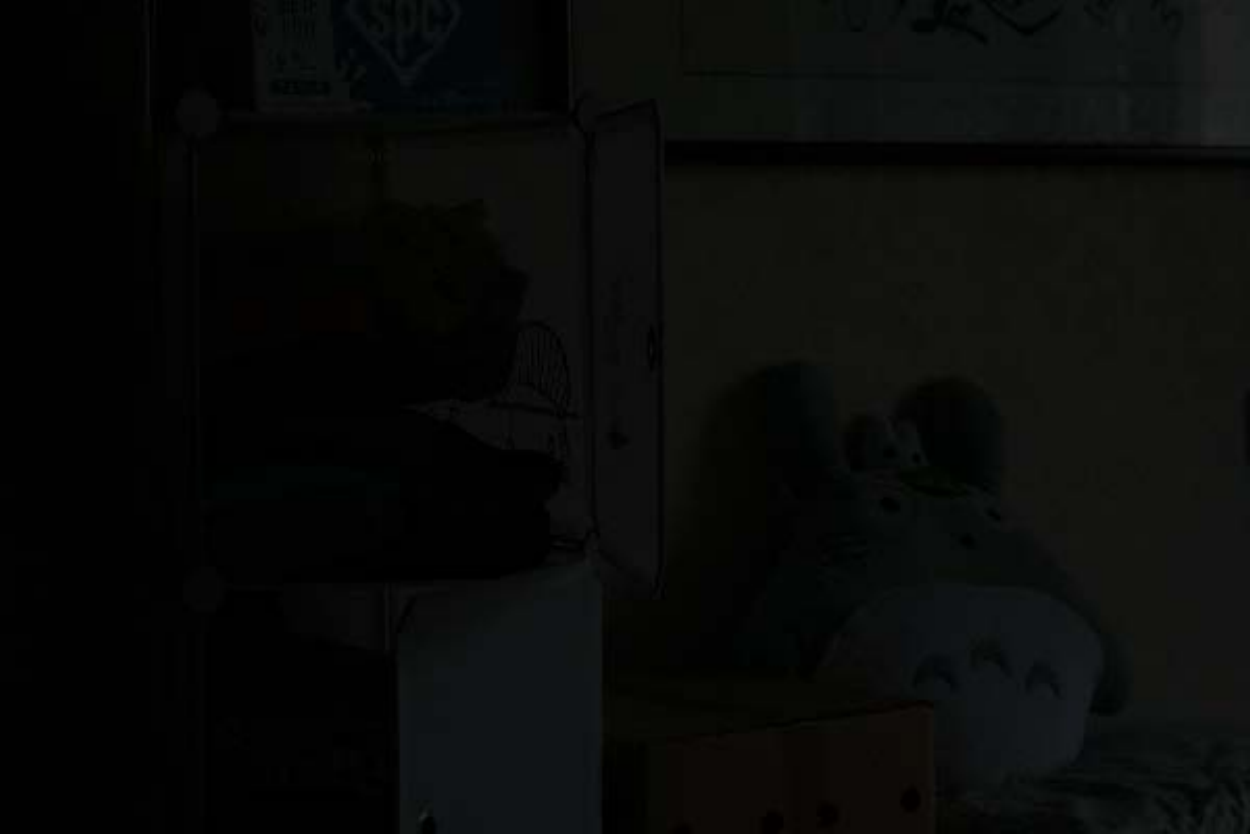}}
    \subfloat{\includegraphics[width = .10 \linewidth]{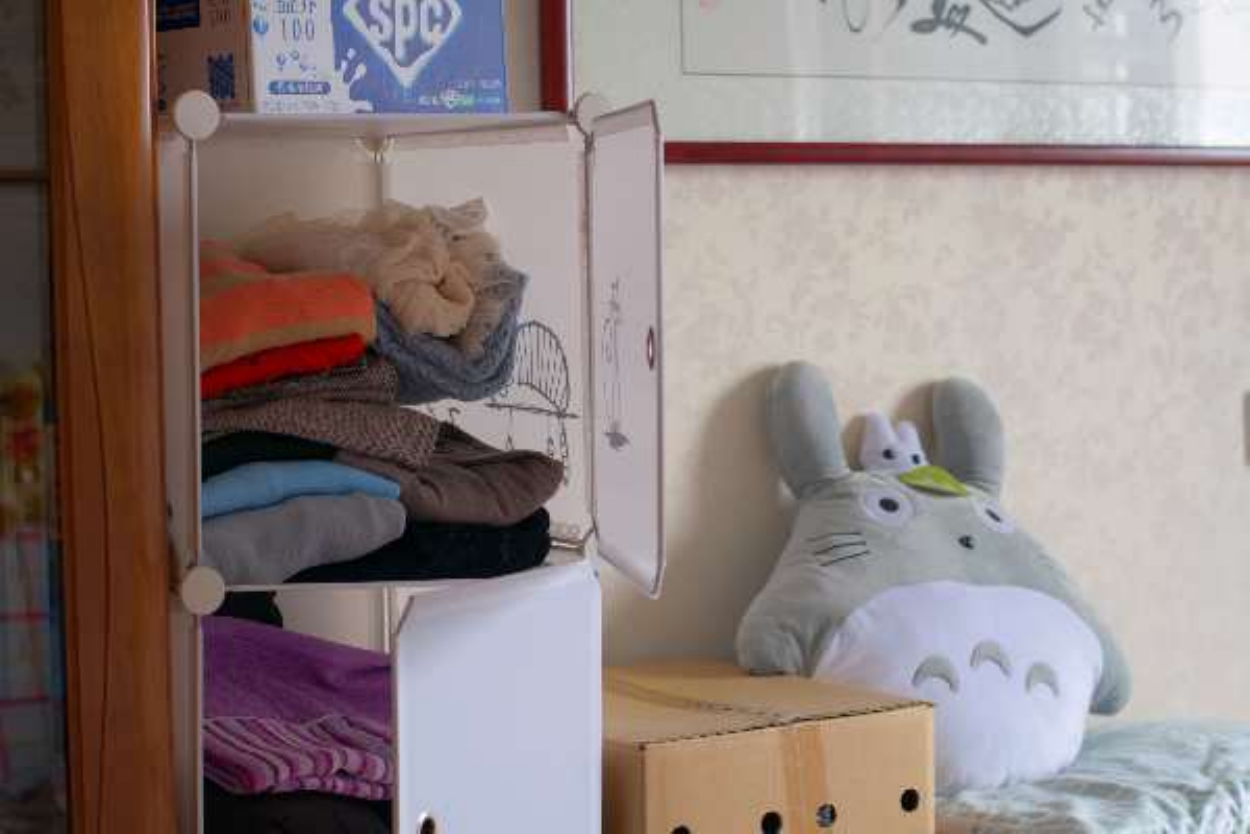}}
    \subfloat{\includegraphics[width = .10 \linewidth]{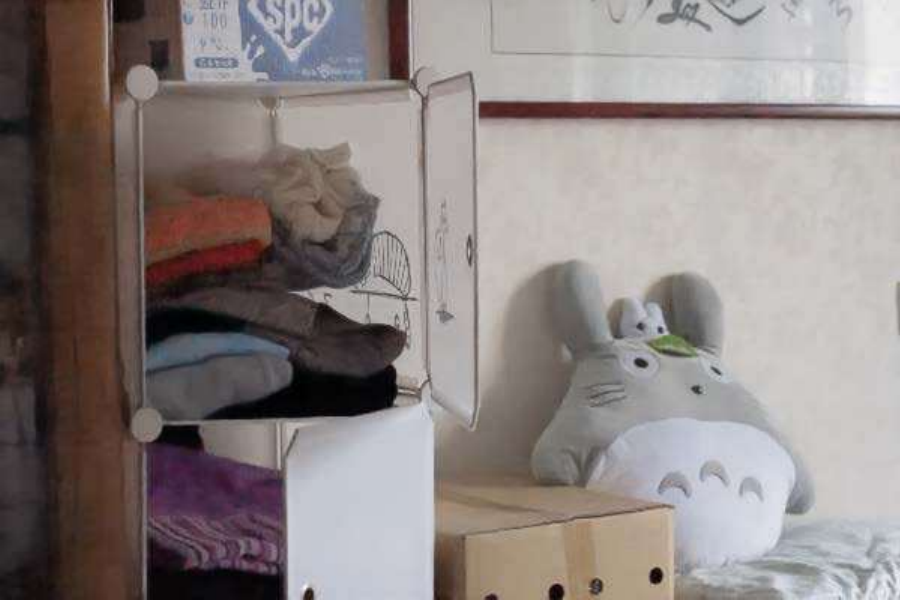}}
    \subfloat{\includegraphics[width = .10 \linewidth]{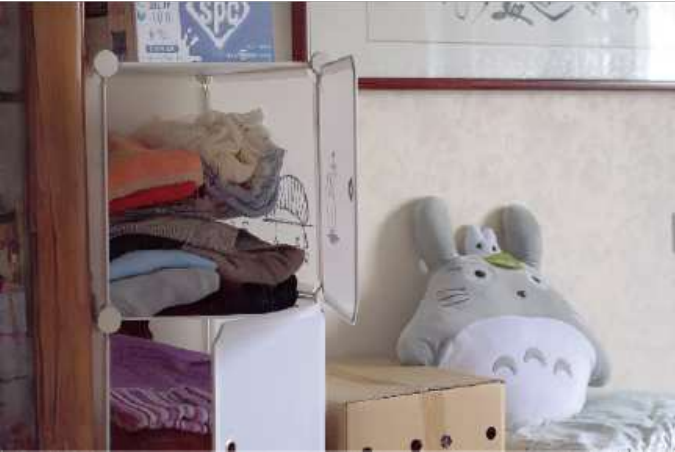}}
    \subfloat{\includegraphics[width = .10 \linewidth]{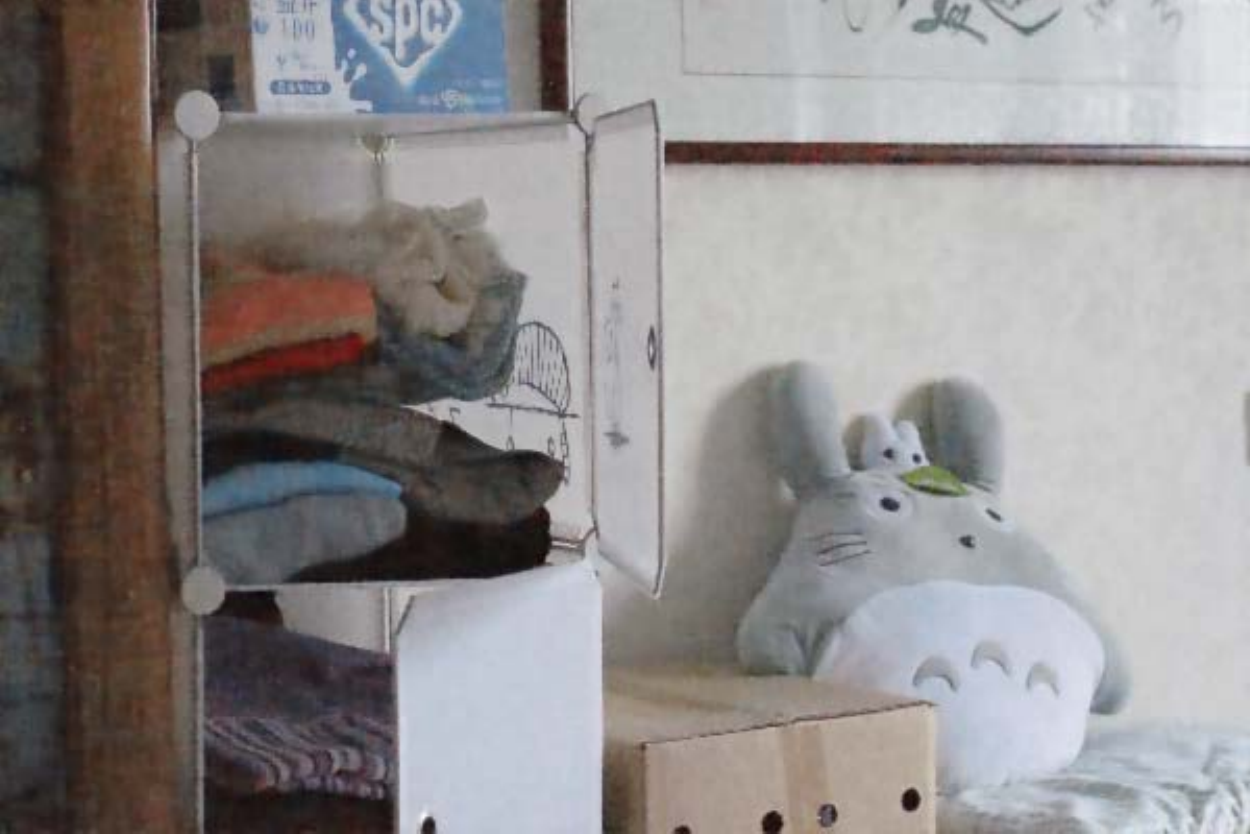}}
    \subfloat{\includegraphics[width = .10 \linewidth]{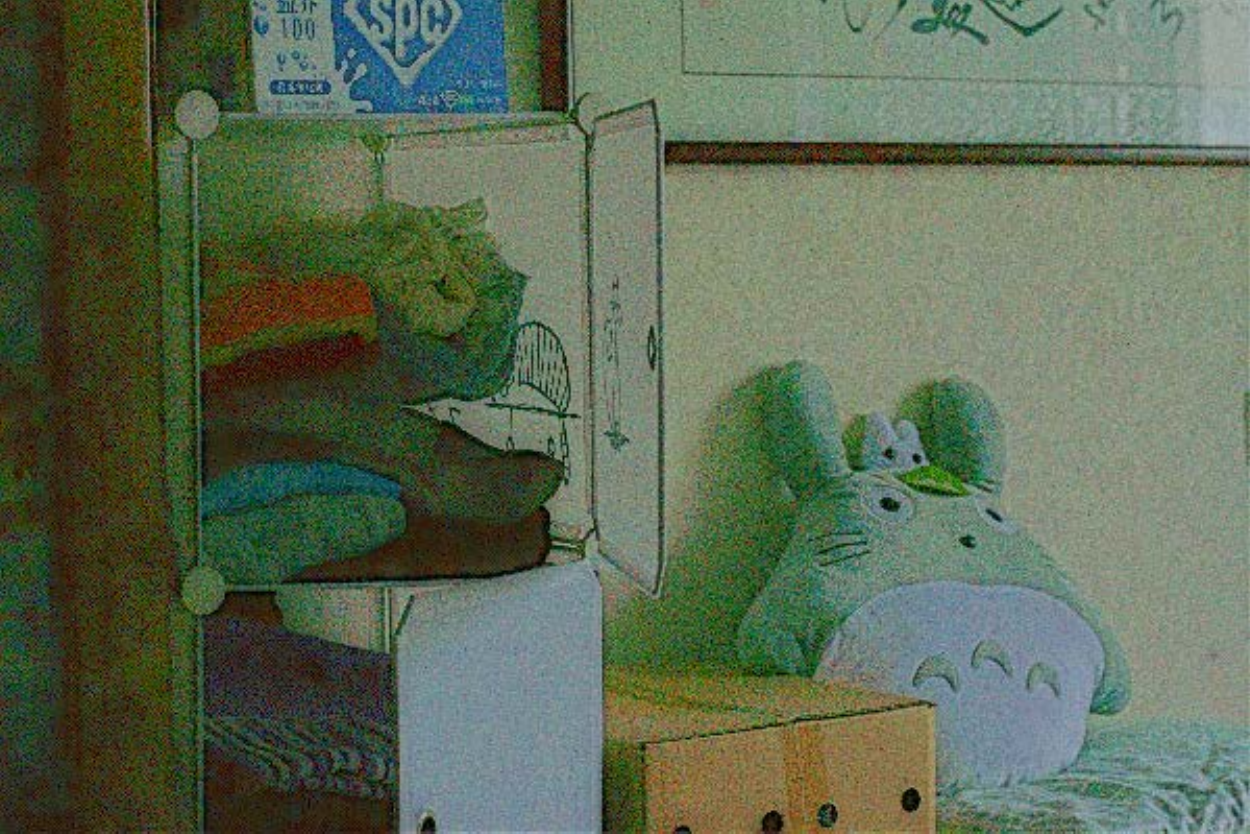}}
    \subfloat{\includegraphics[width = .10 \linewidth]{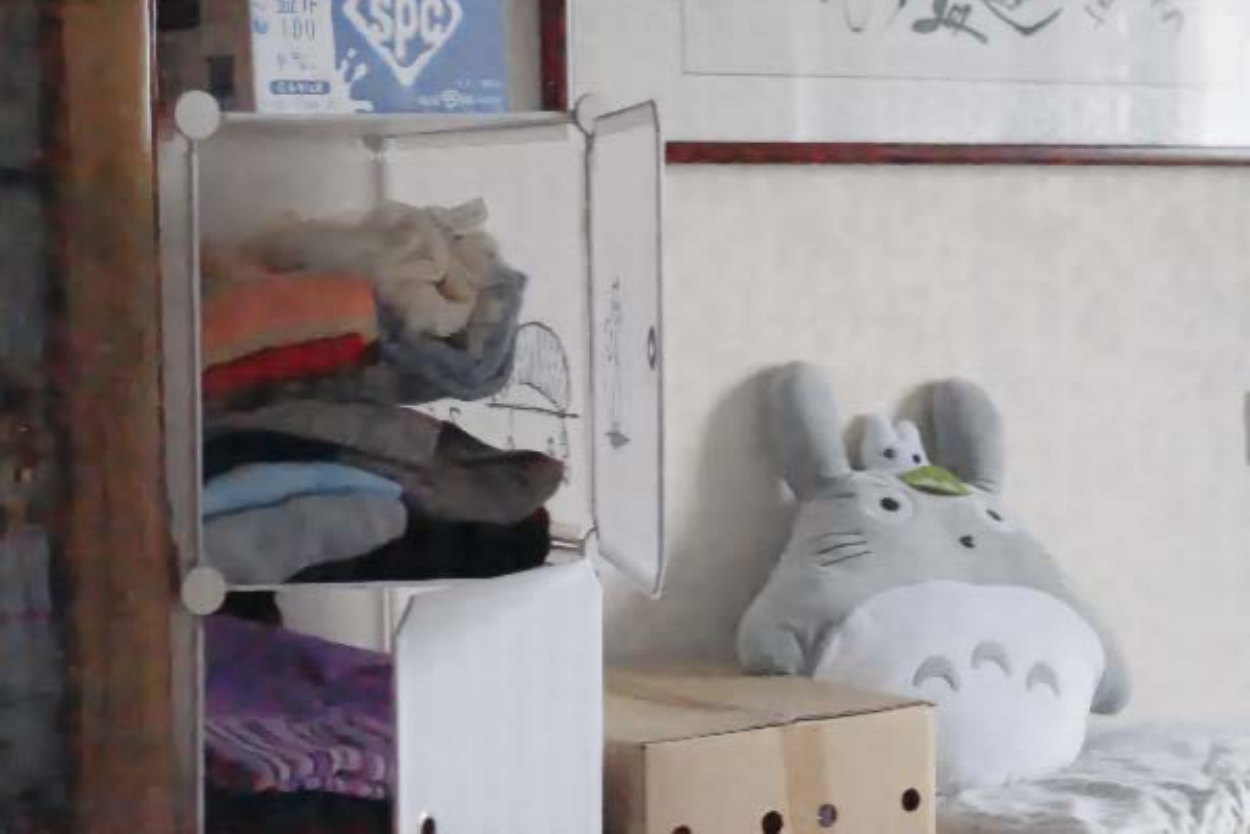}}
    \subfloat{\includegraphics[width = .10 \linewidth]{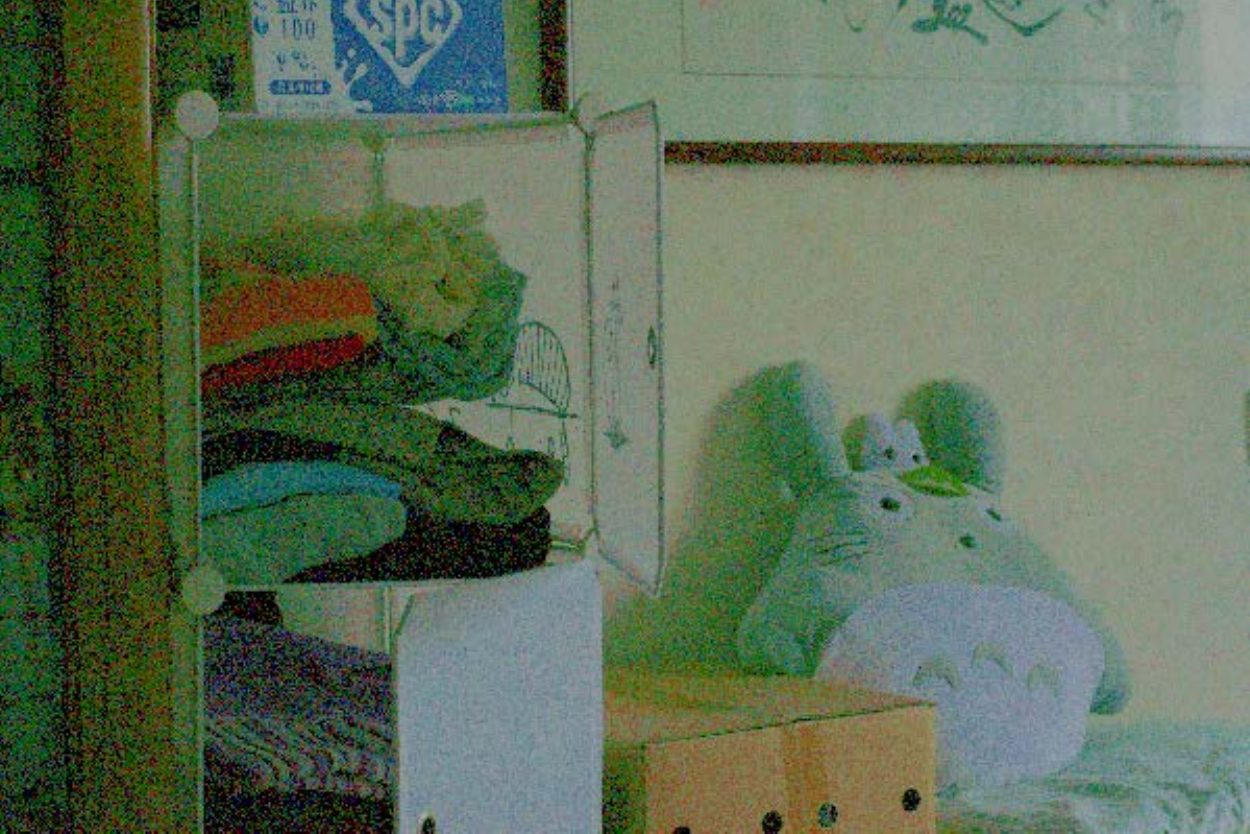}}
    \subfloat{\includegraphics[width = .10 \linewidth]{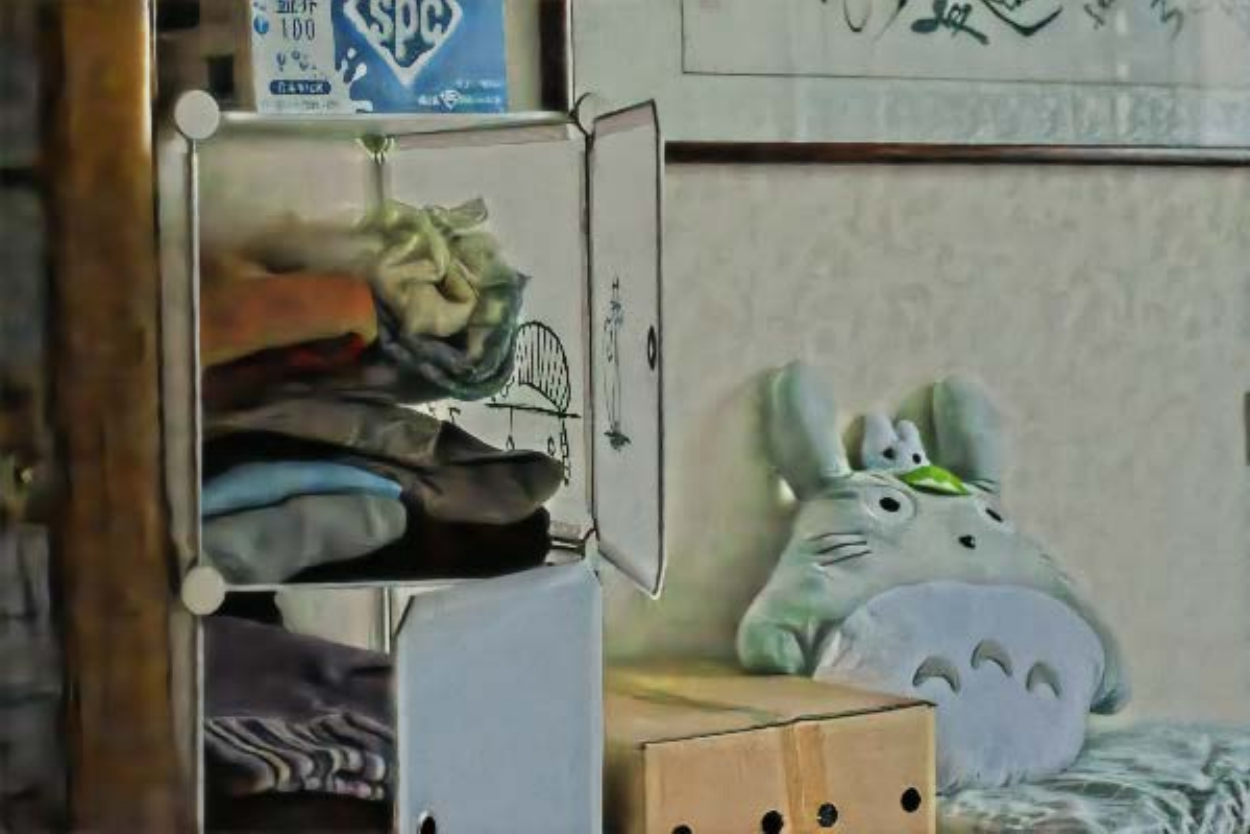}}
    \subfloat{\includegraphics[width = .10 \linewidth]{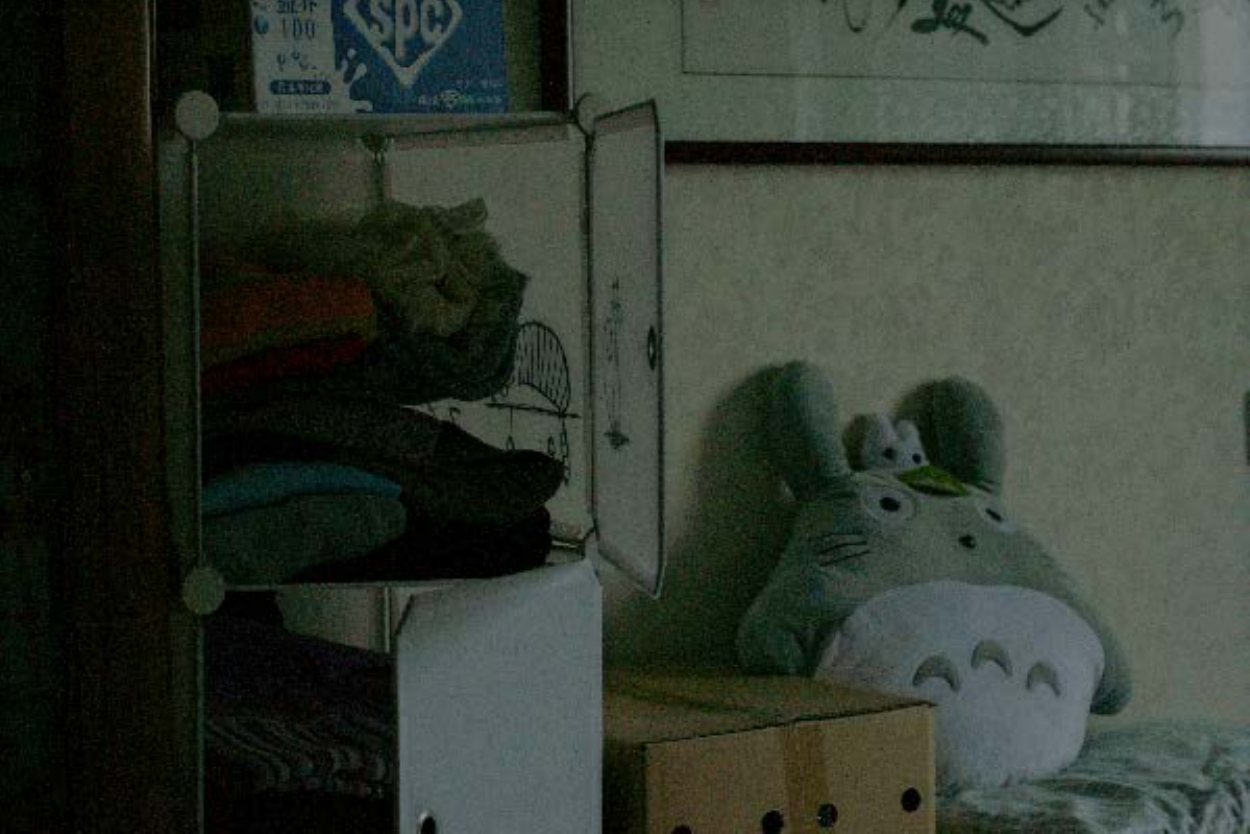}}\\\vspace{-0.16in}
    \subfloat{\includegraphics[width = .10 \linewidth]{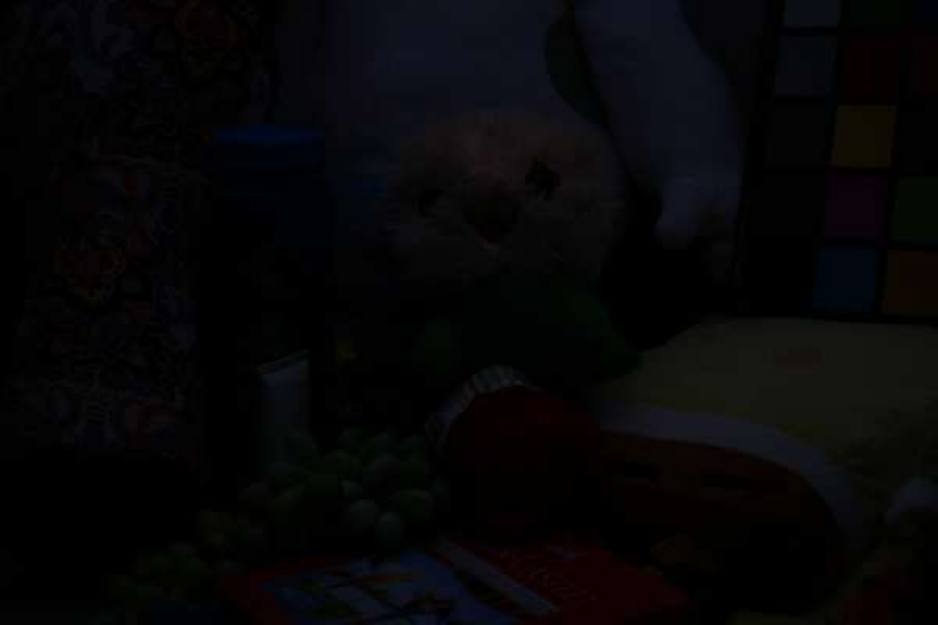}}
    \subfloat{\includegraphics[width = .10 \linewidth]{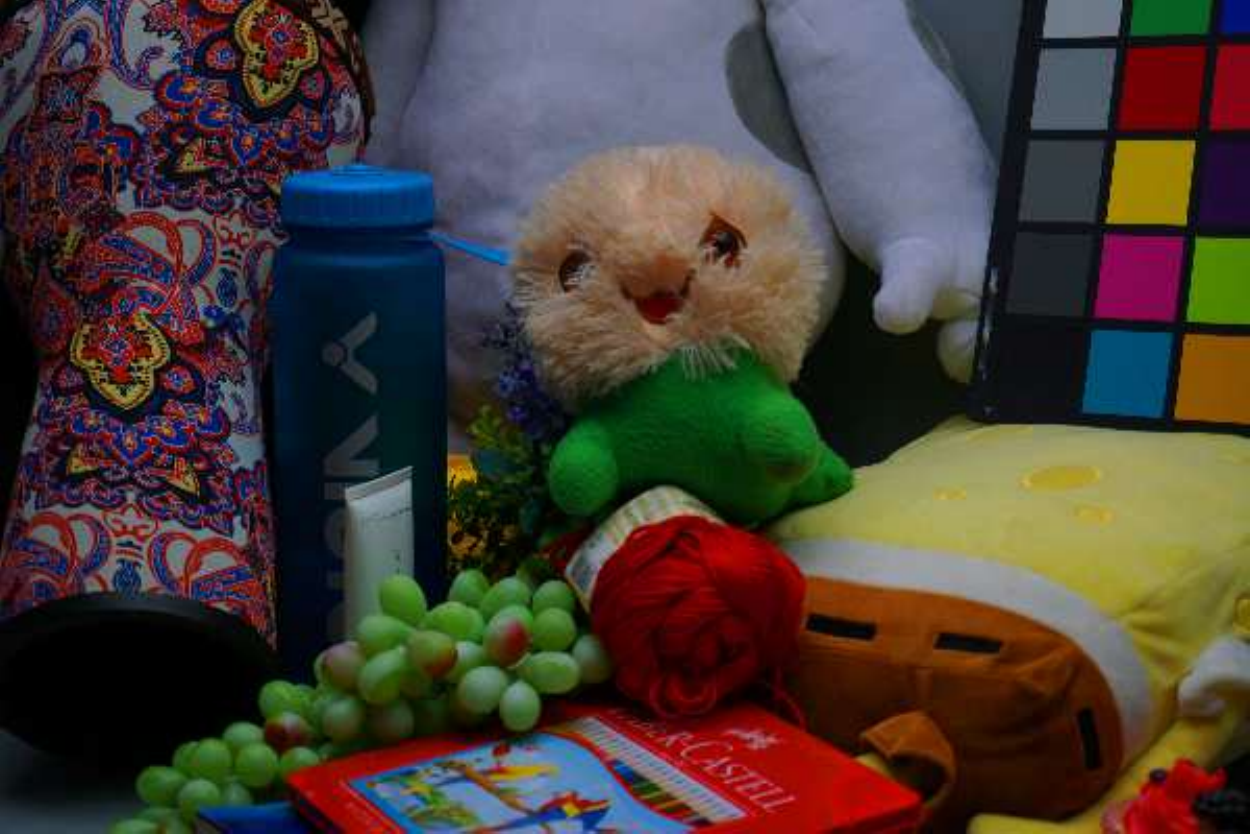}}
    \subfloat{\includegraphics[width = .10 \linewidth]{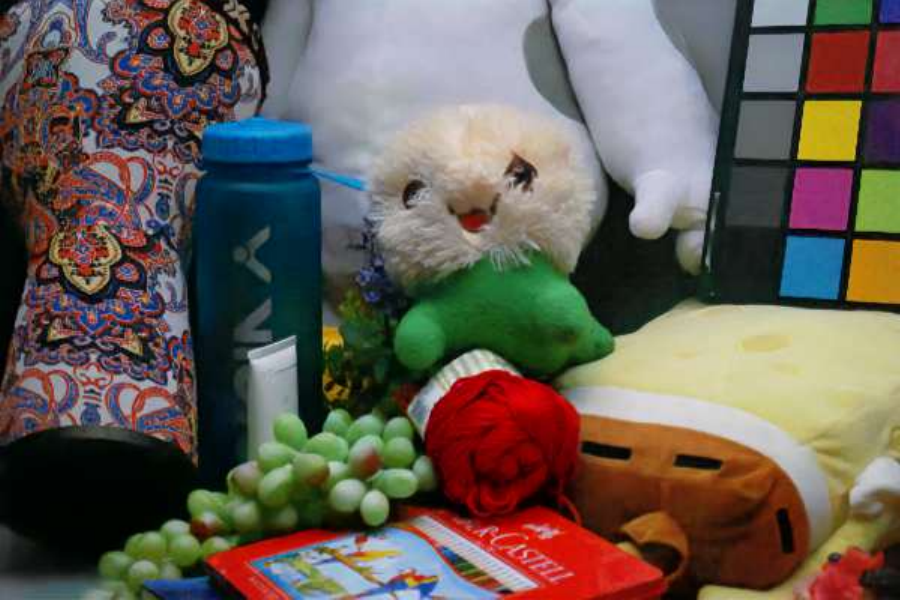}}
    \subfloat{\includegraphics[width = .10 \linewidth]{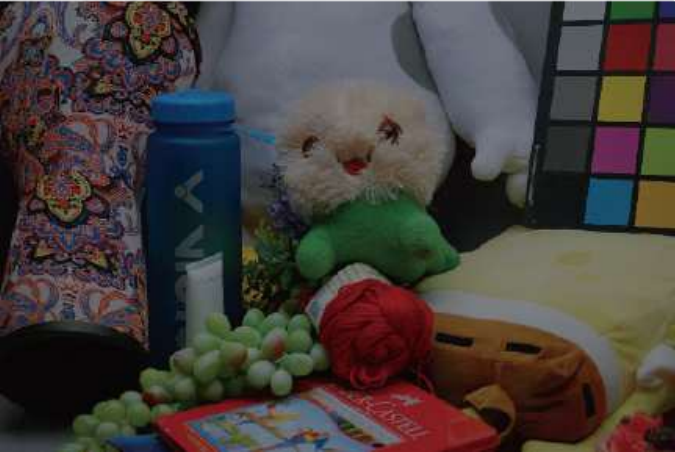}}
    \subfloat{\includegraphics[width = .10 \linewidth]{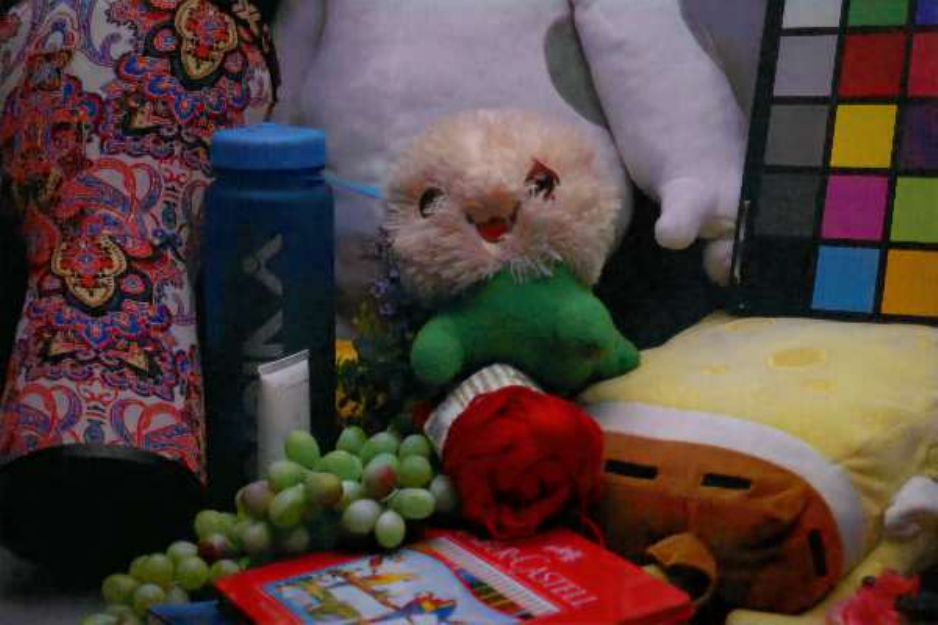}}
    \subfloat{\includegraphics[width = .10 \linewidth]{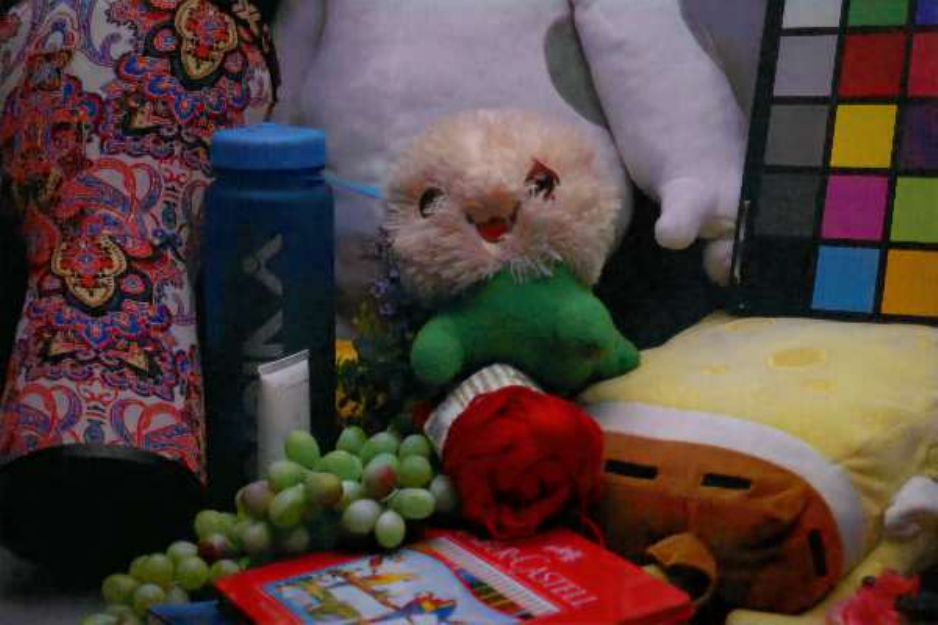}}
    \subfloat{\includegraphics[width = .10 \linewidth]{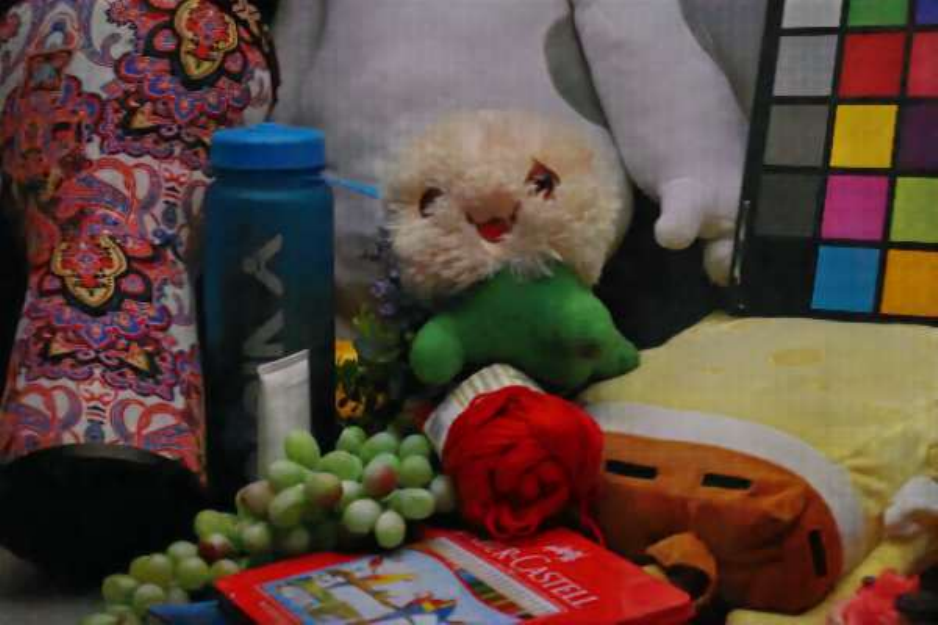}}
    \subfloat{\includegraphics[width = .10 \linewidth]{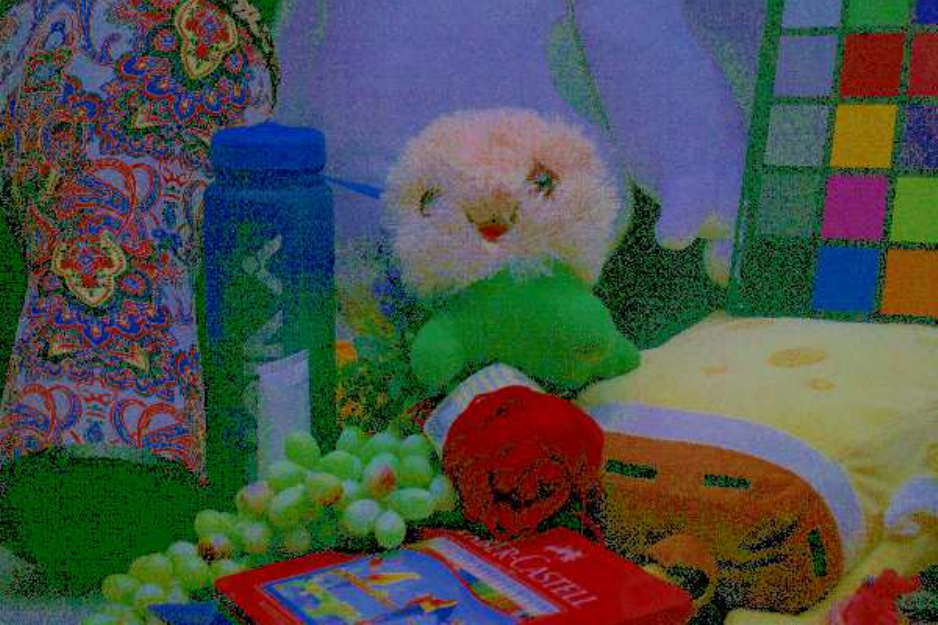}}
    \subfloat{\includegraphics[width = .10 \linewidth]{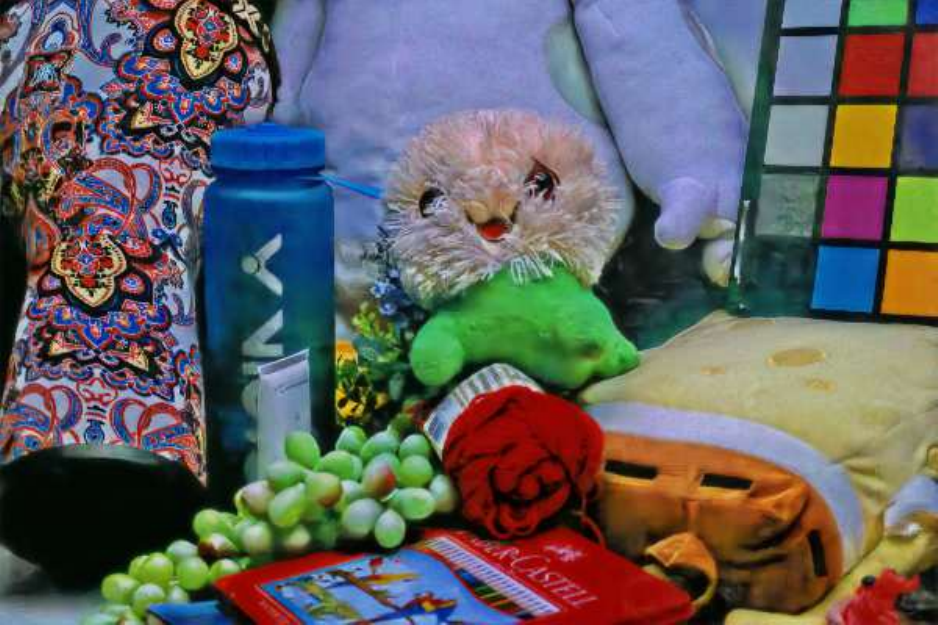}}
    \subfloat{\includegraphics[width = .10 \linewidth]{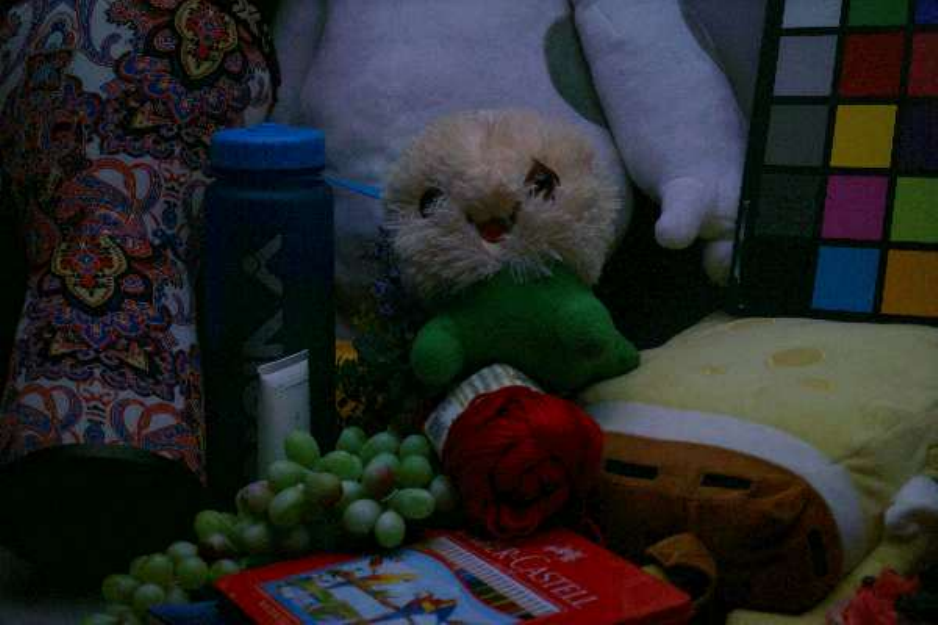}}\\\vspace{-0.16in}
    \subfloat{\includegraphics[width = .10 \linewidth]{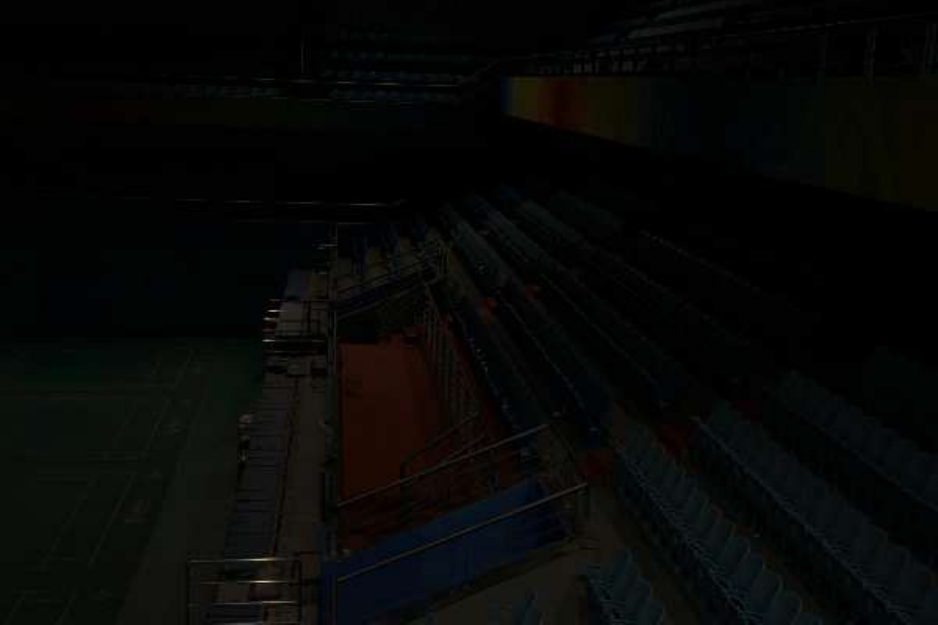}}
    \subfloat{\includegraphics[width = .10 \linewidth]{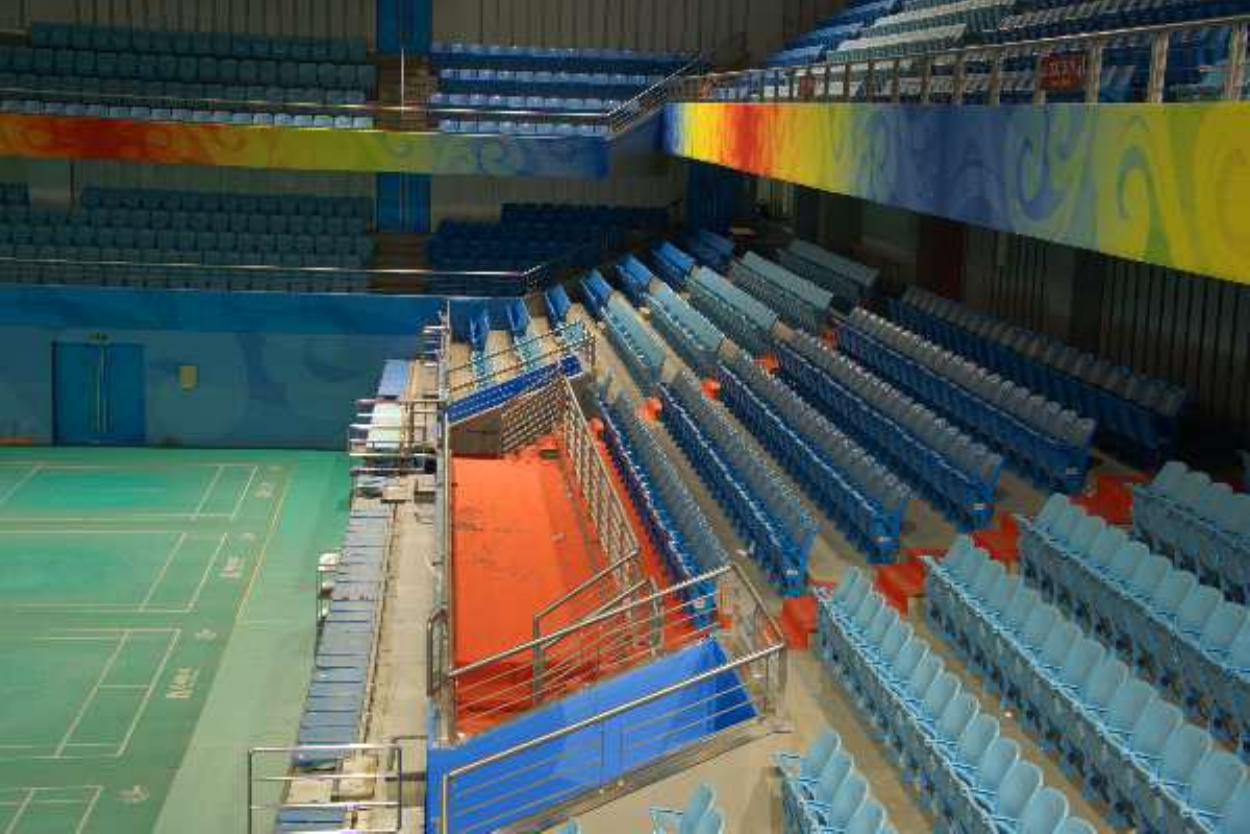}}
    \subfloat{\includegraphics[width = .10 \linewidth]{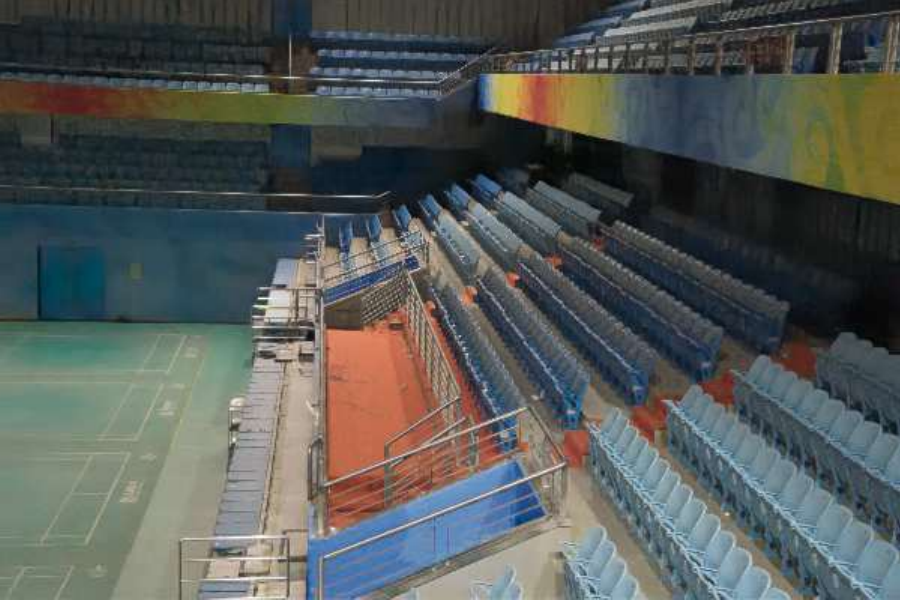}}
    \subfloat{\includegraphics[width = .10 \linewidth]{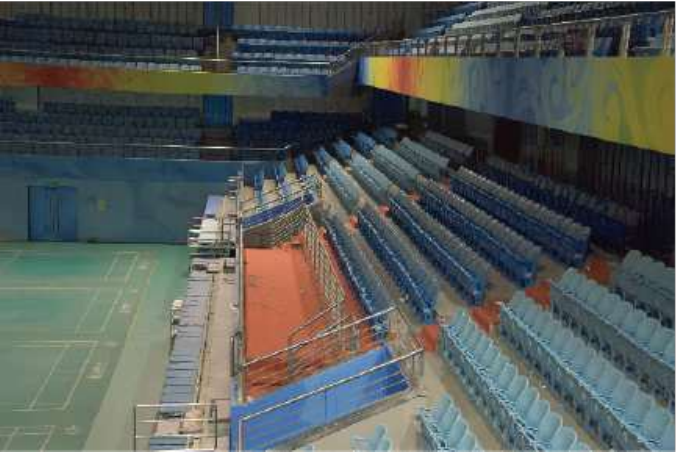}}
    \subfloat{\includegraphics[width = .10 \linewidth]{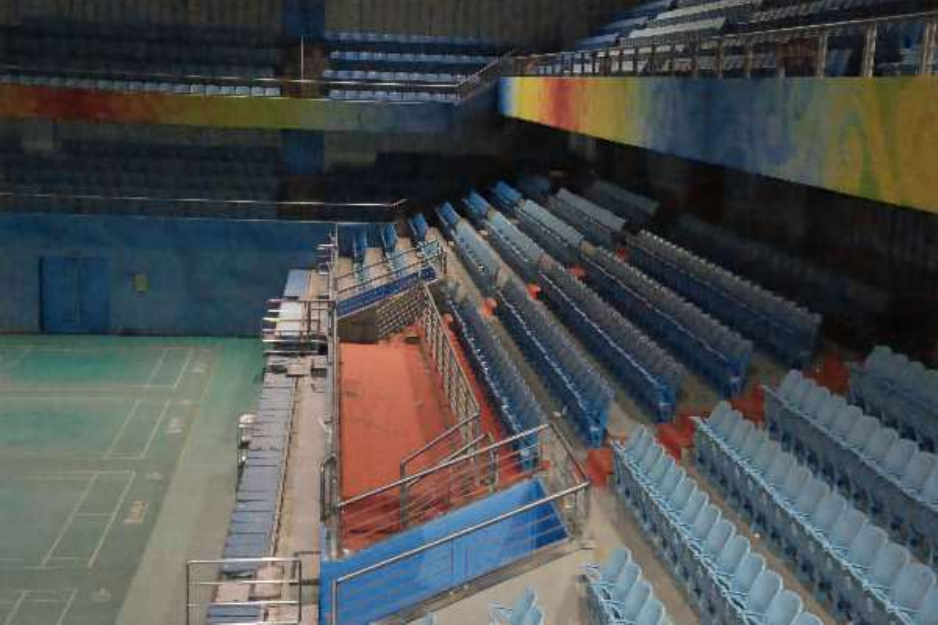}}
    \subfloat{\includegraphics[width = .10 \linewidth]{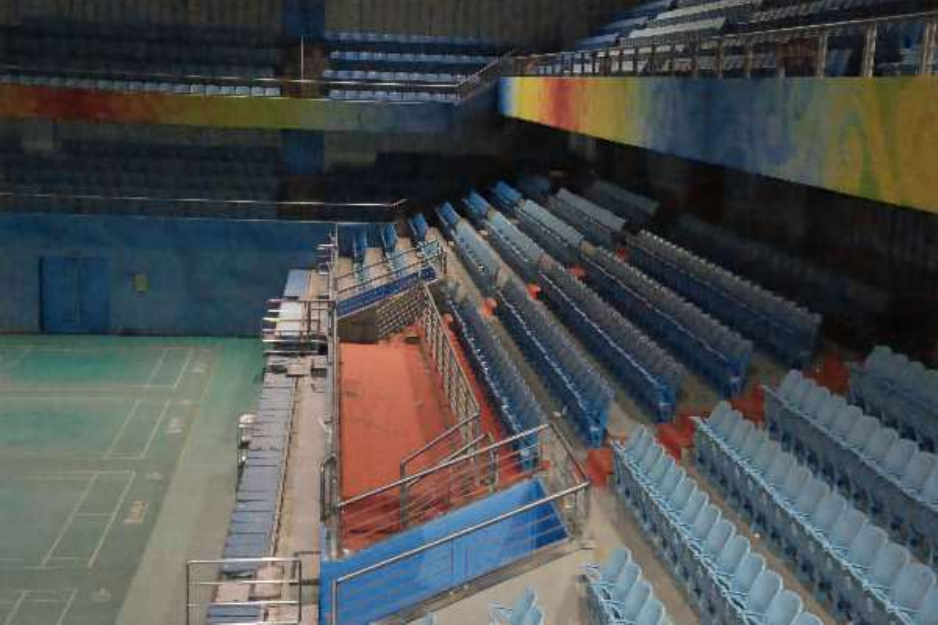}}
    \subfloat{\includegraphics[width = .10 \linewidth]{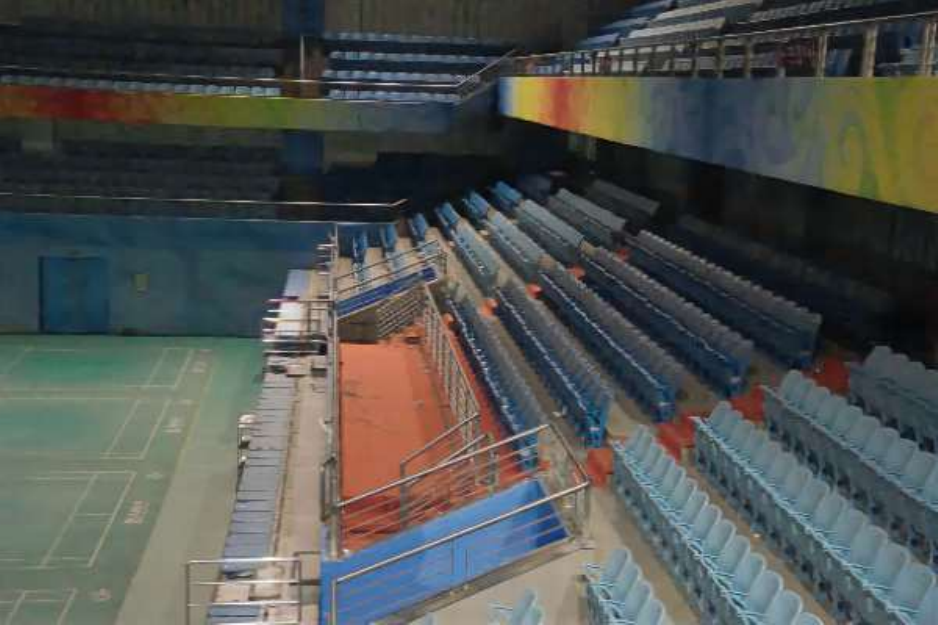}}
    \subfloat{\includegraphics[width = .10 \linewidth]{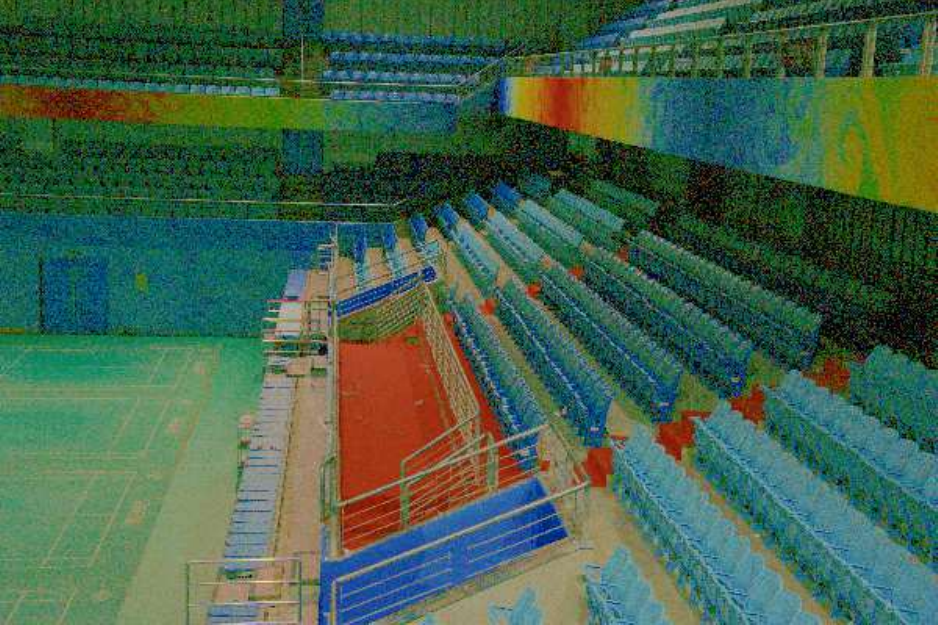}}
    \subfloat{\includegraphics[width = .10 \linewidth]{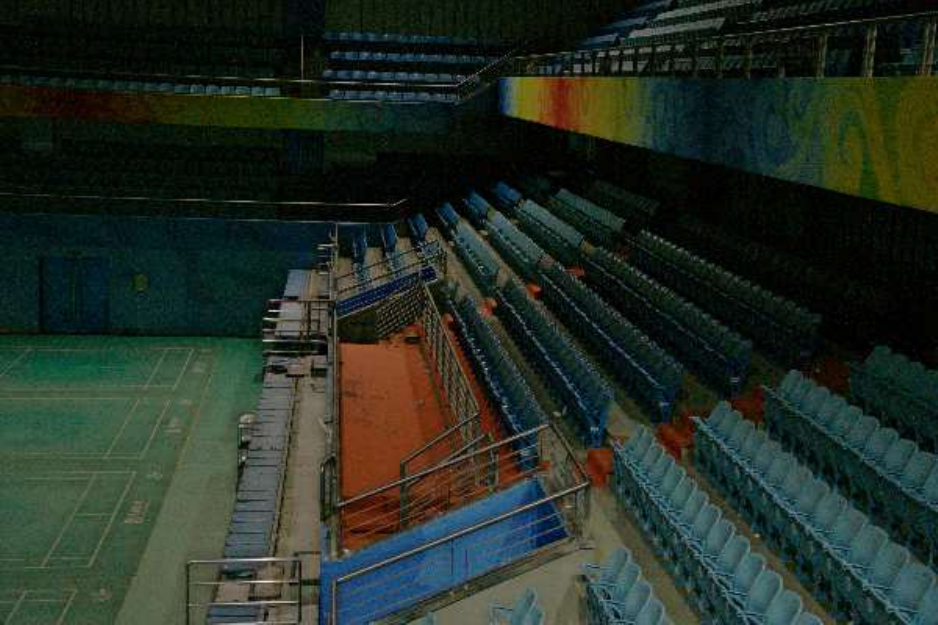}}
    \subfloat{\includegraphics[width = .10 \linewidth]{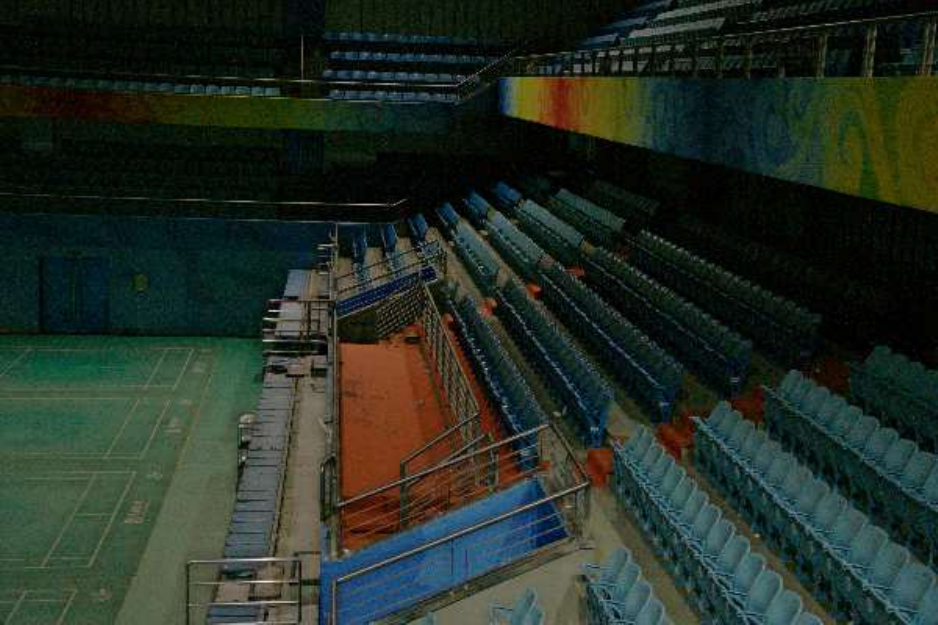}}\\\vspace{-0.1in}
    \subfloat{\includegraphics[width = .10 \linewidth]{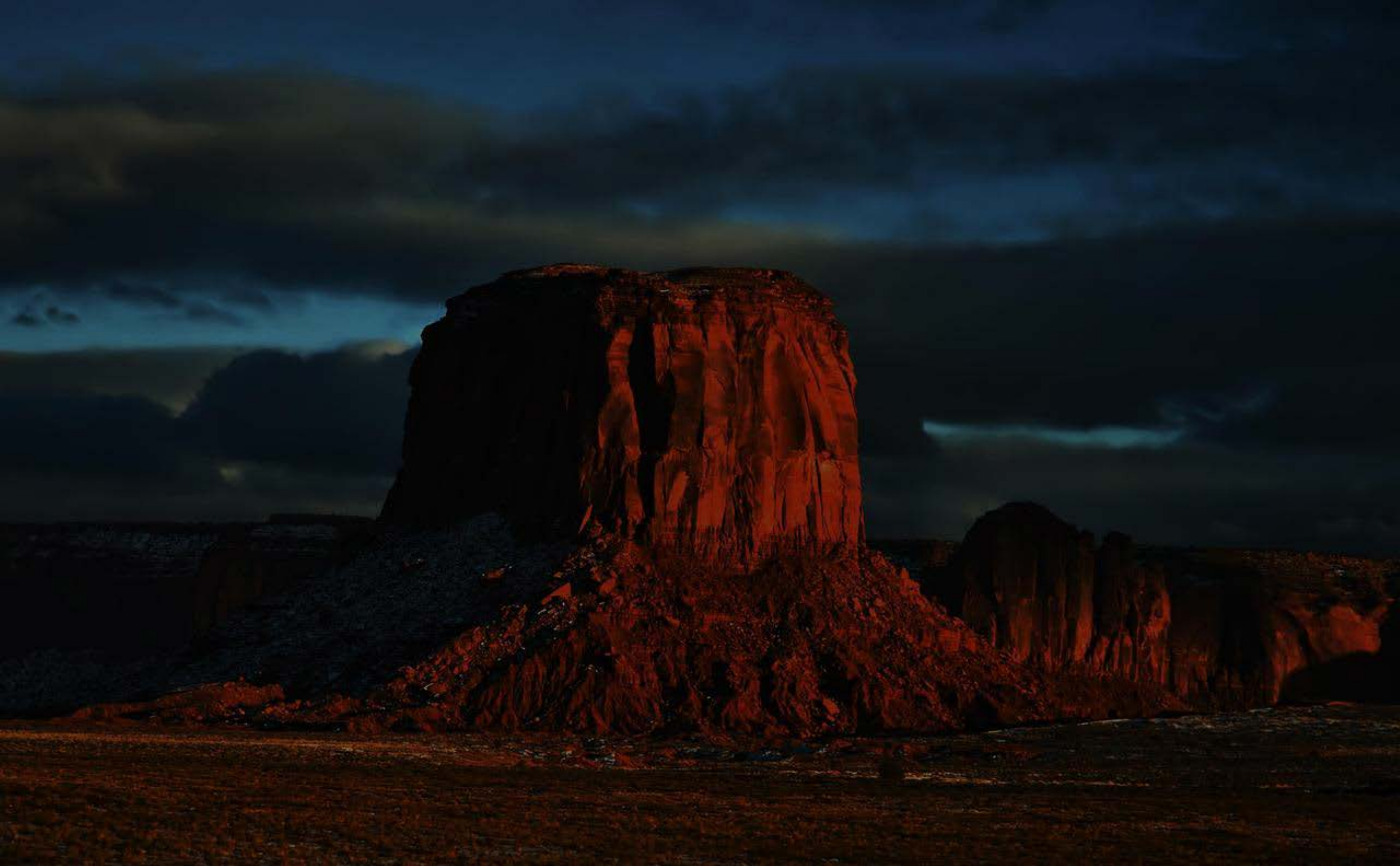}}
    \subfloat{\includegraphics[width = .10 \linewidth]{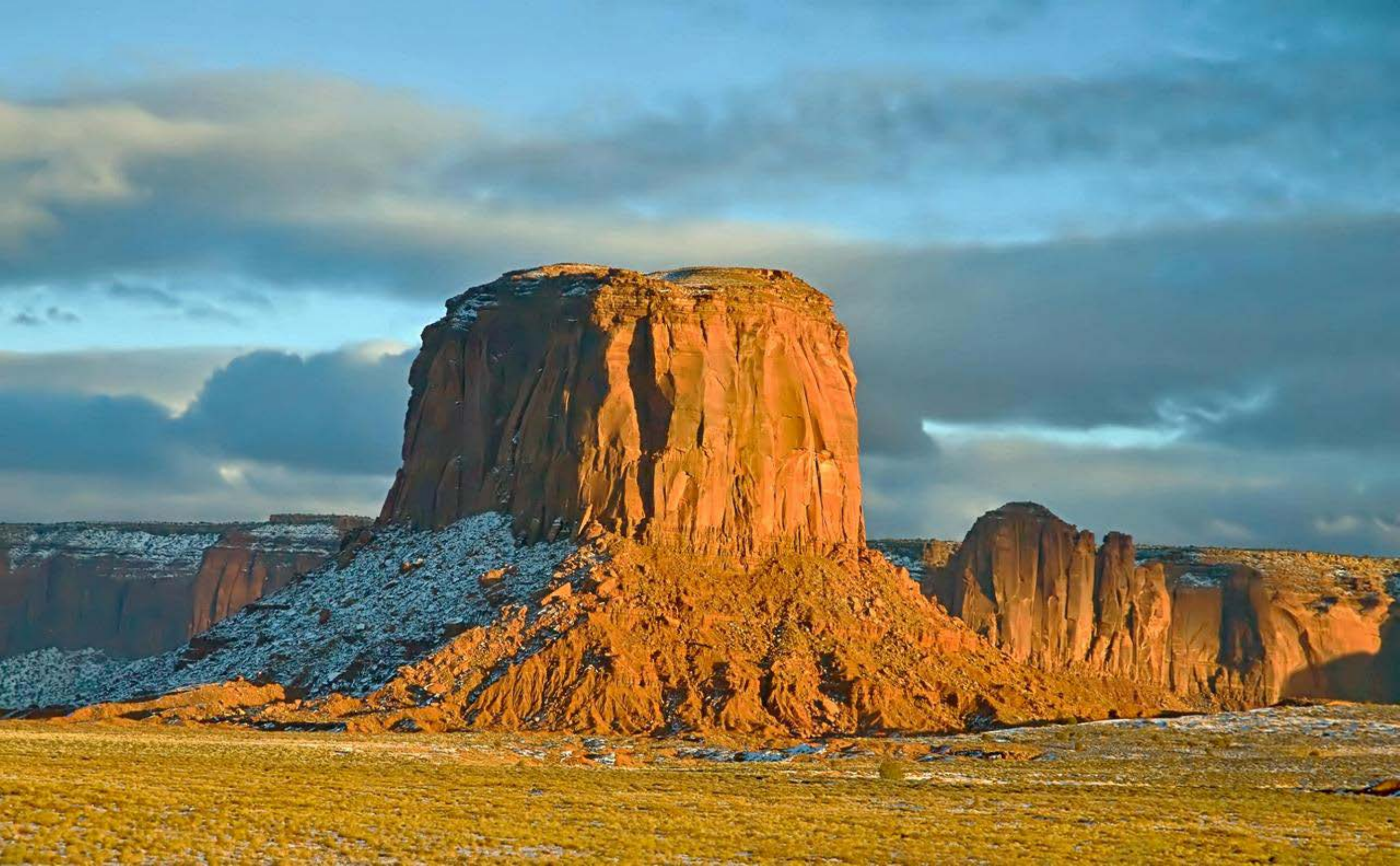}}
    \subfloat{\includegraphics[width = .10 \linewidth]{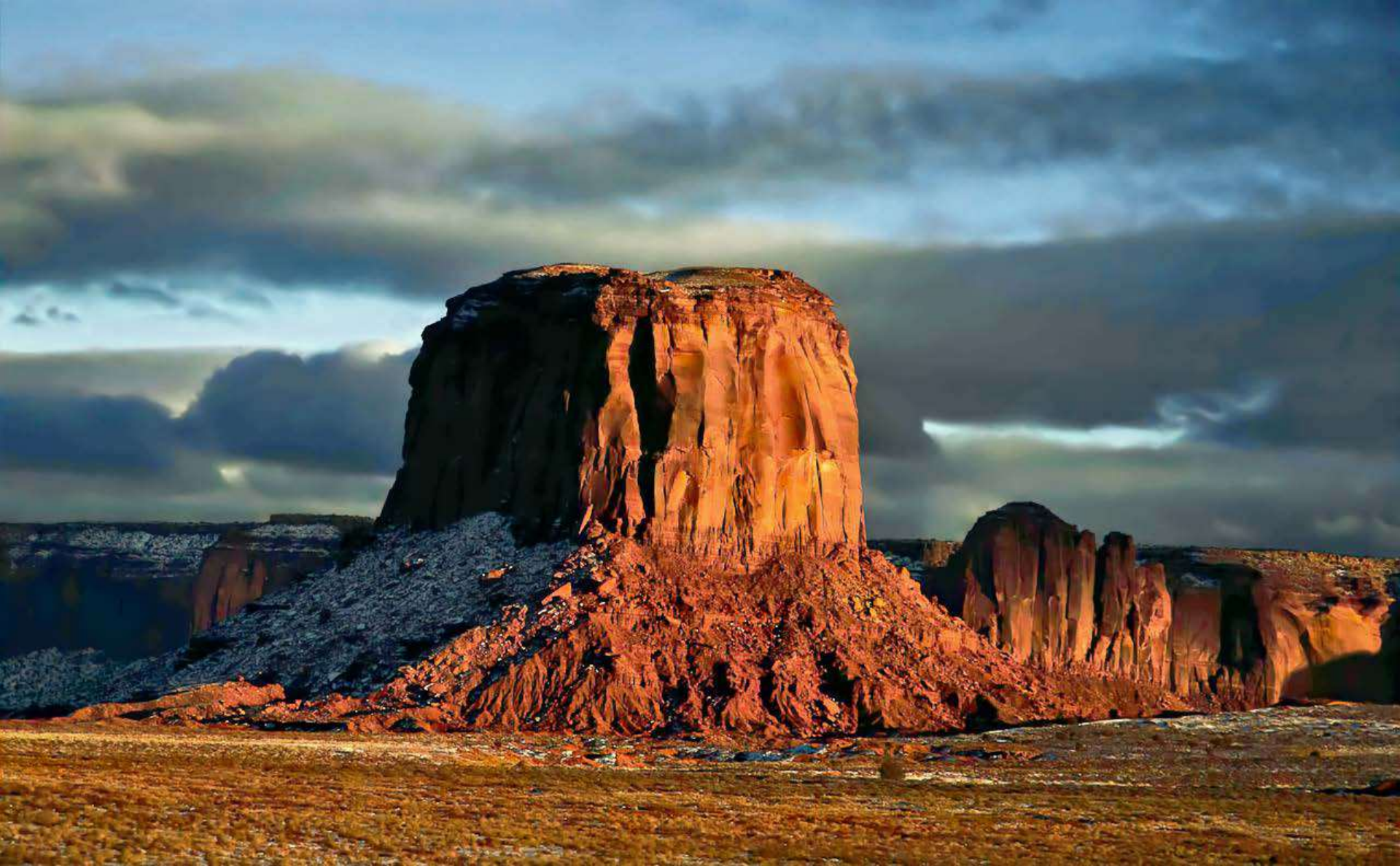}}
    \subfloat{\includegraphics[width = .10 \linewidth]{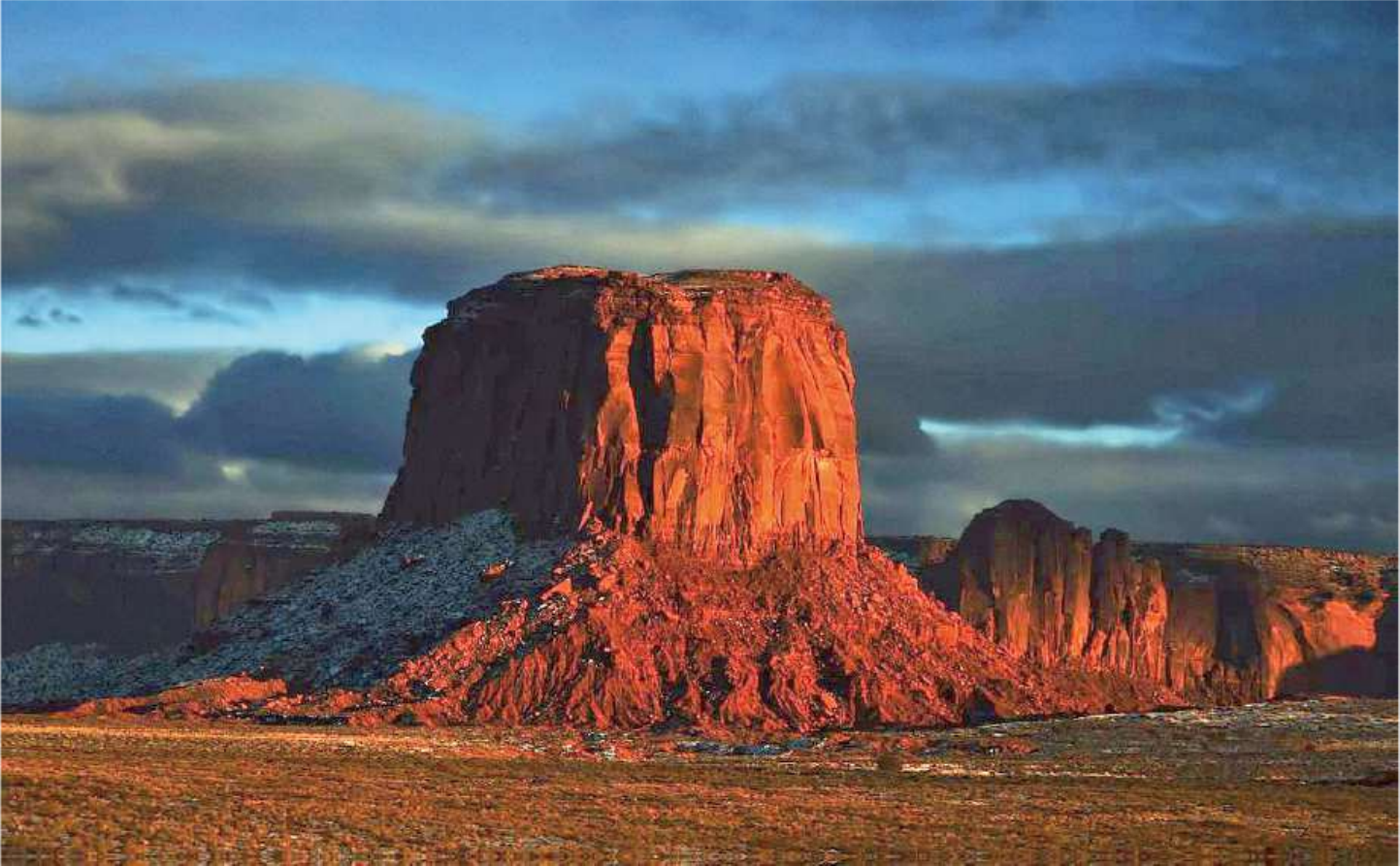}}
    \subfloat{\includegraphics[width = .10 \linewidth]{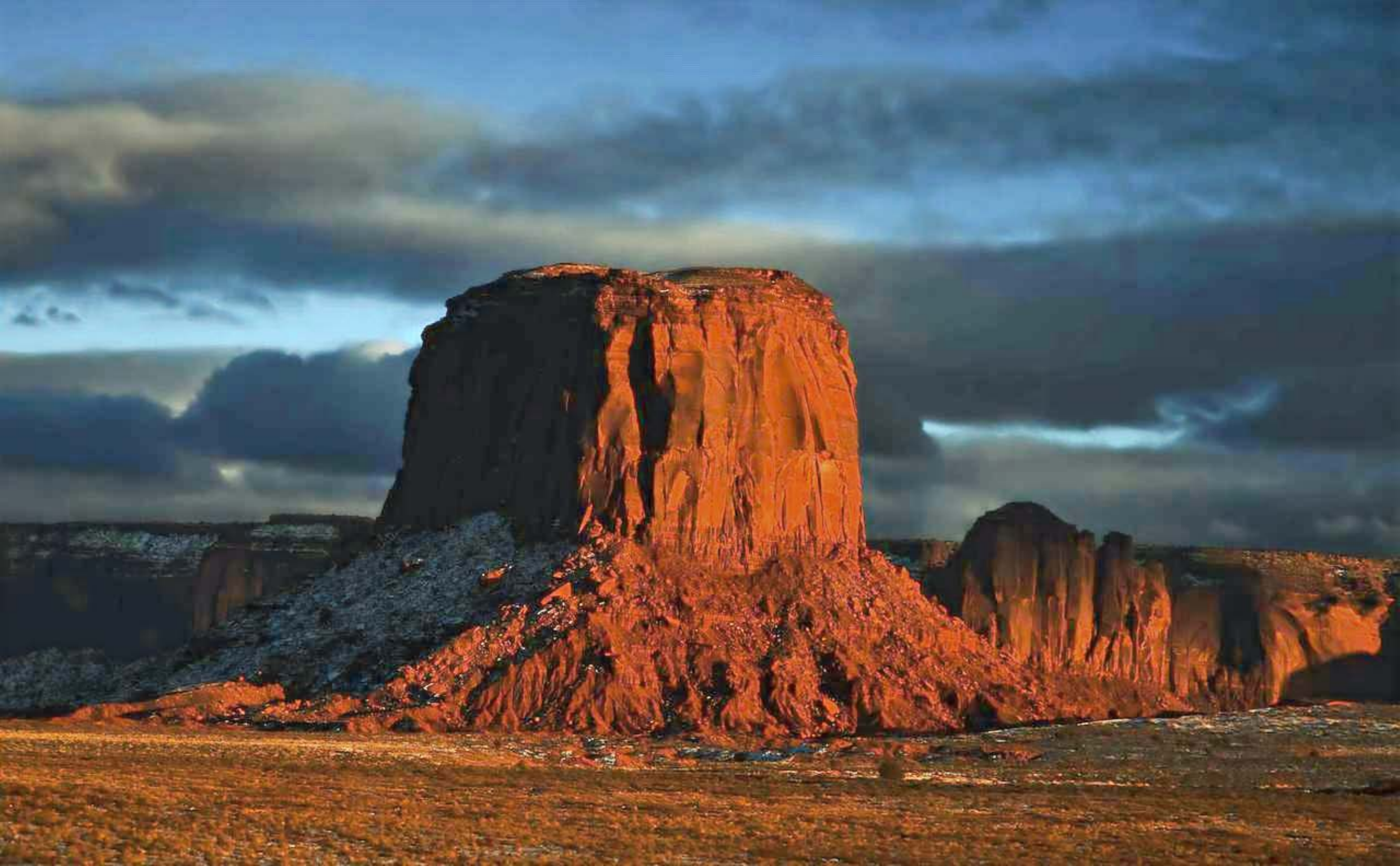}}
    \subfloat{\includegraphics[width = .10 \linewidth]{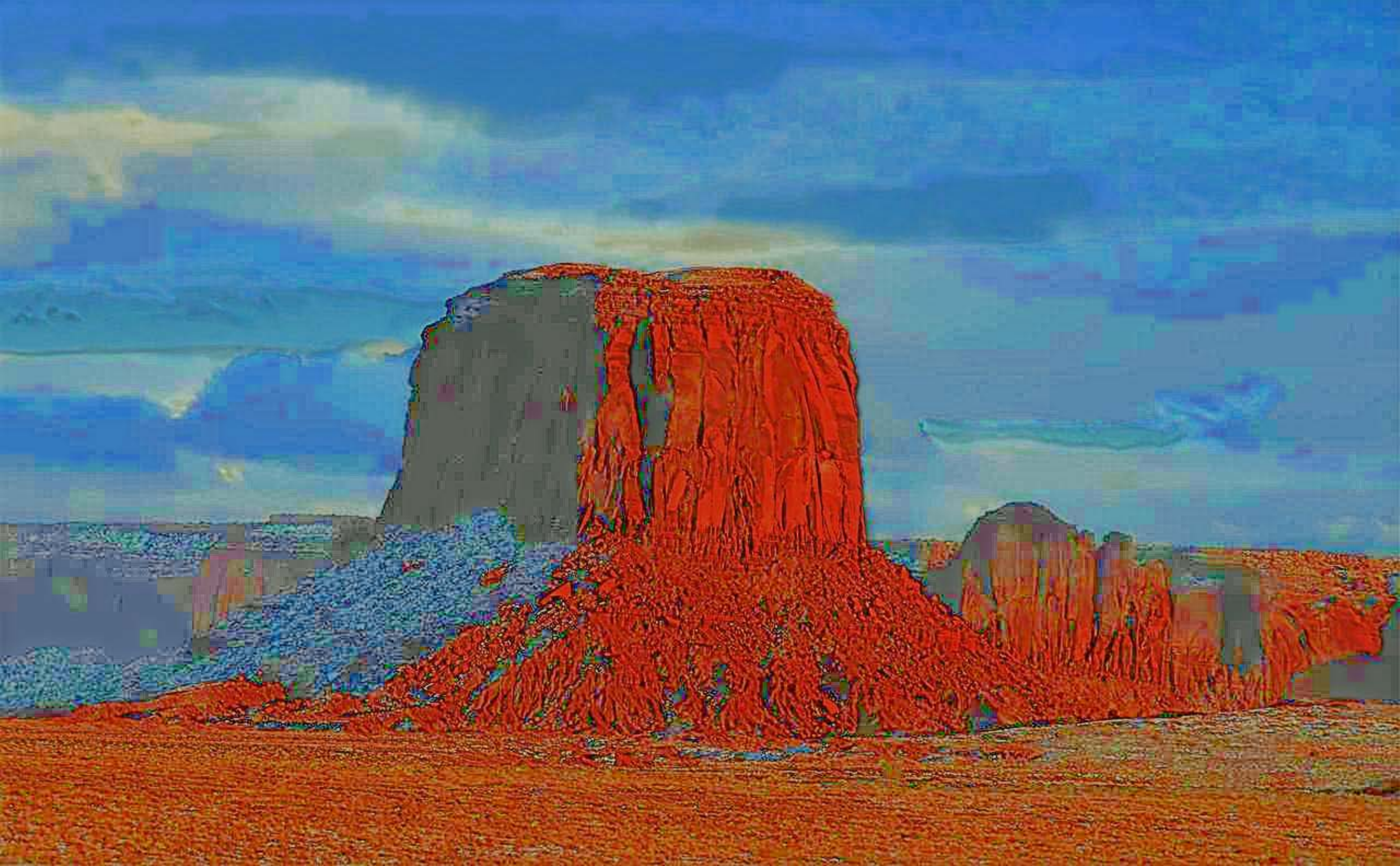}}
    \subfloat{\includegraphics[width = .10 \linewidth]{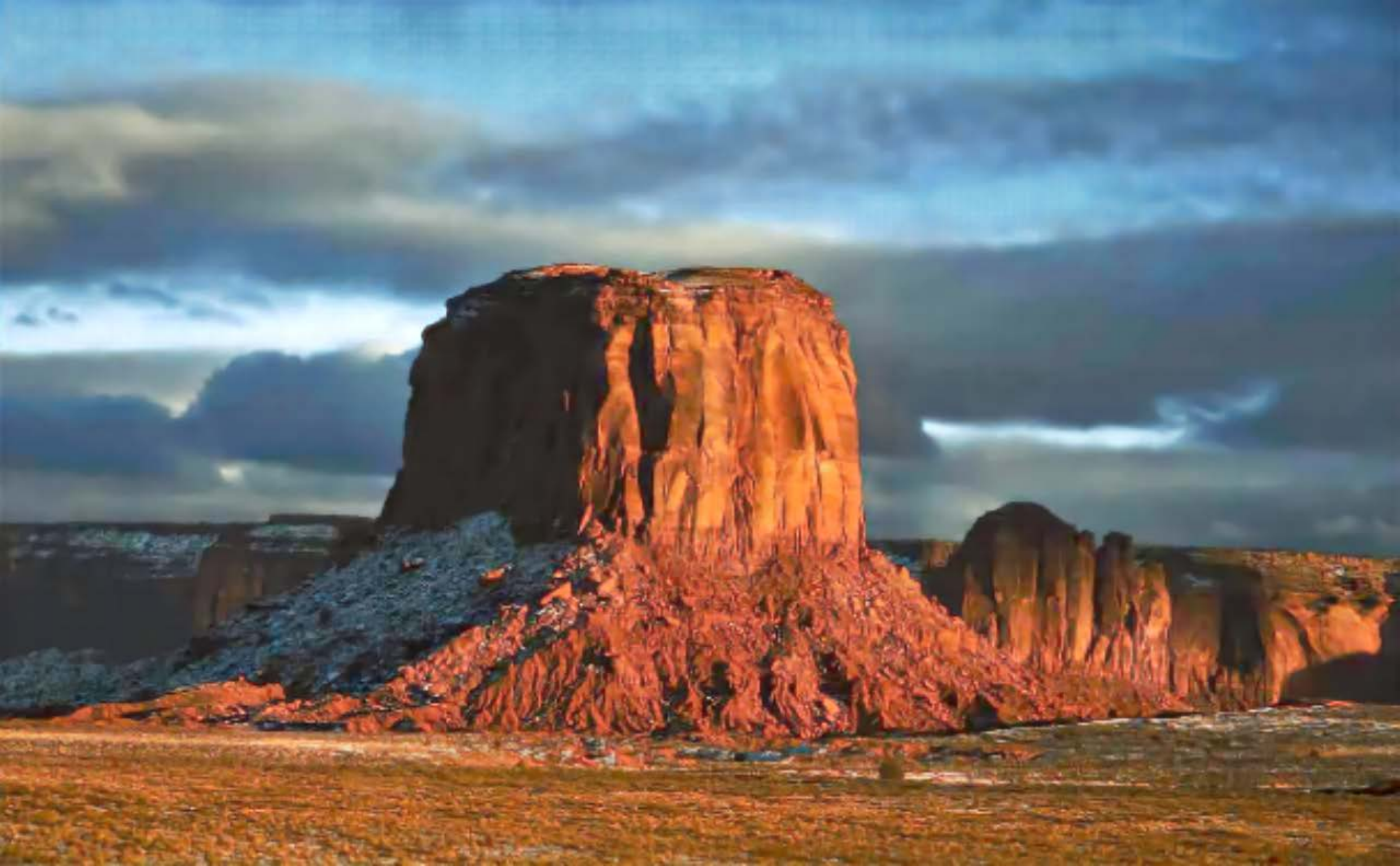}}
    \subfloat{\includegraphics[width = .10 \linewidth]{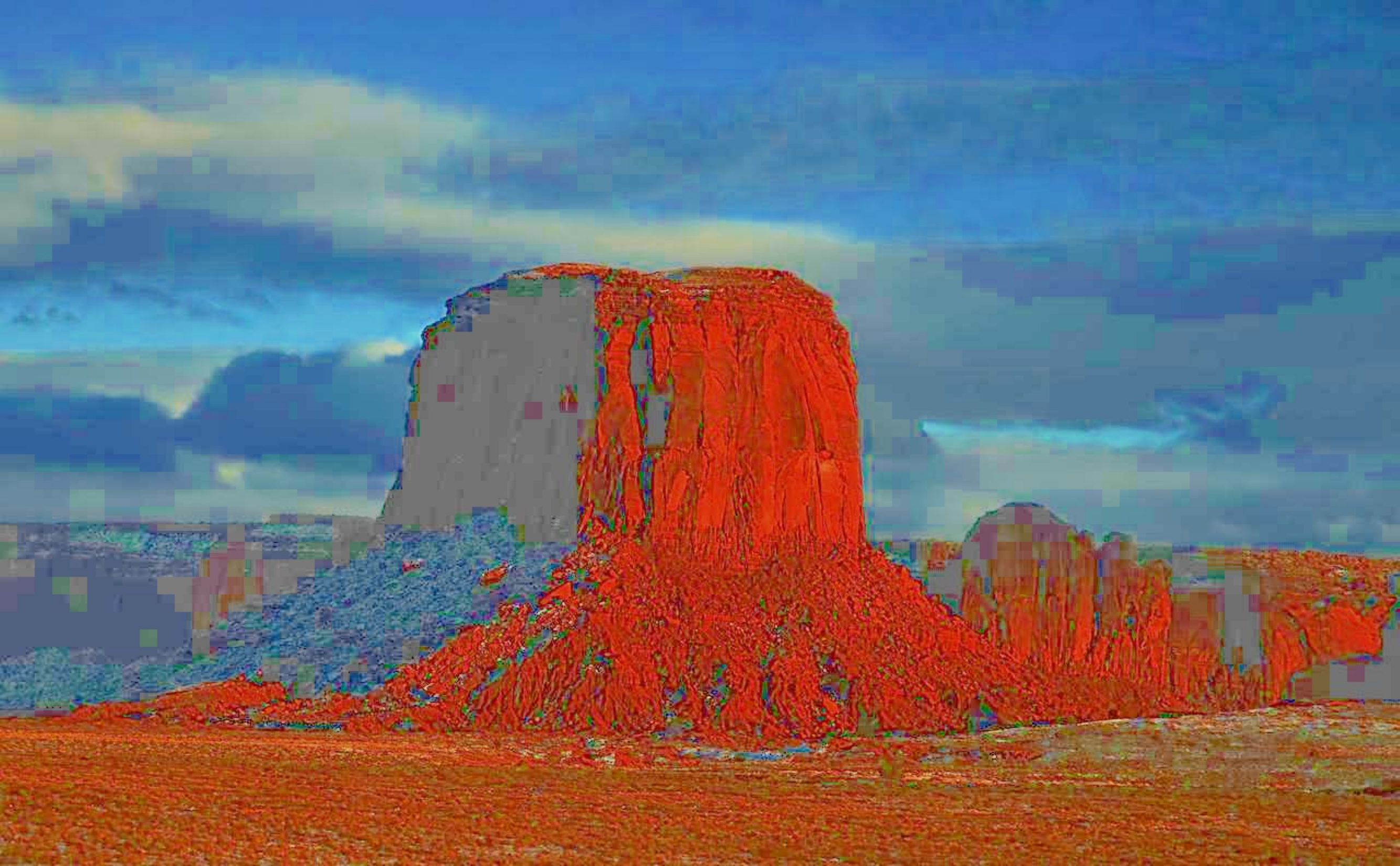}}
    \subfloat{\includegraphics[width = .10 \linewidth]{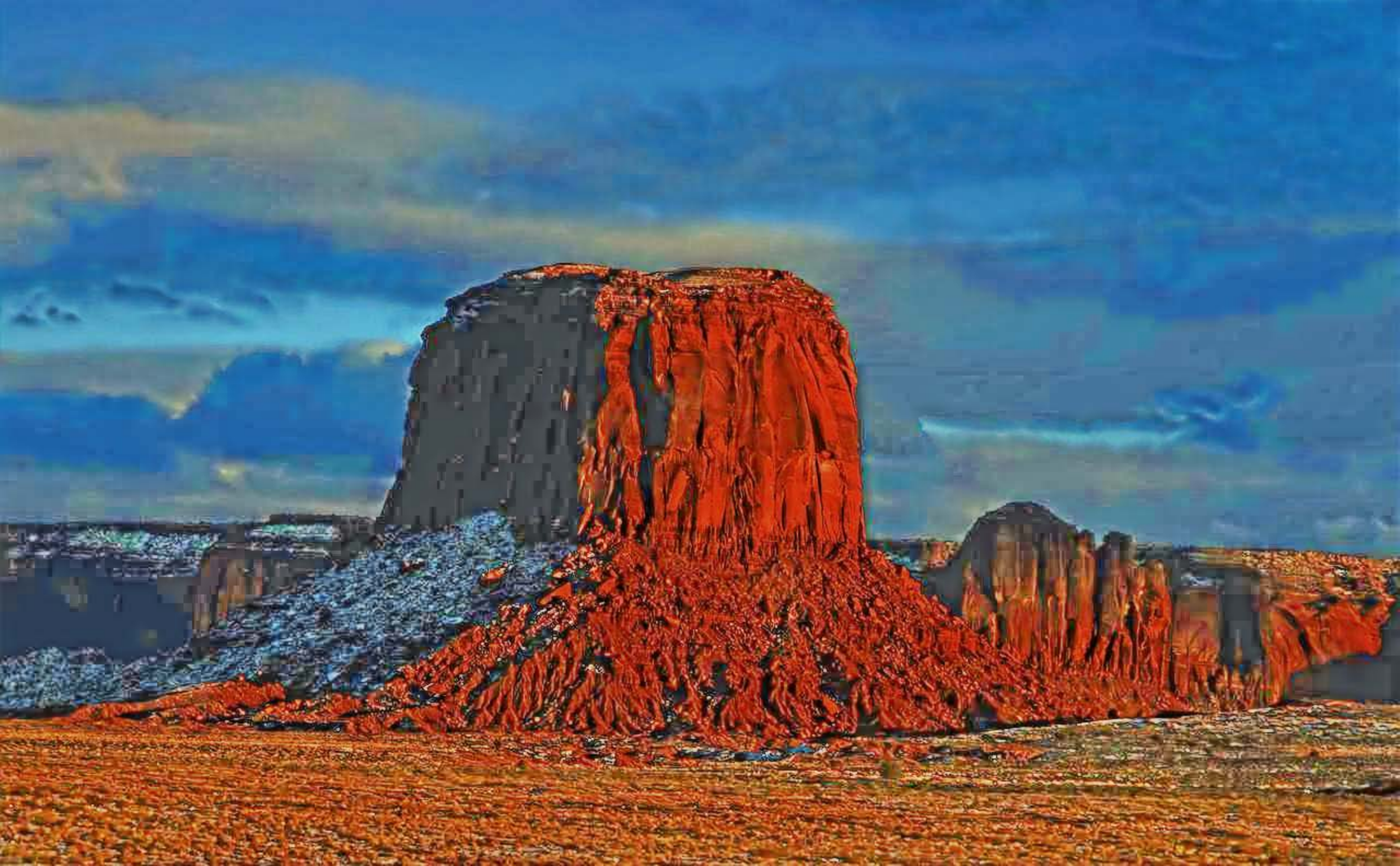}}
    \subfloat{\includegraphics[width = .10 \linewidth]{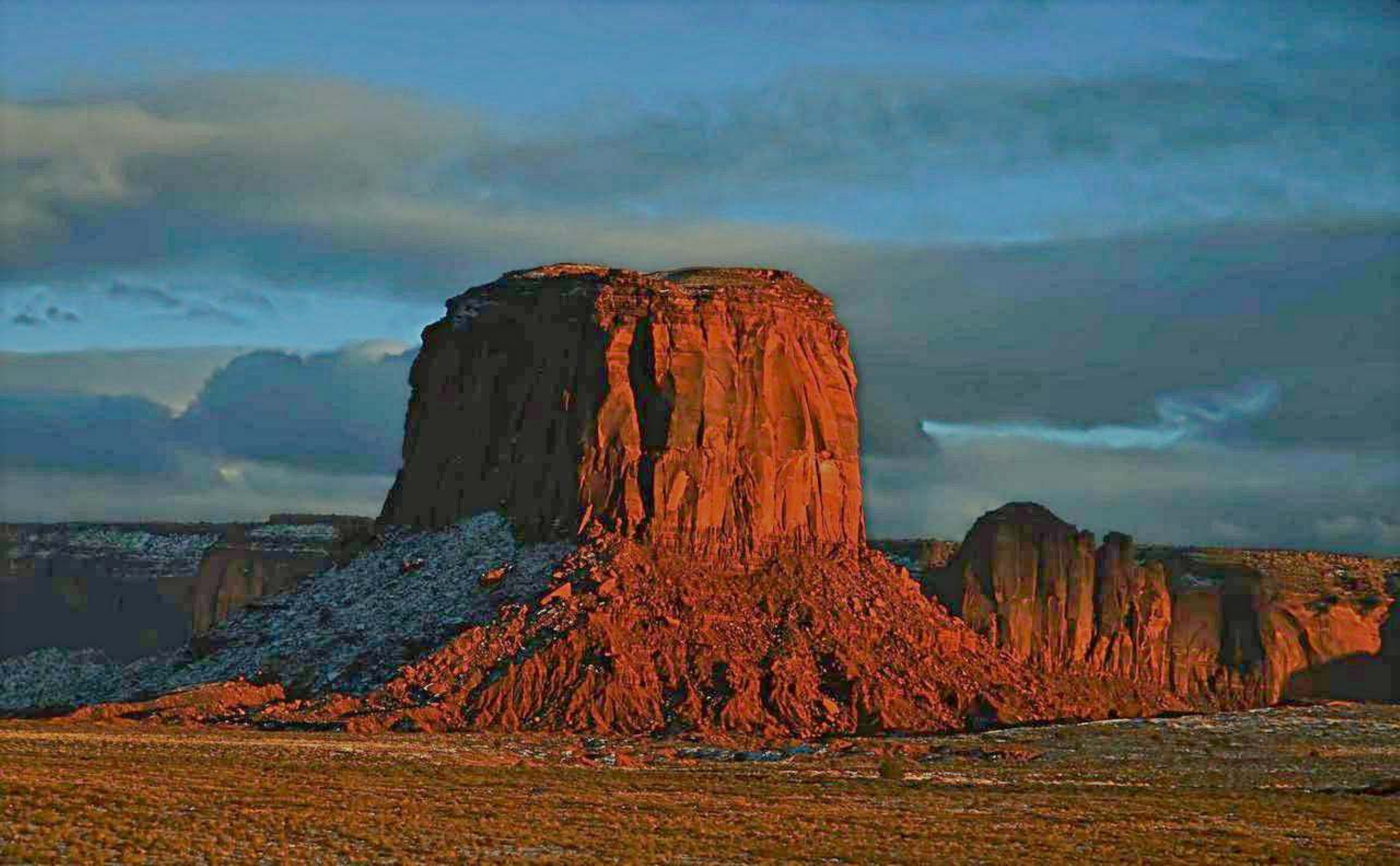}}\\\vspace{-0.16in}
    \subfloat{\includegraphics[width = .10 \linewidth]{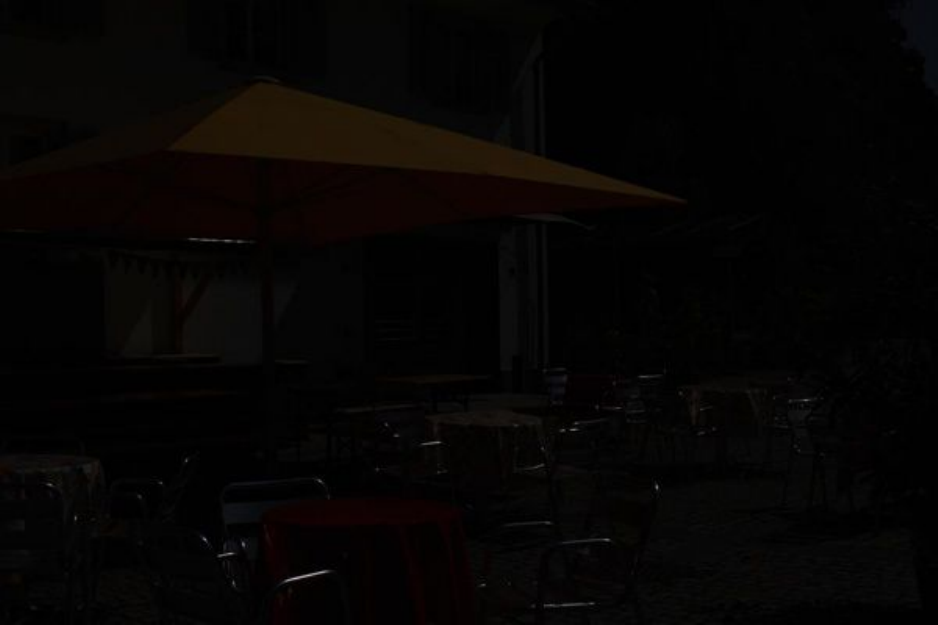}}
    \subfloat{\includegraphics[width = .10 \linewidth]{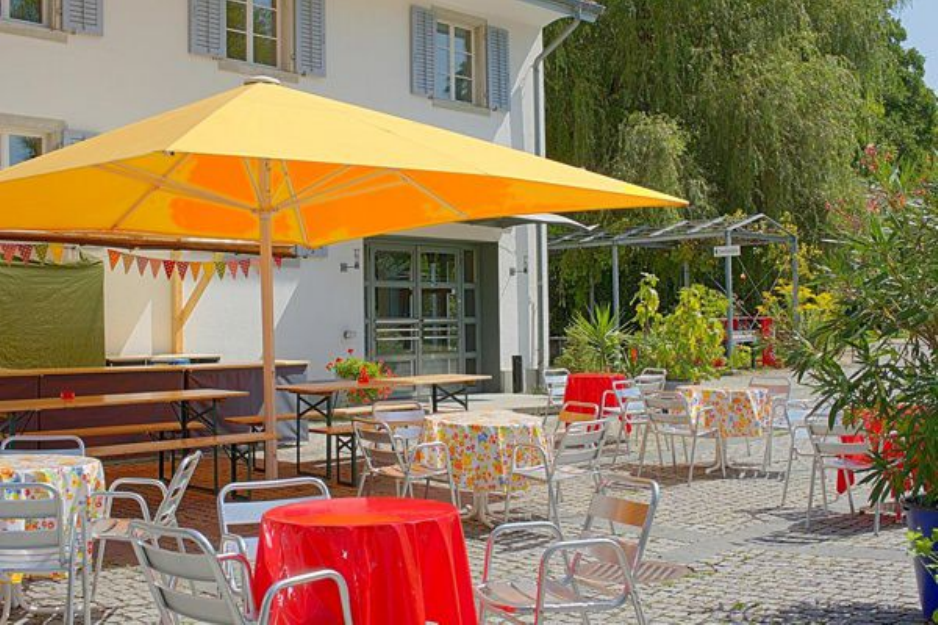}}
    \subfloat{\includegraphics[width = .10 \linewidth]{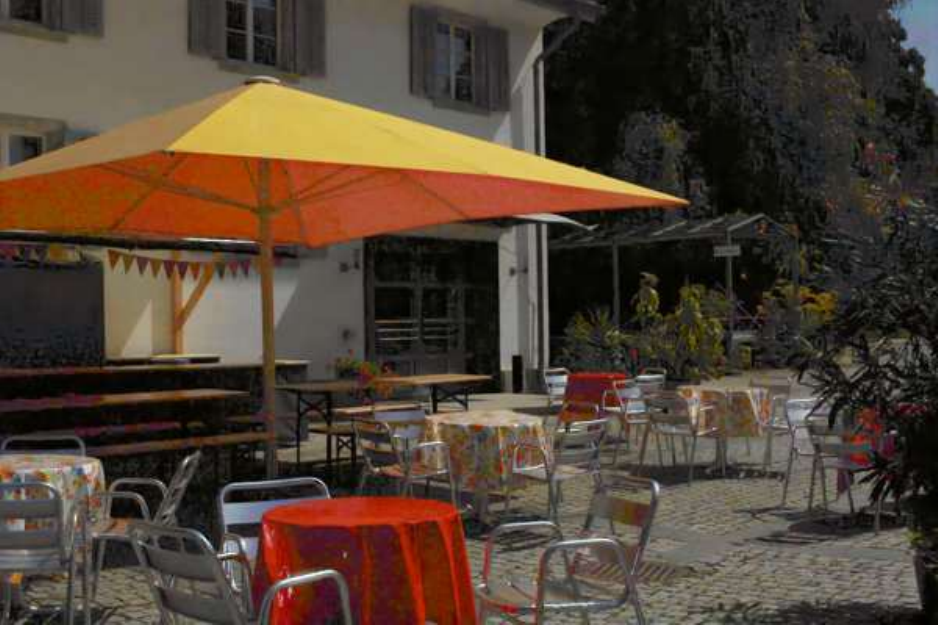}}
    \subfloat{\includegraphics[width = .10 \linewidth]{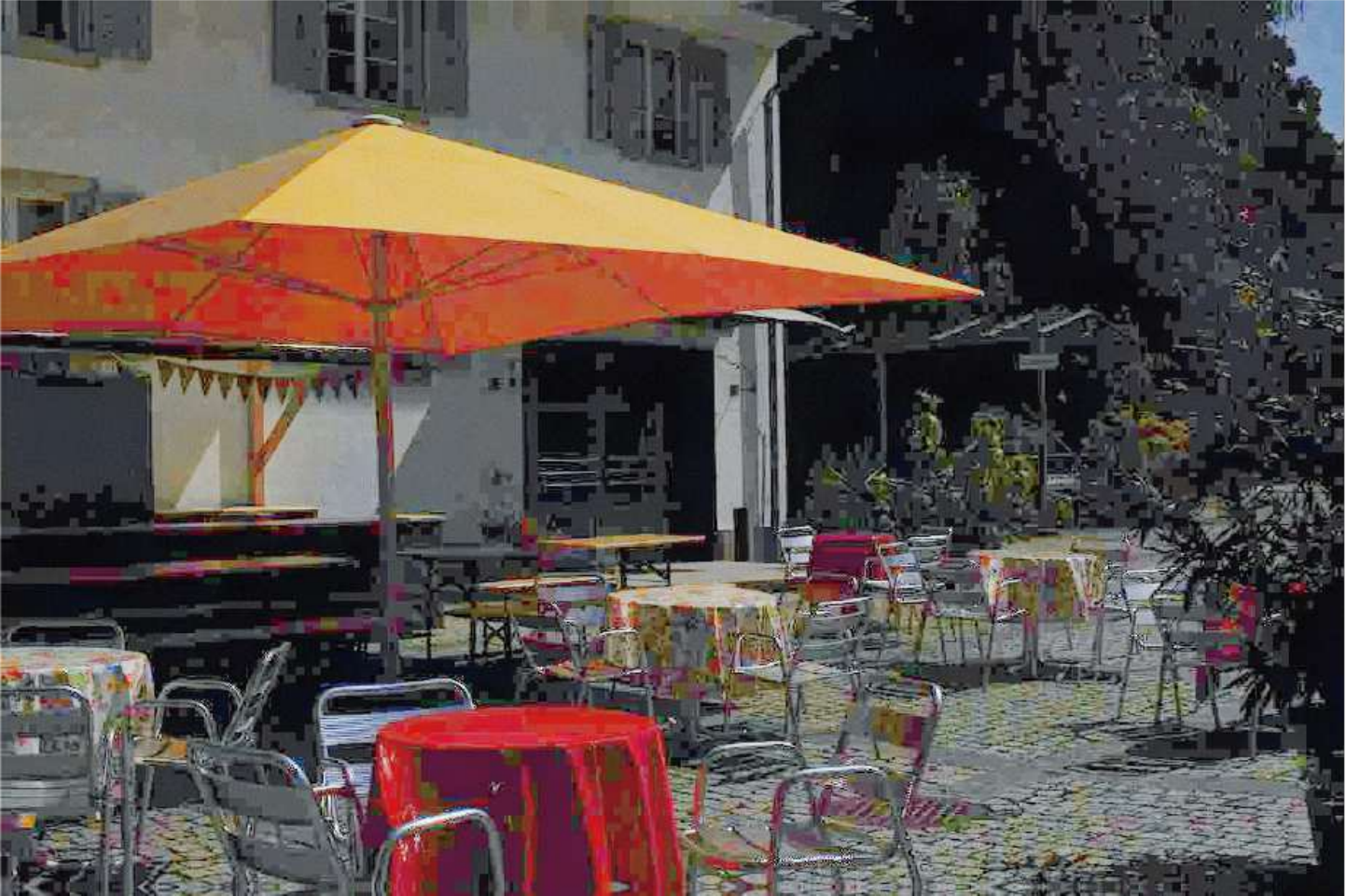}}
    \subfloat{\includegraphics[width = .10 \linewidth]{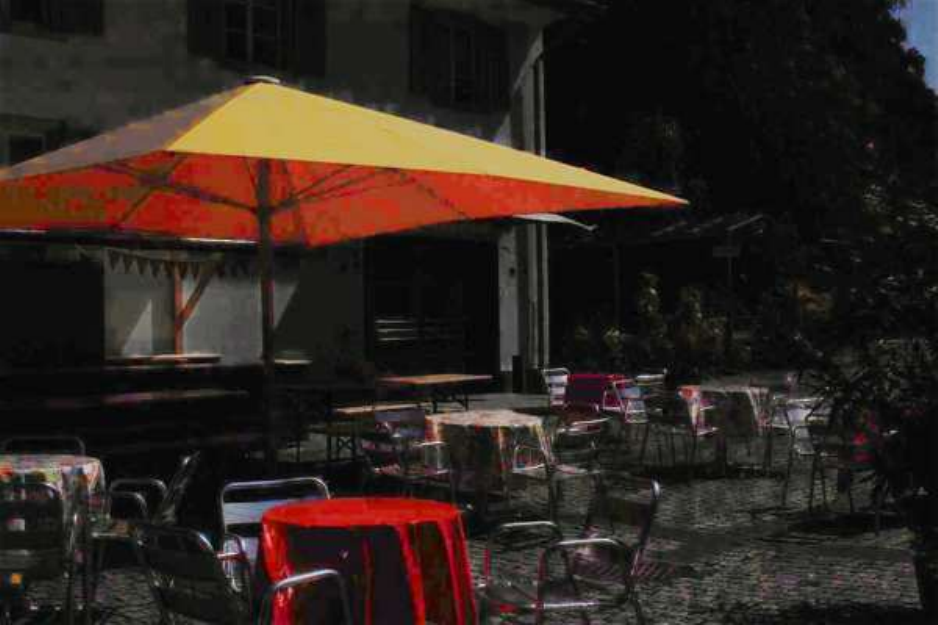}}
    \subfloat{\includegraphics[width = .10 \linewidth]{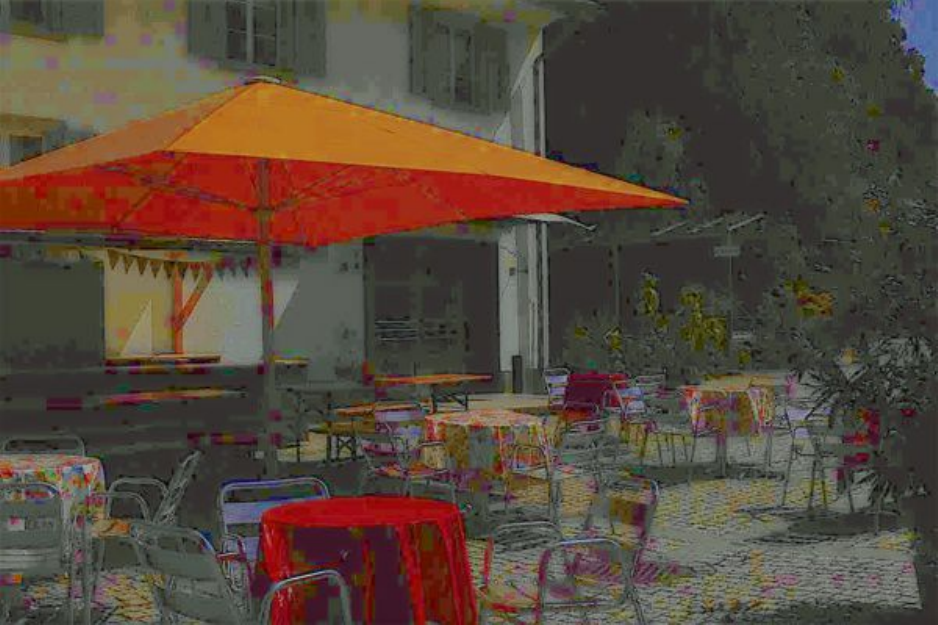}}
    \subfloat{\includegraphics[width = .10 \linewidth]{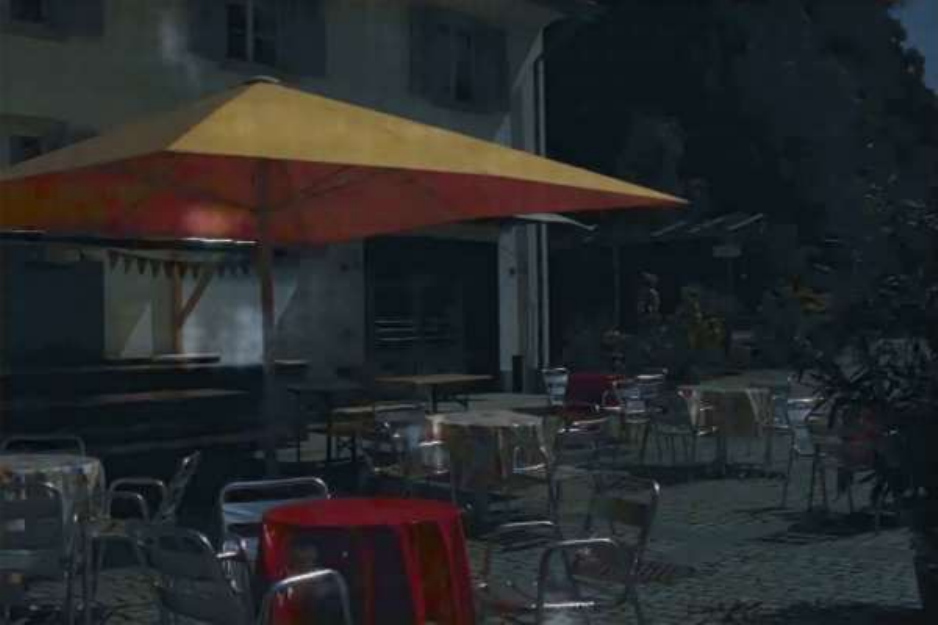}}
    \subfloat{\includegraphics[width = .10 \linewidth]{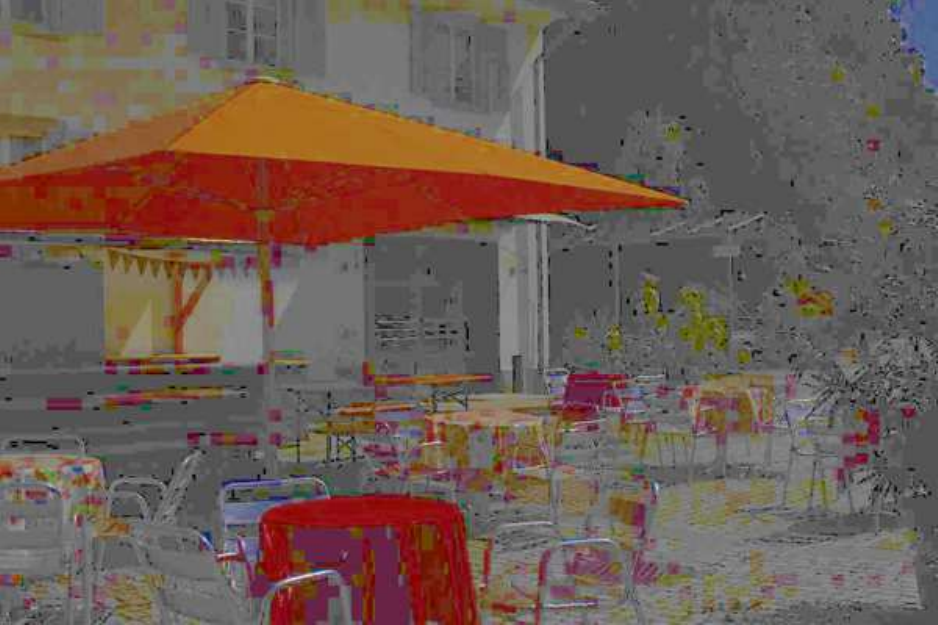}}
    \subfloat{\includegraphics[width = .10 \linewidth]{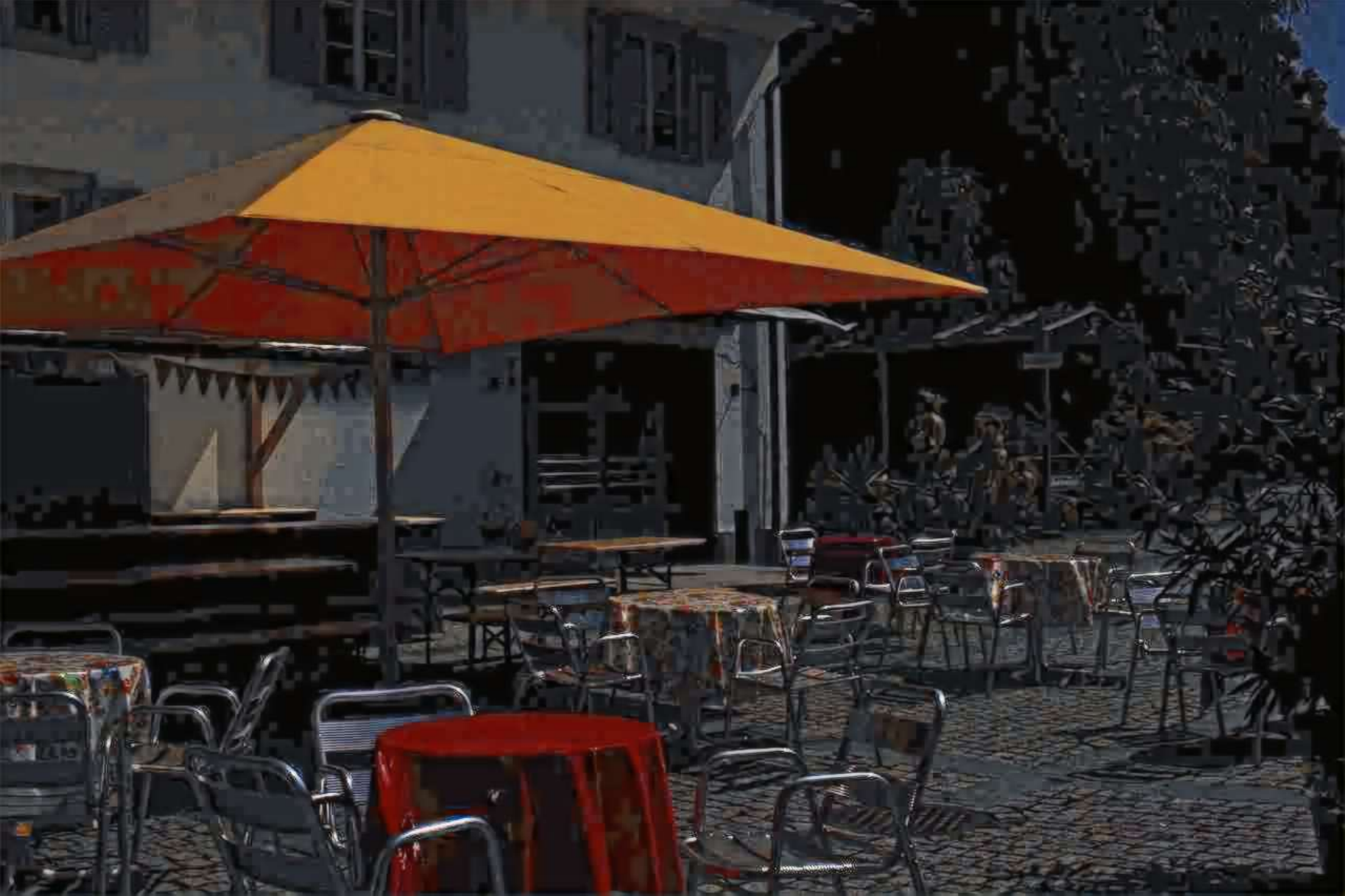}}
    \subfloat{\includegraphics[width = .10 \linewidth]{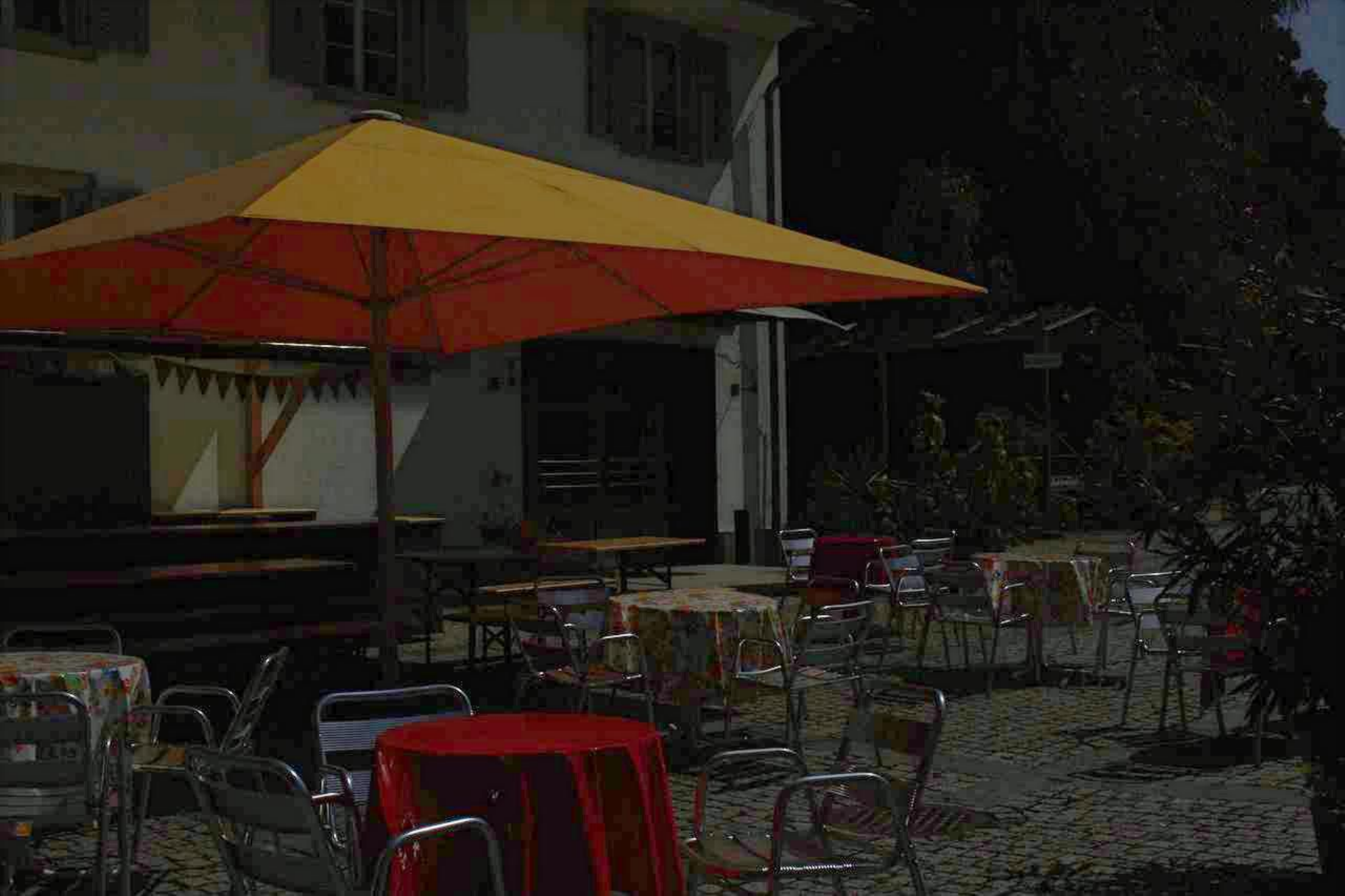}}\\\vspace{-0.16in}
    \subfloat{\includegraphics[width = .10 \linewidth]{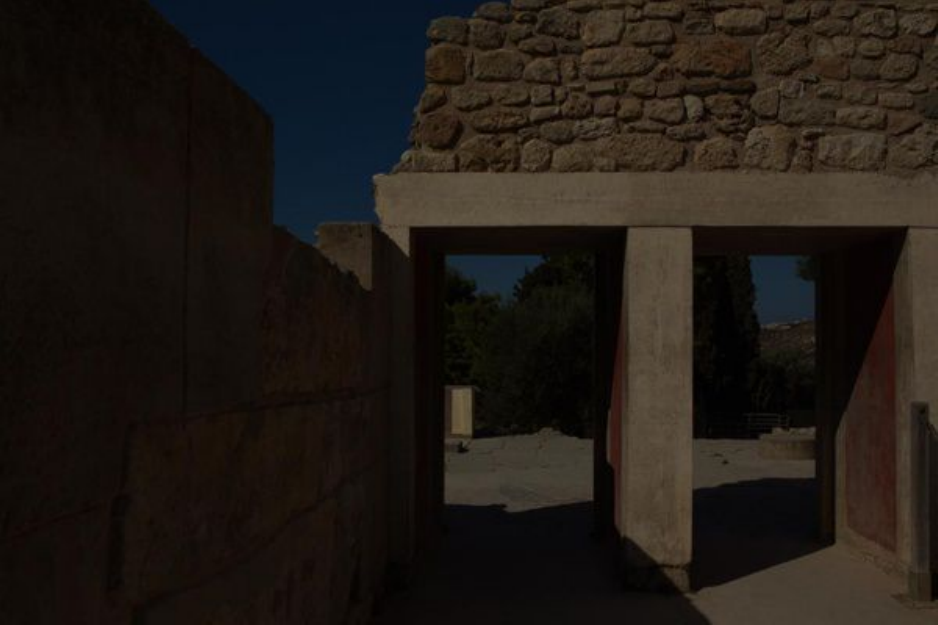}}
    \subfloat{\includegraphics[width = .10 \linewidth]{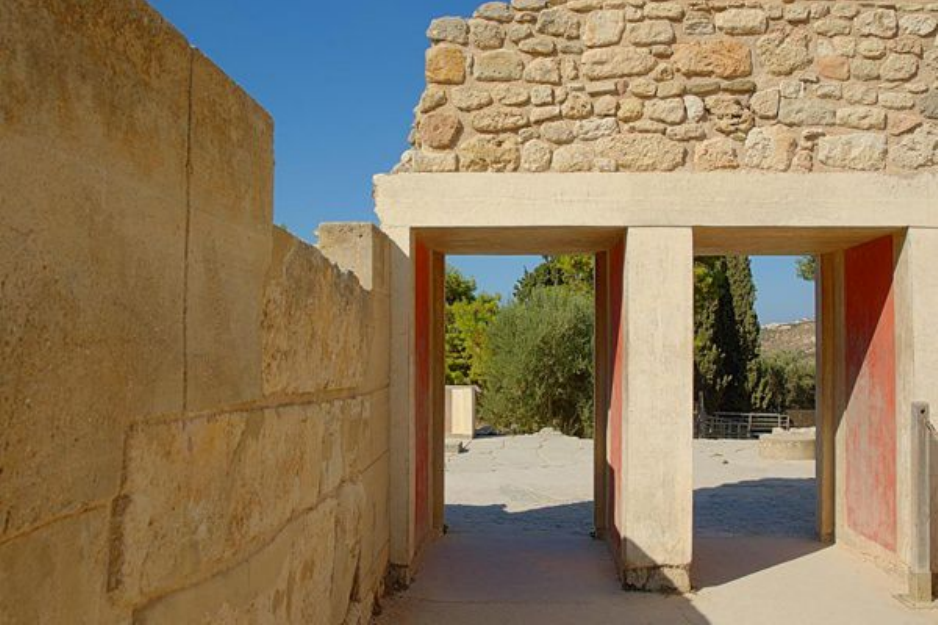}}
    \subfloat{\includegraphics[width = .10 \linewidth]{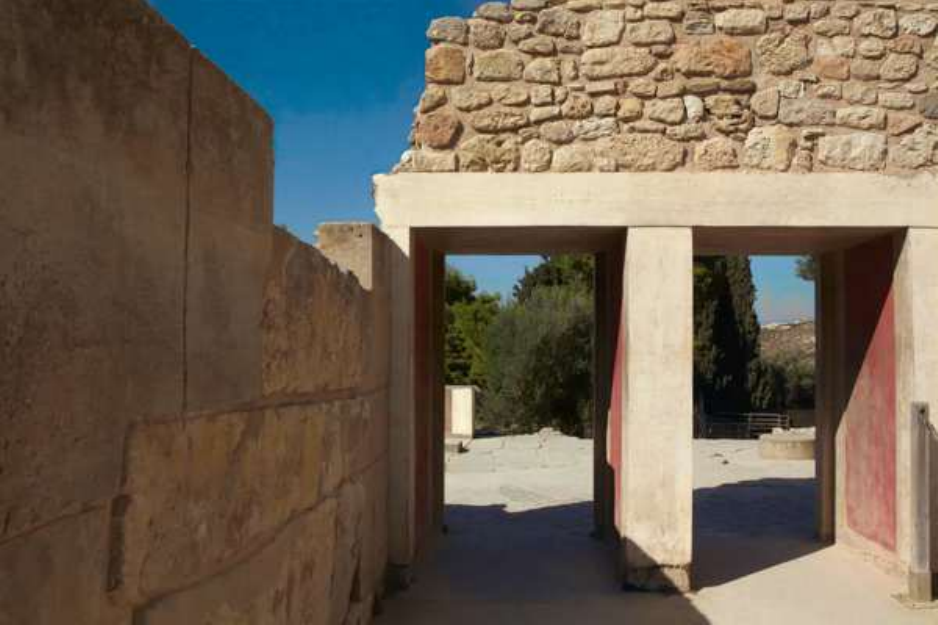}}
    \subfloat{\includegraphics[width = .10 \linewidth]{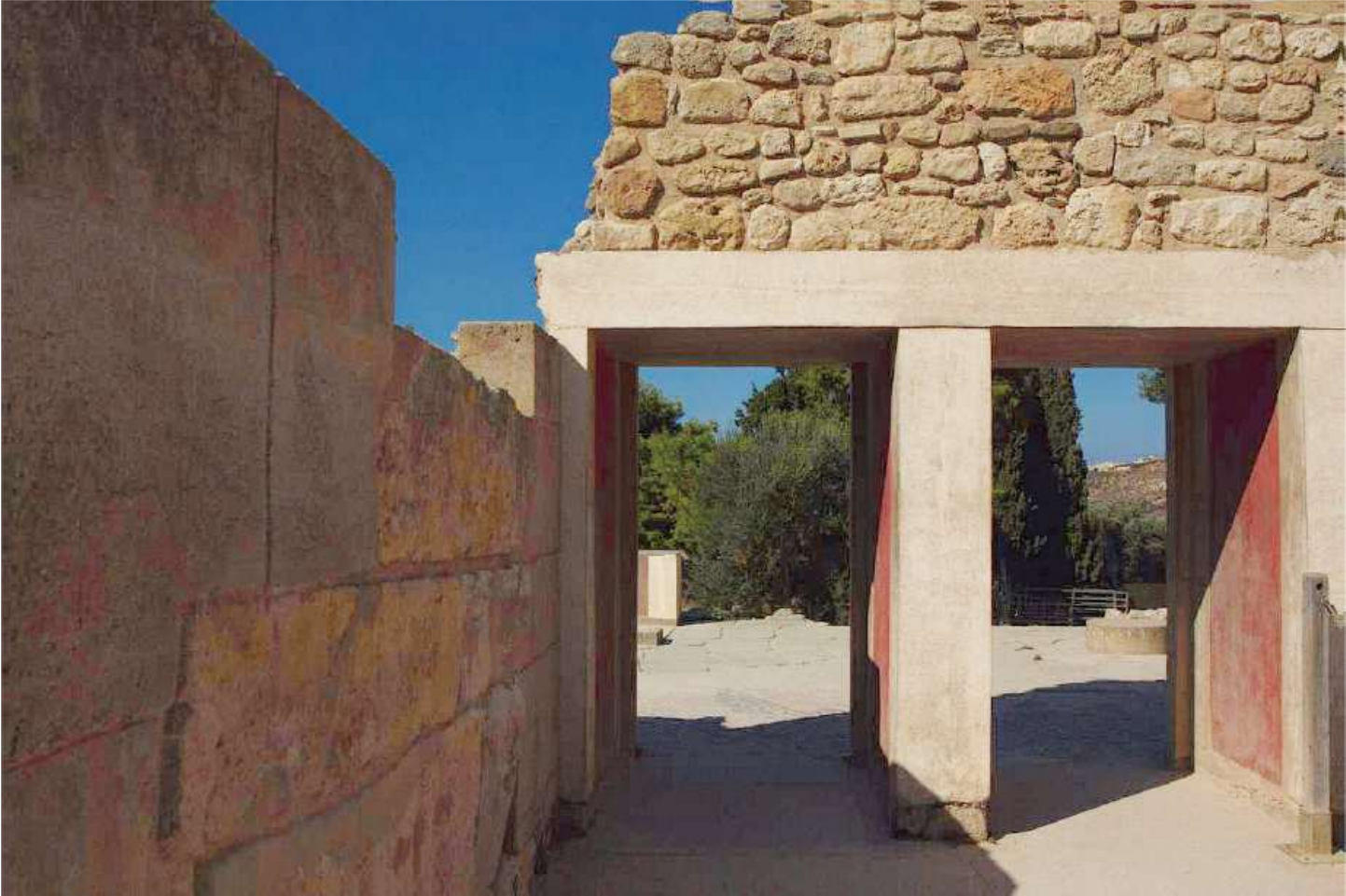}}
    \subfloat{\includegraphics[width = .10 \linewidth]{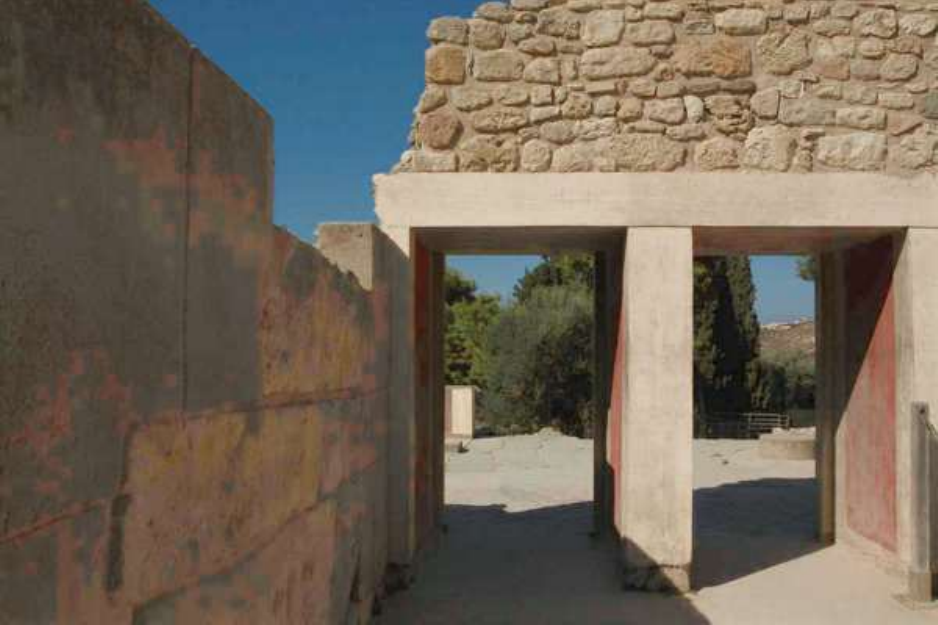}}
    \subfloat{\includegraphics[width = .10 \linewidth]{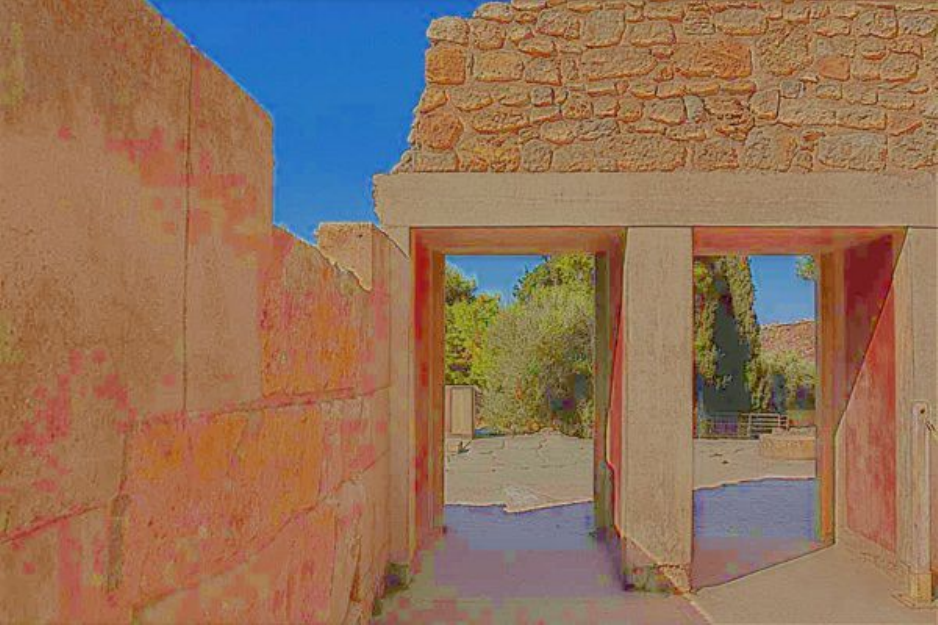}}
    \subfloat{\includegraphics[width = .10 \linewidth]{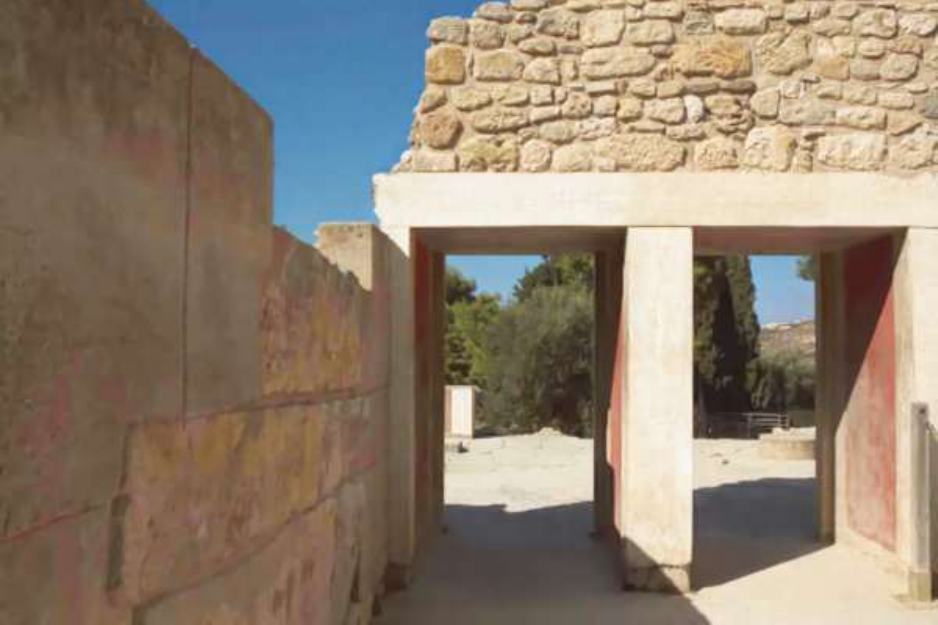}}
    \subfloat{\includegraphics[width = .10 \linewidth]{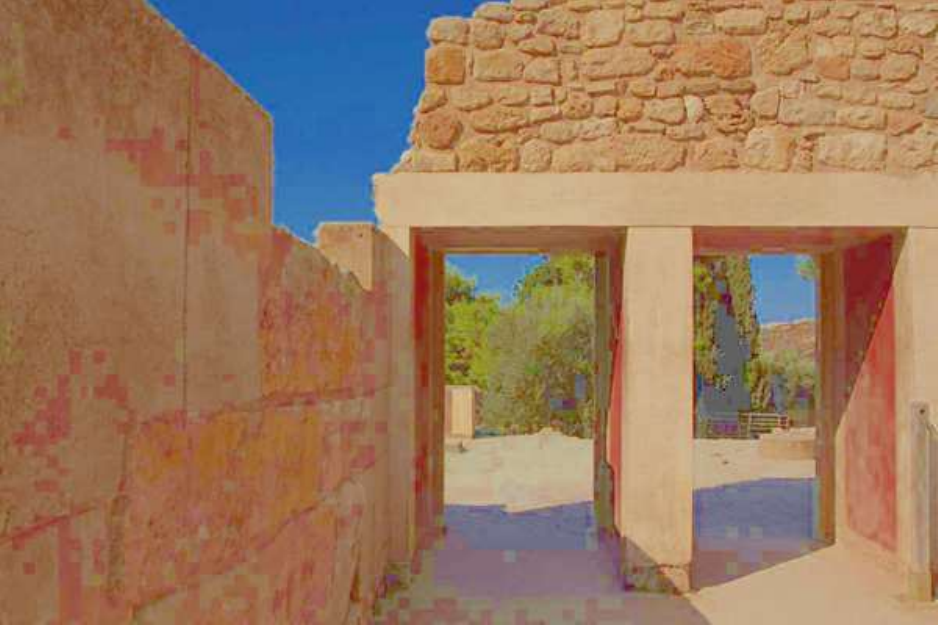}}
    \subfloat{\includegraphics[width = .10 \linewidth]{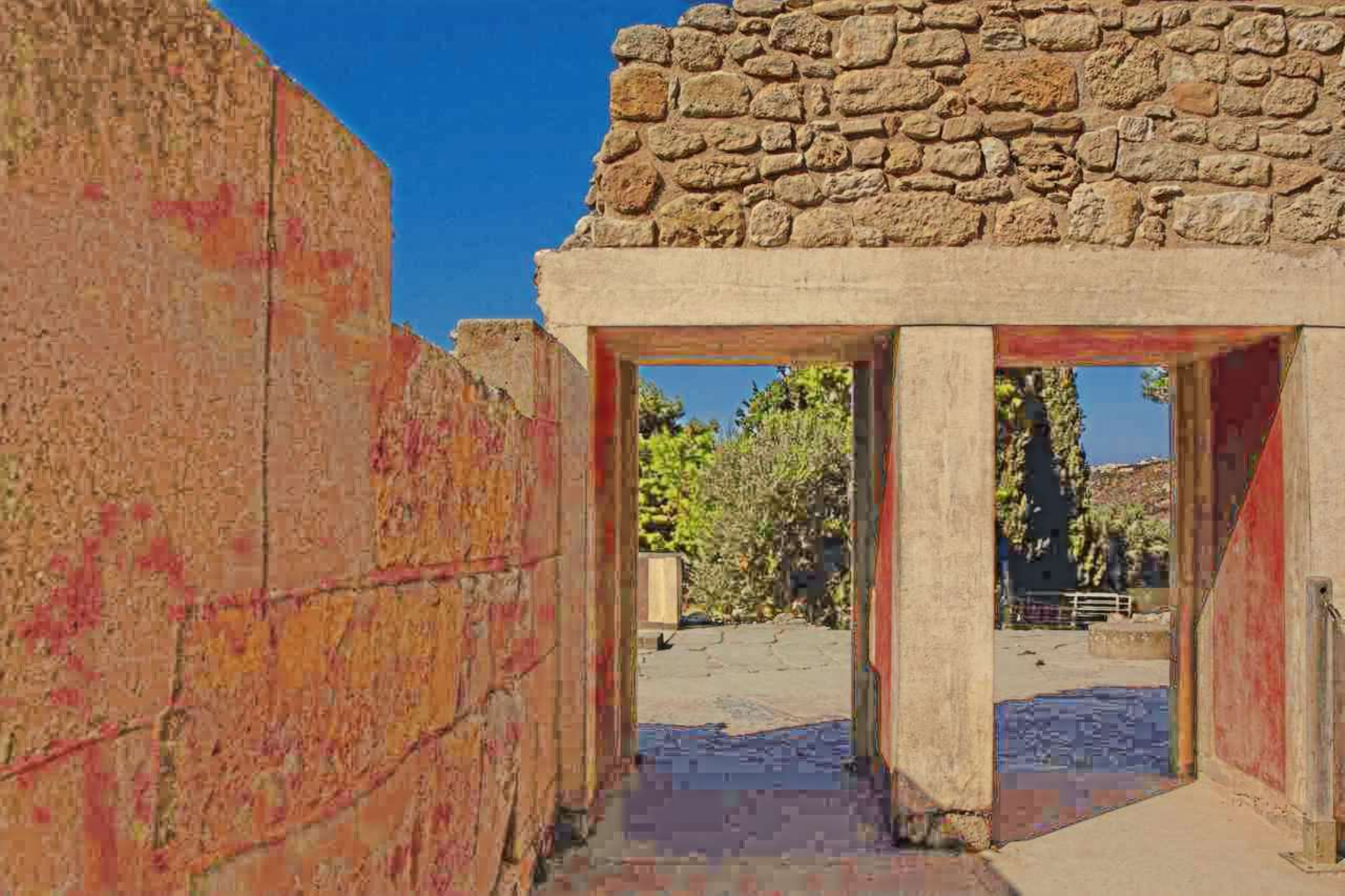}}
    \subfloat{\includegraphics[width = .10 \linewidth]{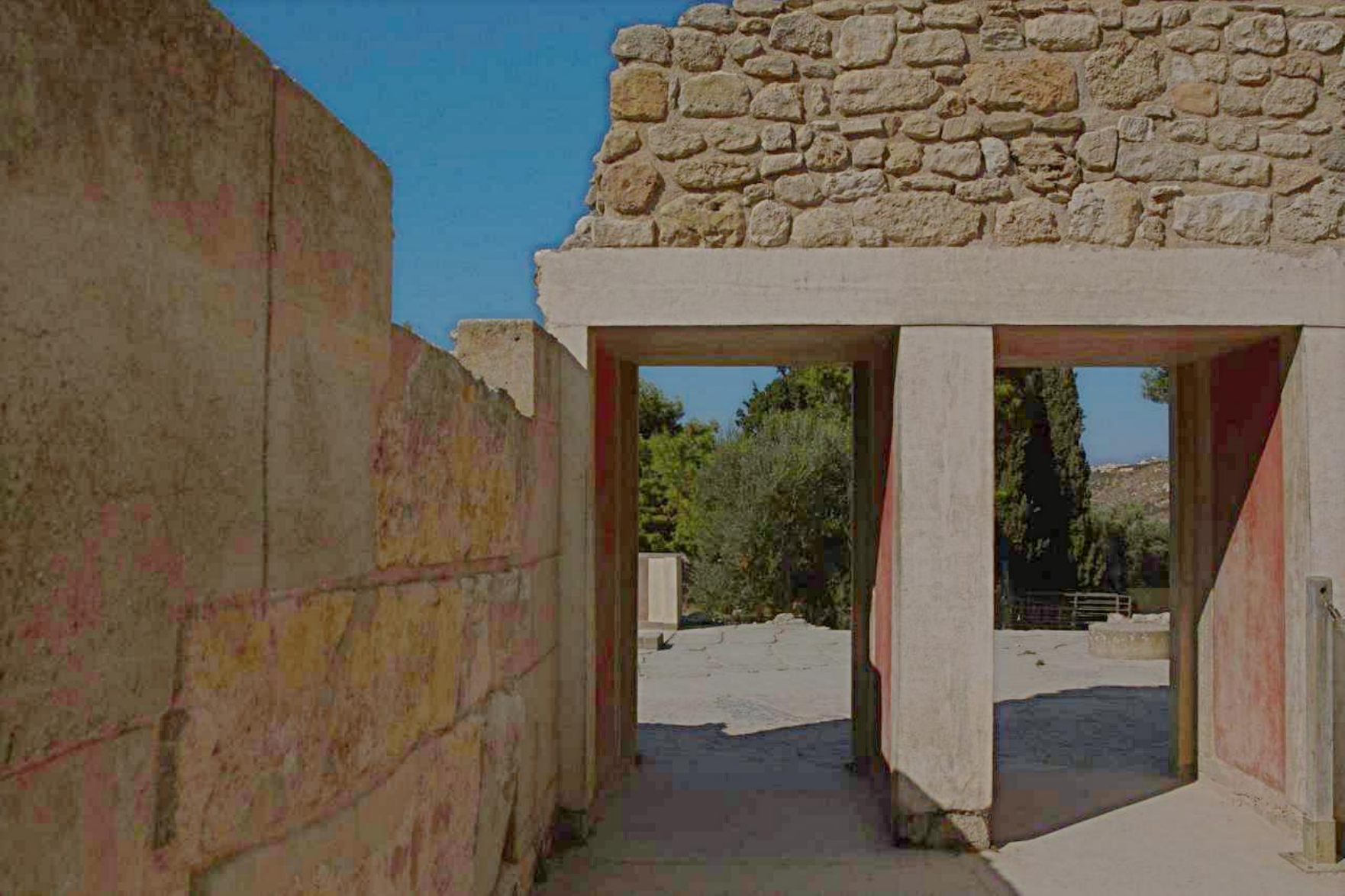}}\\\vspace{-0.1in}
    \subfloat{\includegraphics[width = .10 \linewidth]{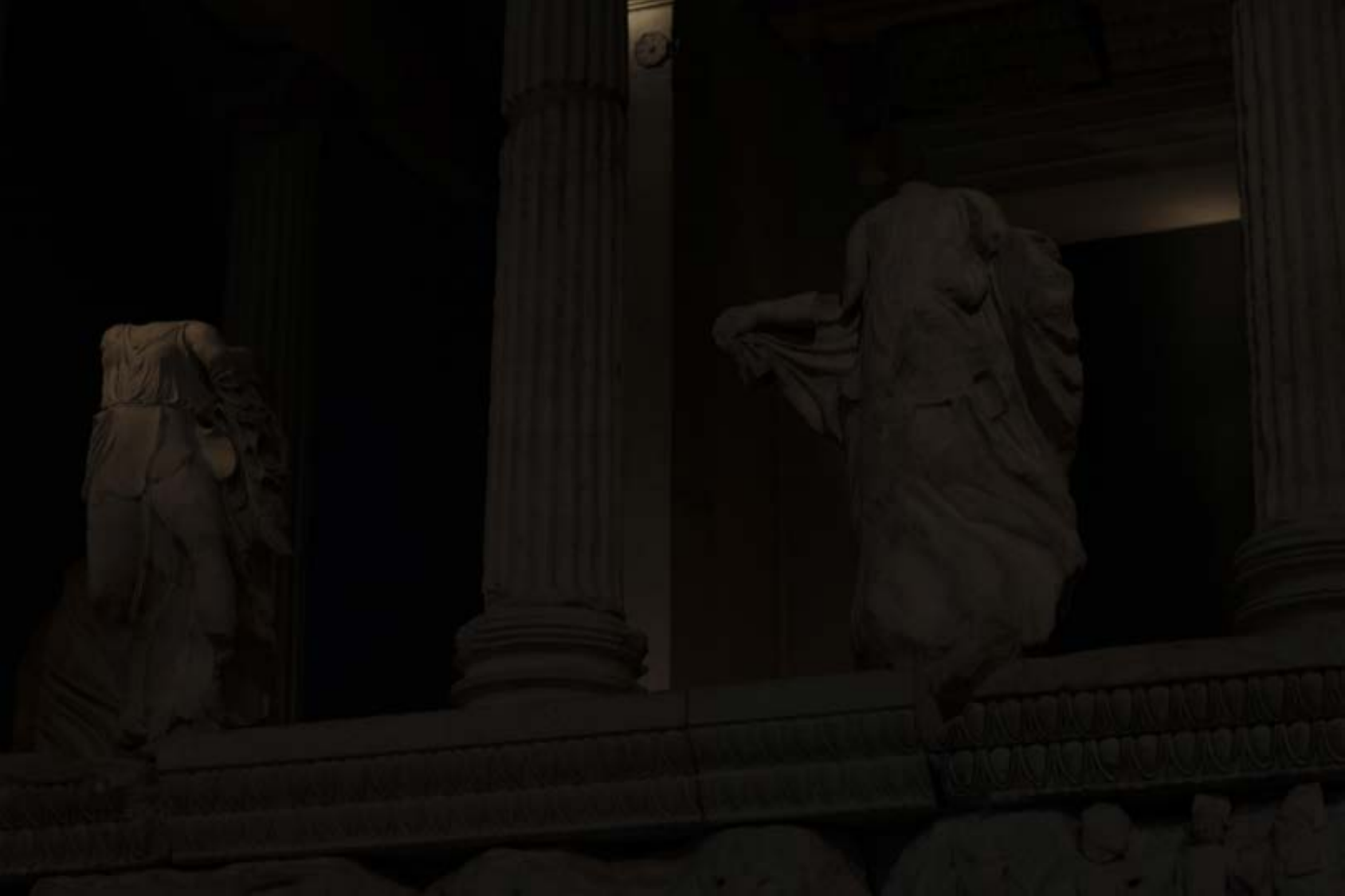}}
    \subfloat{\includegraphics[width = .10 \linewidth]{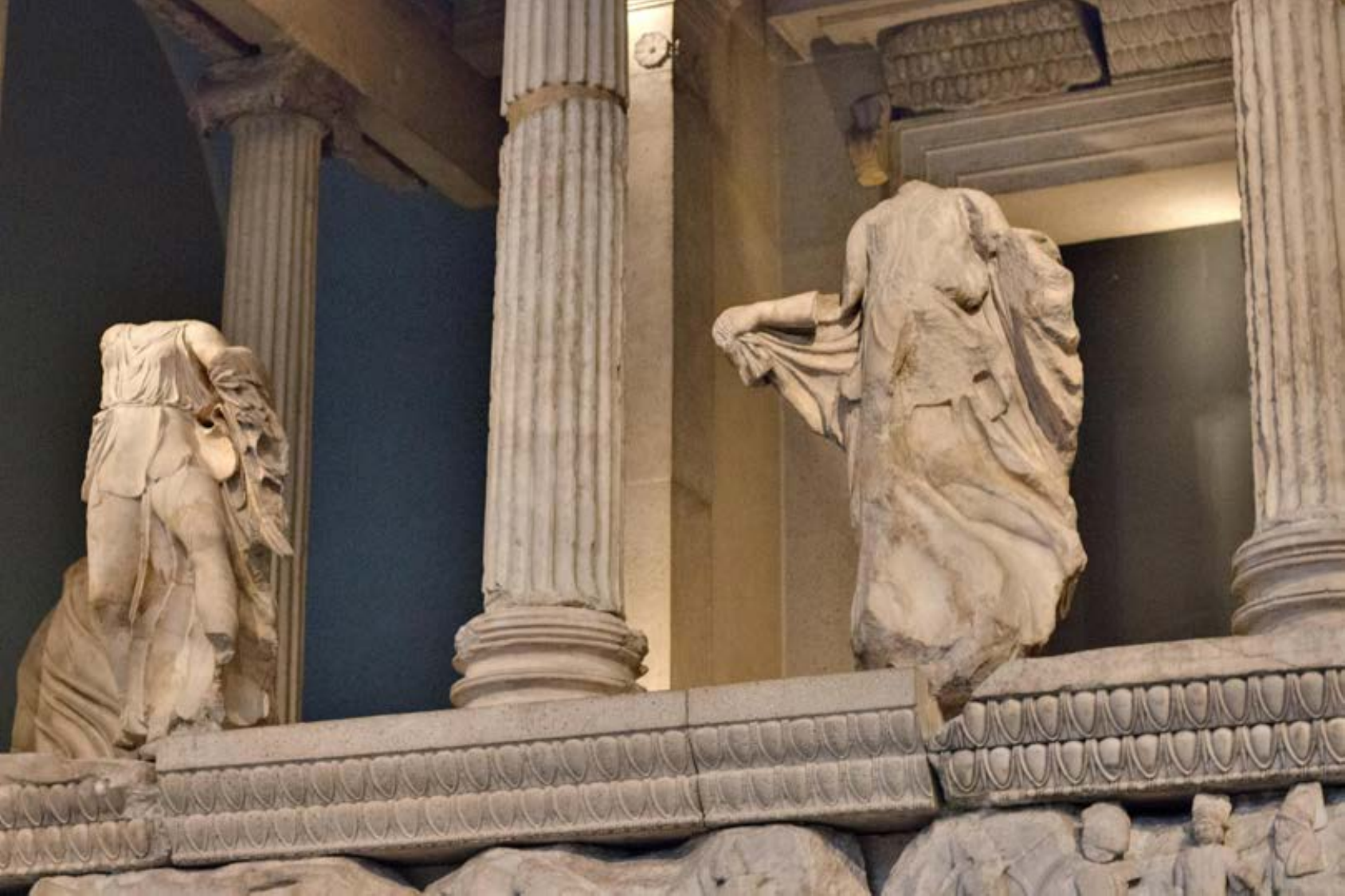}}
    \subfloat{\includegraphics[width = .10 \linewidth]{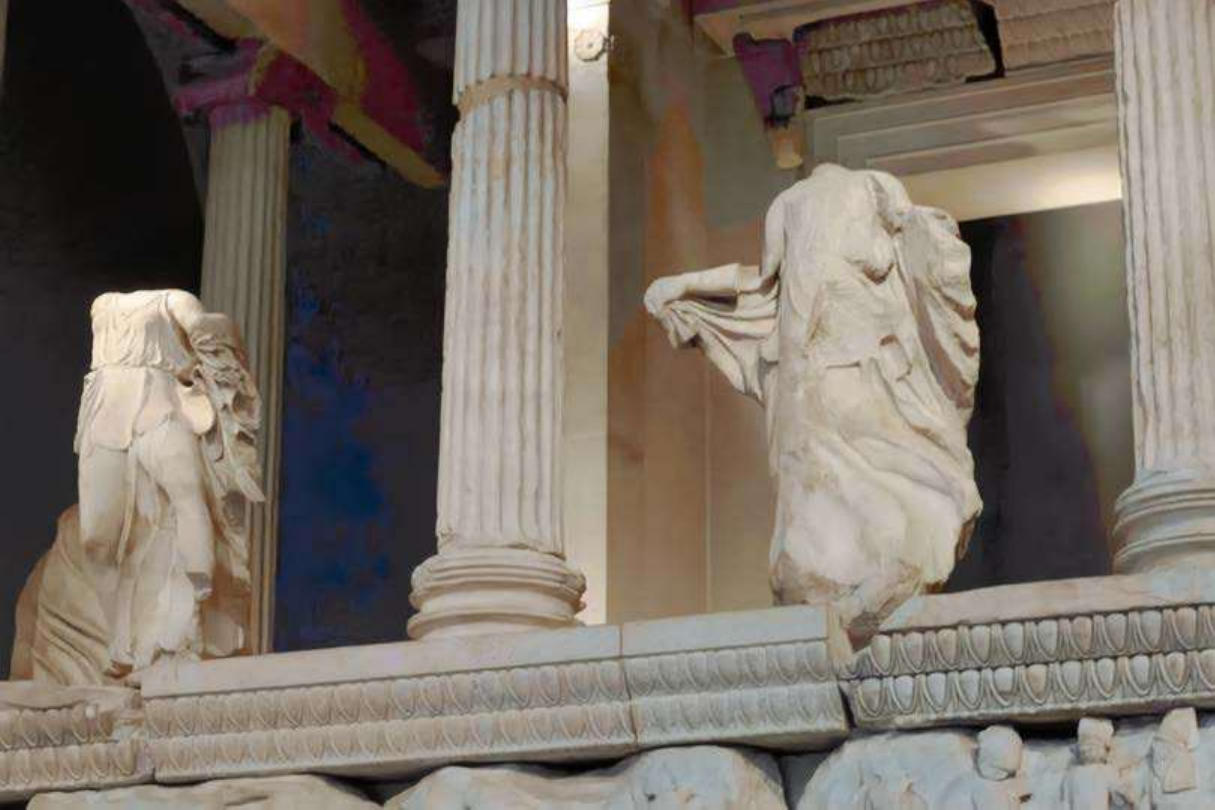}}
    \subfloat{\includegraphics[width = .10 \linewidth]{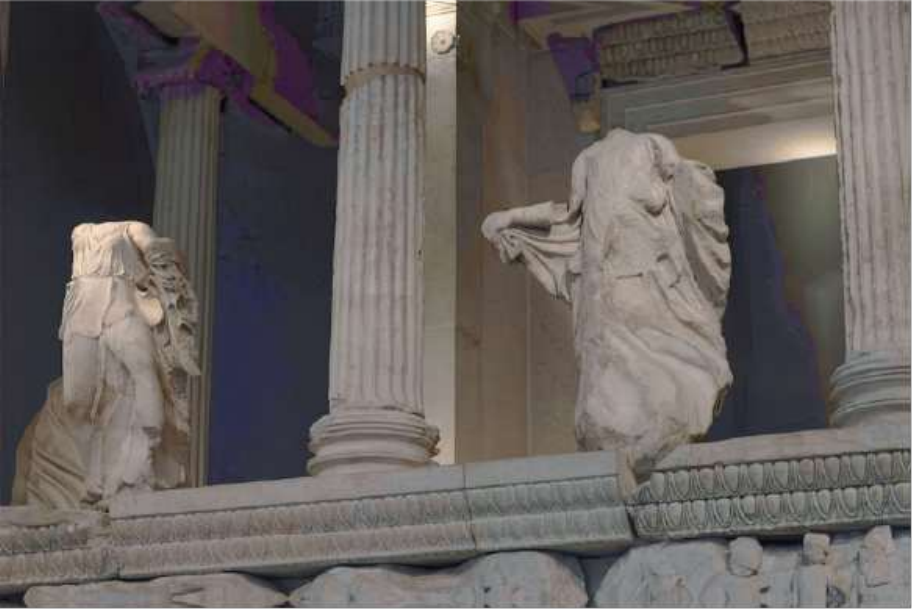}}
    \subfloat{\includegraphics[width = .10 \linewidth]{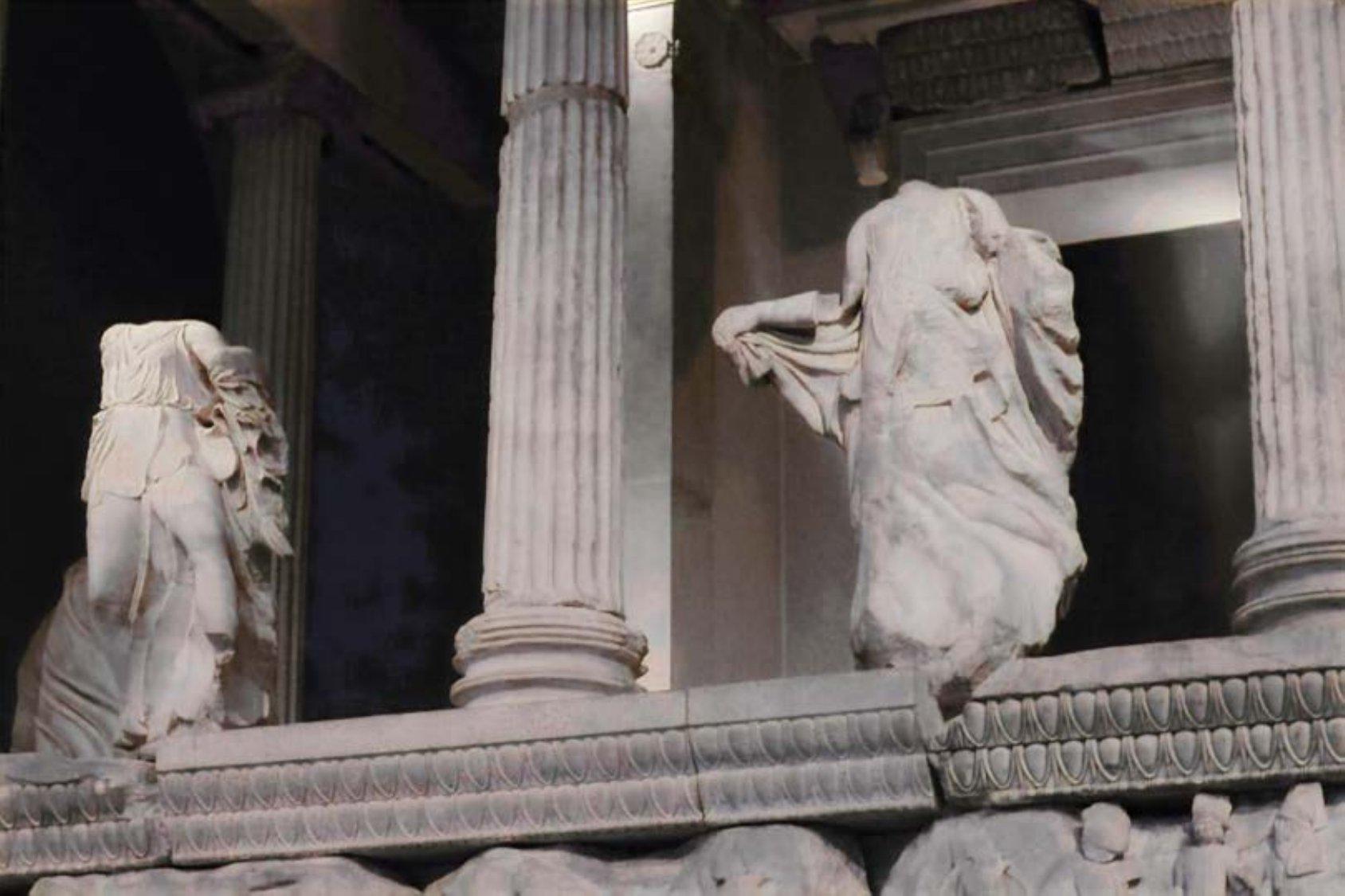}}
    \subfloat{\includegraphics[width = .10 \linewidth]{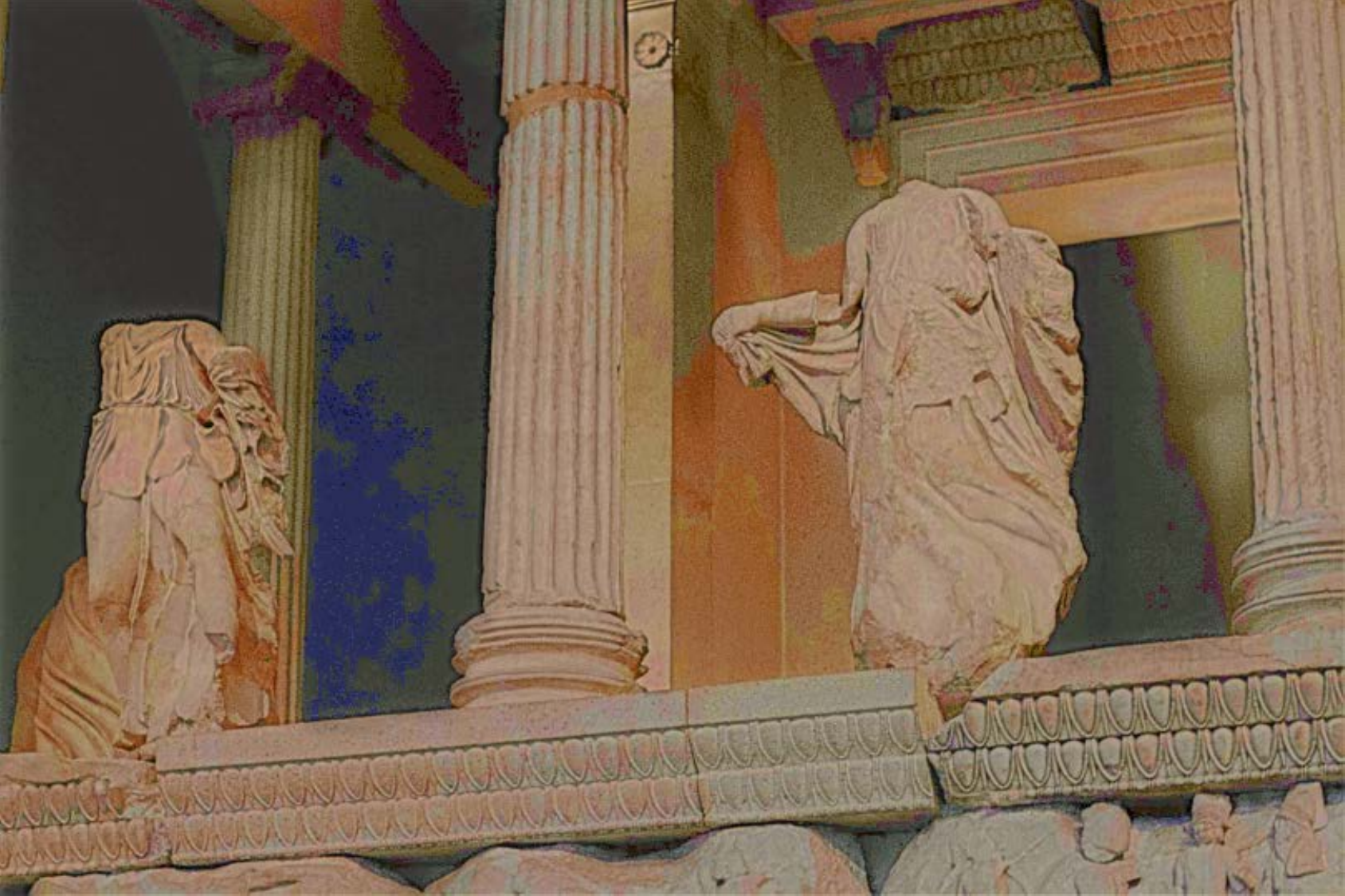}}
    \subfloat{\includegraphics[width = .10 \linewidth]{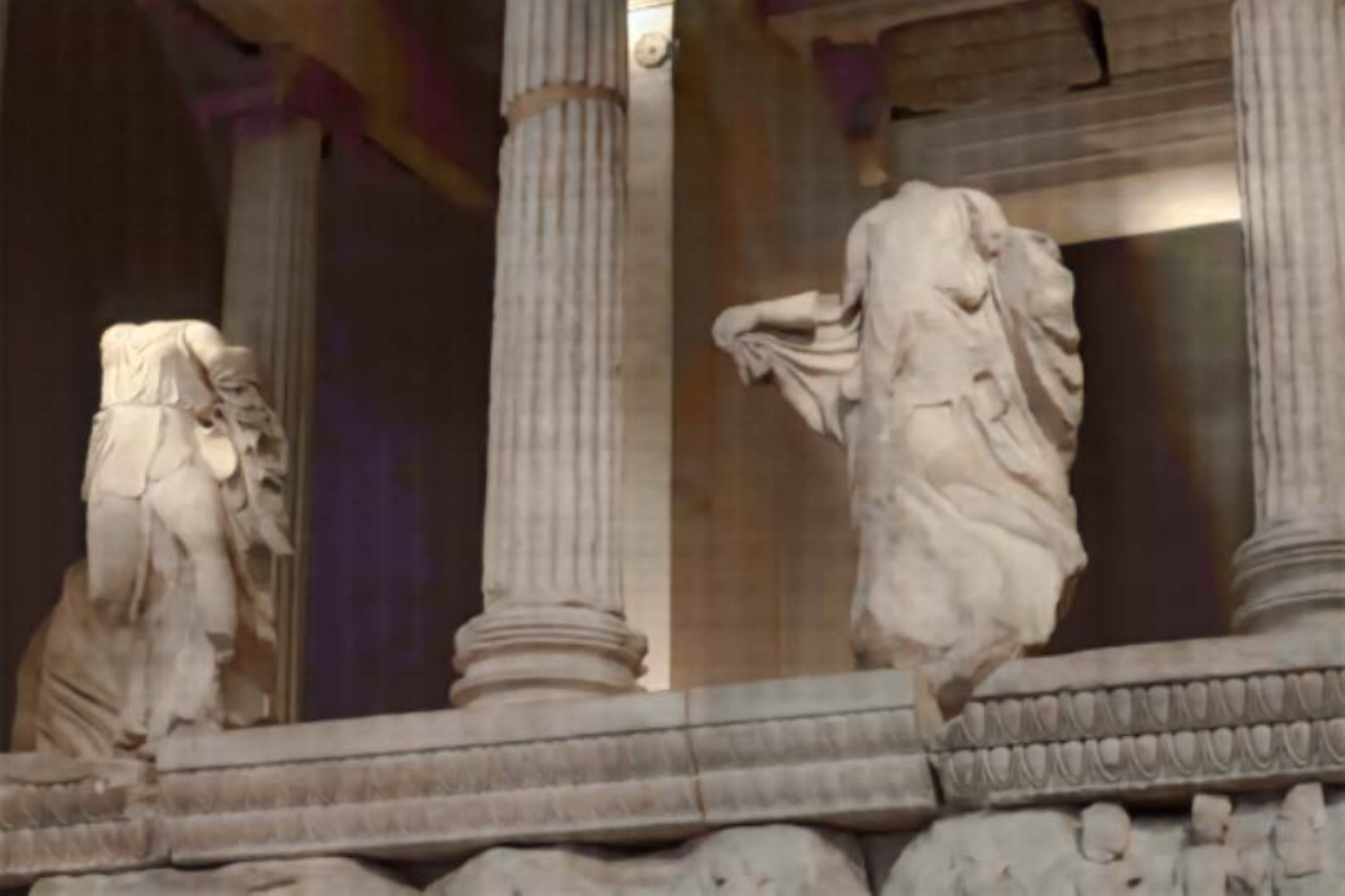}}
    \subfloat{\includegraphics[width = .10 \linewidth]{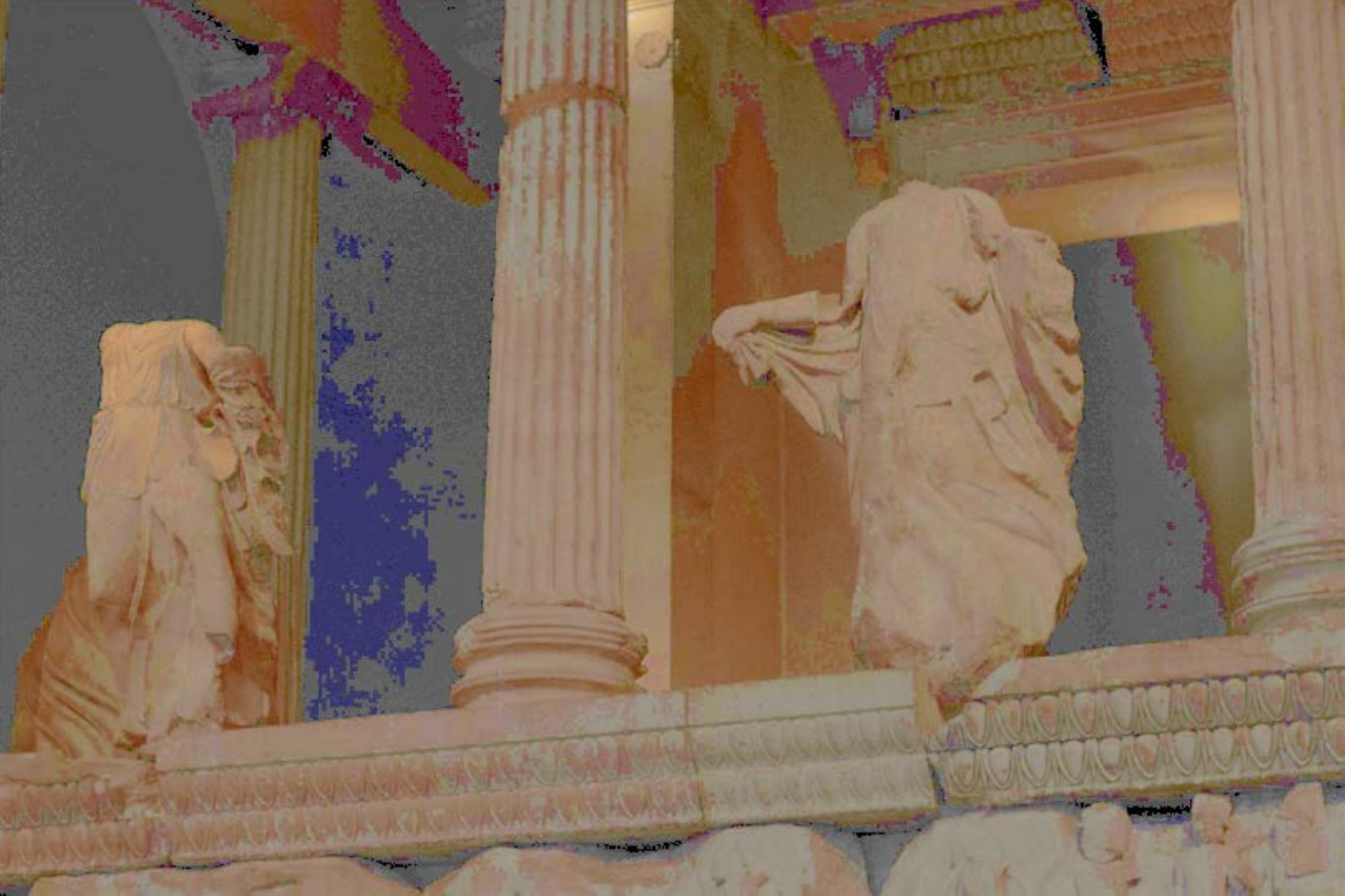}}
    \subfloat{\includegraphics[width = .10 \linewidth]{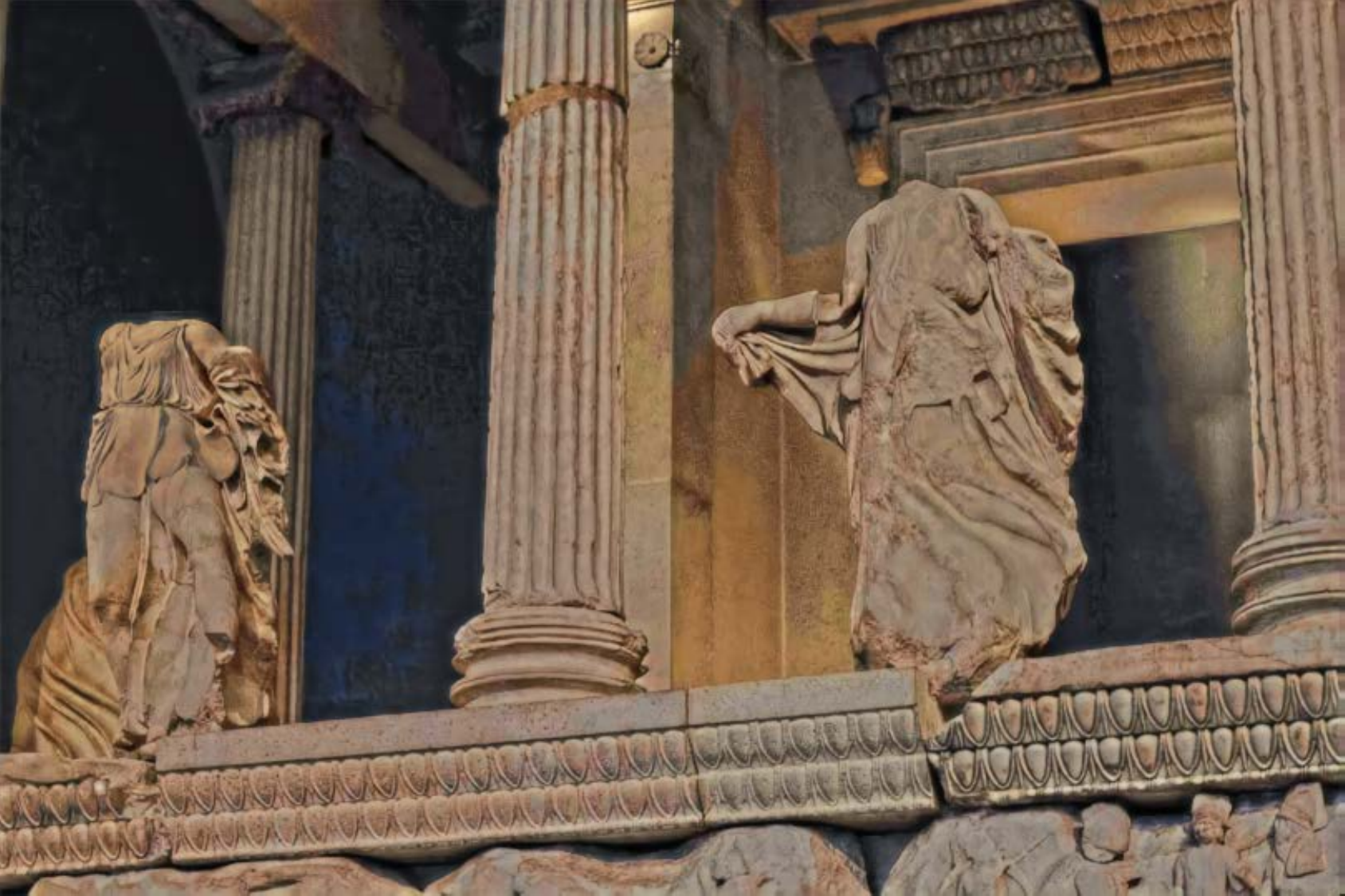}}
    \subfloat{\includegraphics[width = .10 \linewidth]{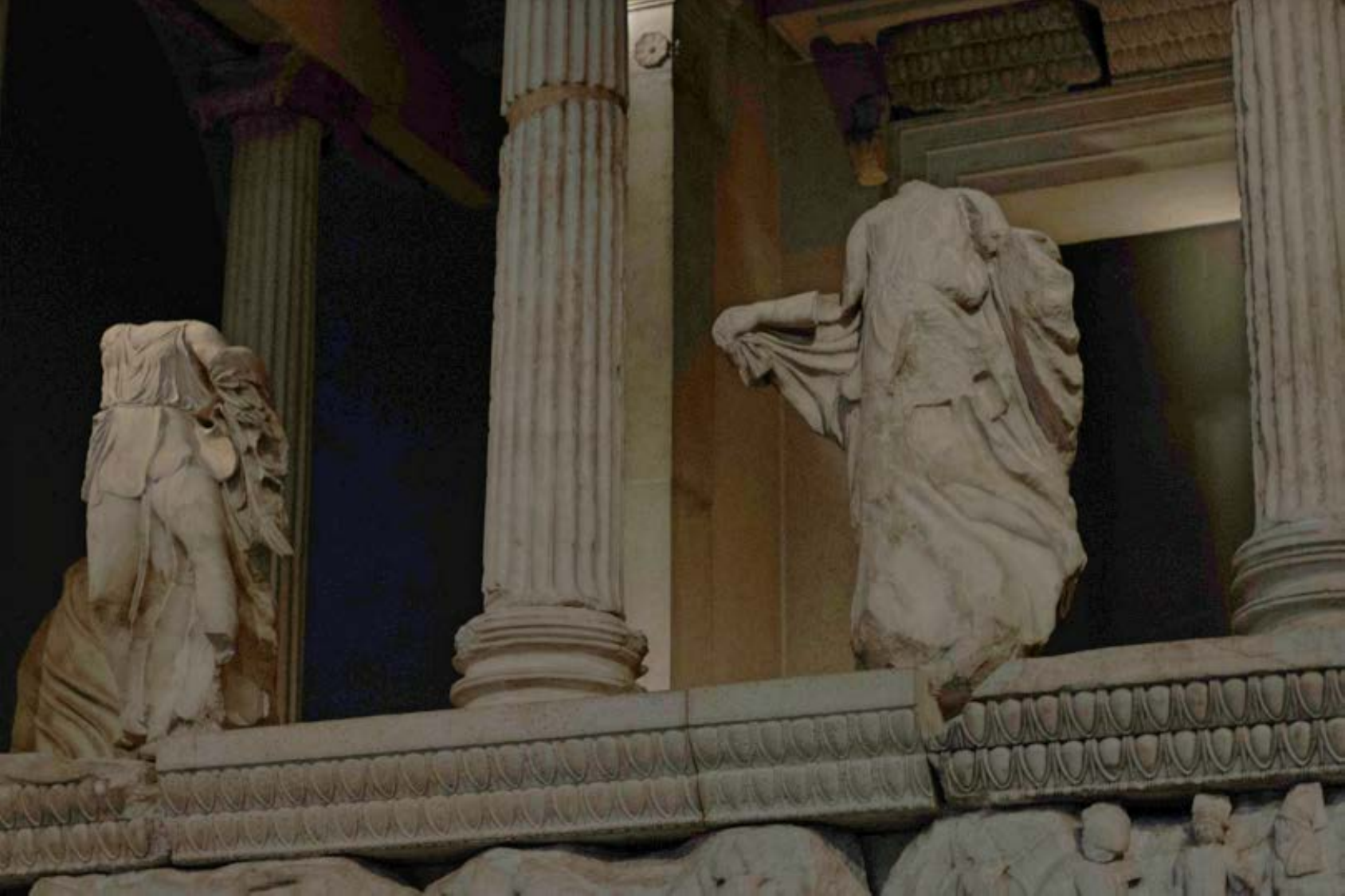}}\\\vspace{-0.16in}
    \subfloat{\includegraphics[width = .10 \linewidth]{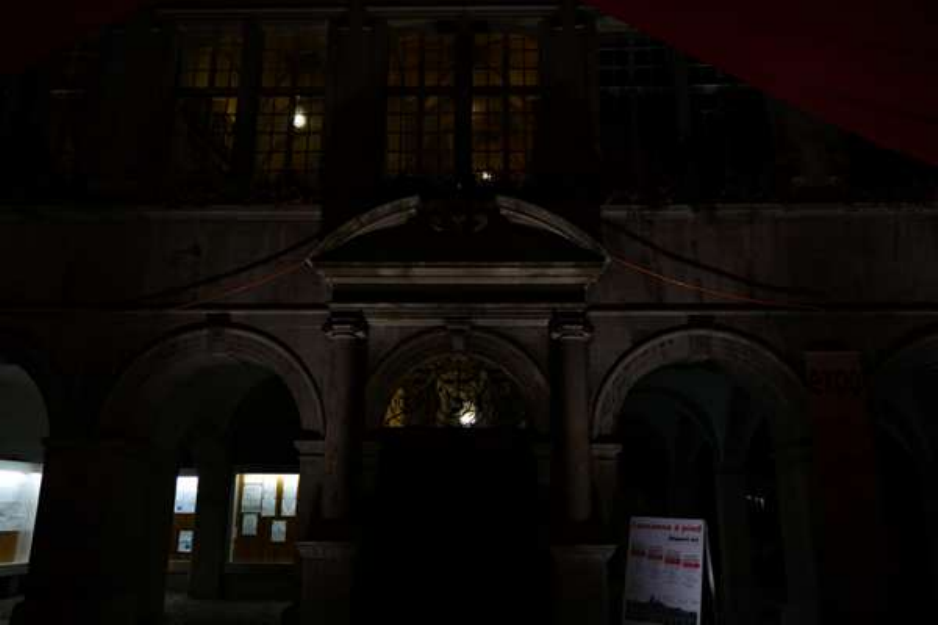}}
    \subfloat{\includegraphics[width = .10 \linewidth]{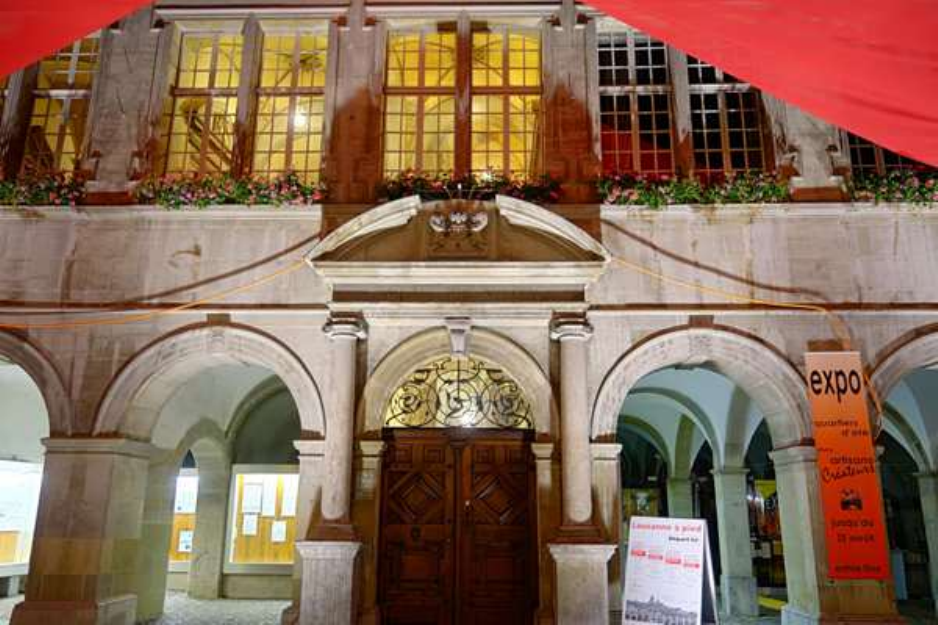}}
    \subfloat{\includegraphics[width = .10 \linewidth]{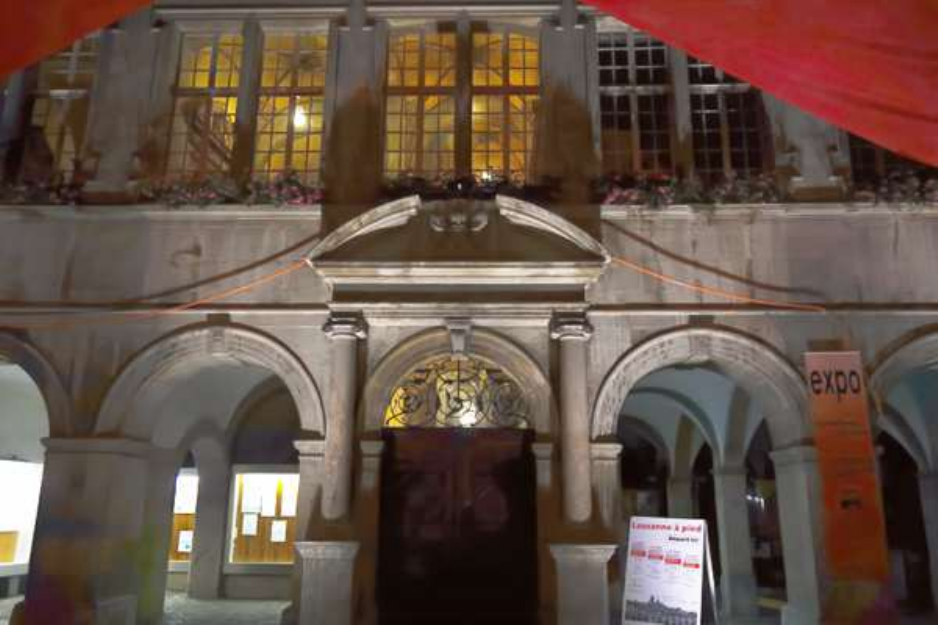}}
    \subfloat{\includegraphics[width = .10 \linewidth]{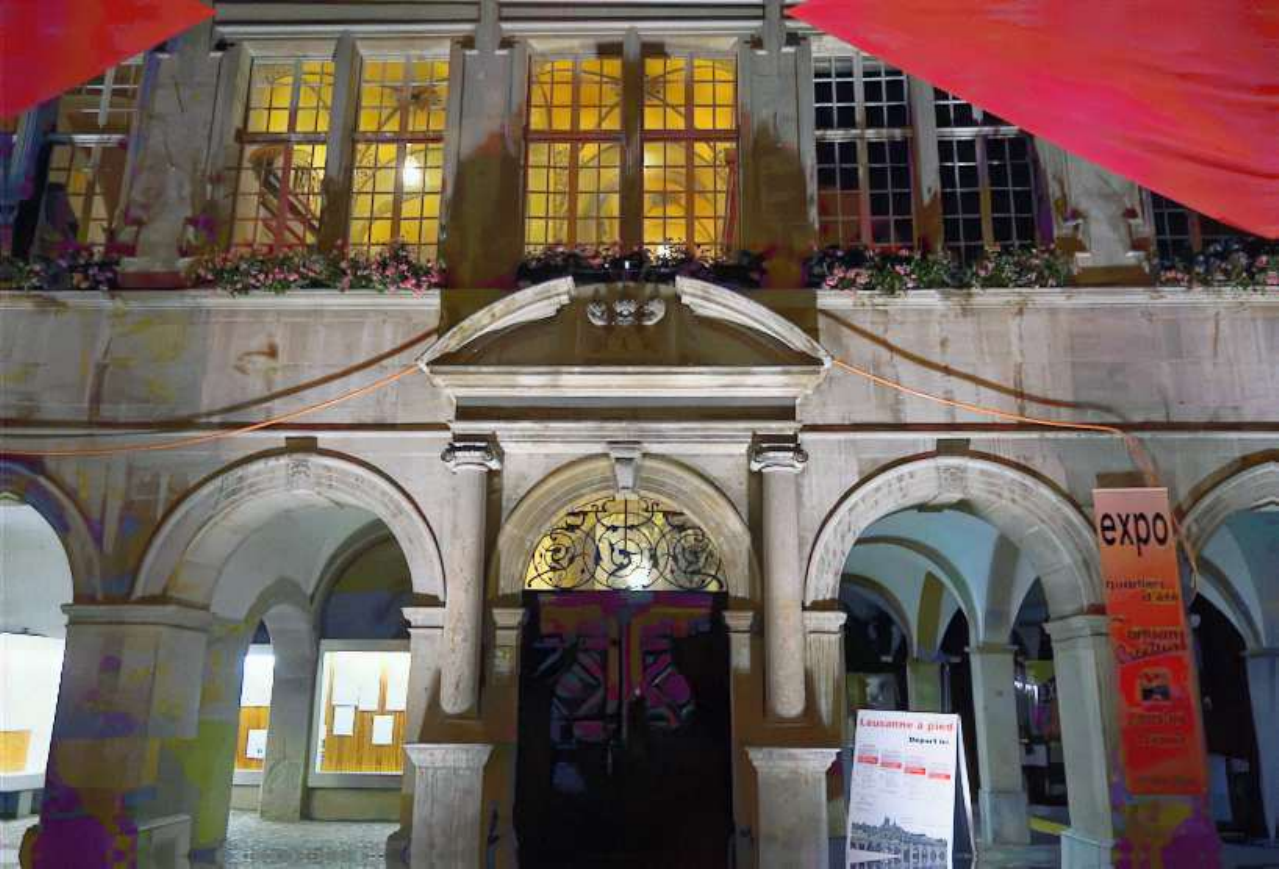}}
    \subfloat{\includegraphics[width = .10 \linewidth]{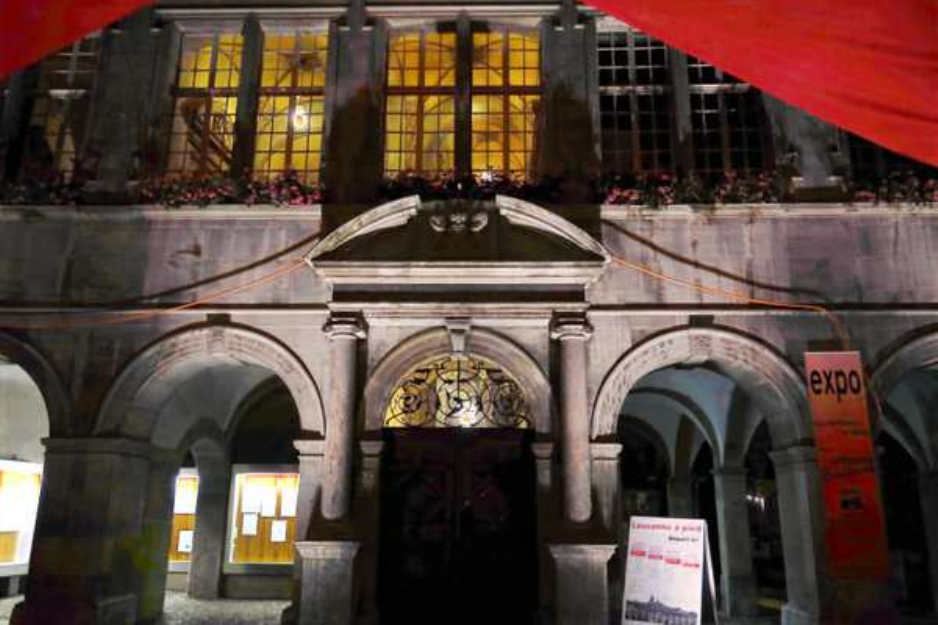}}
    \subfloat{\includegraphics[width = .10 \linewidth]{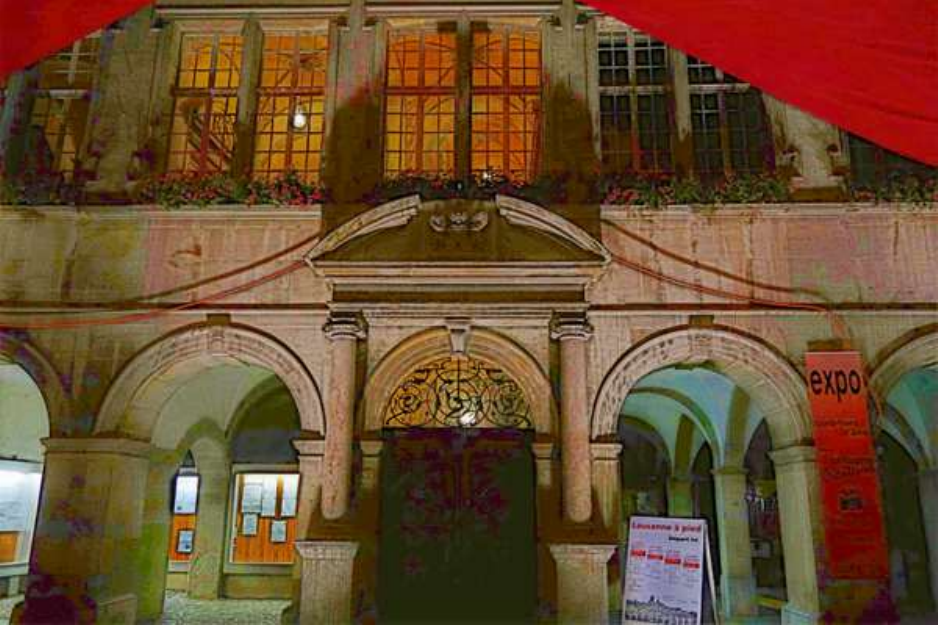}}
    \subfloat{\includegraphics[width = .10 \linewidth]{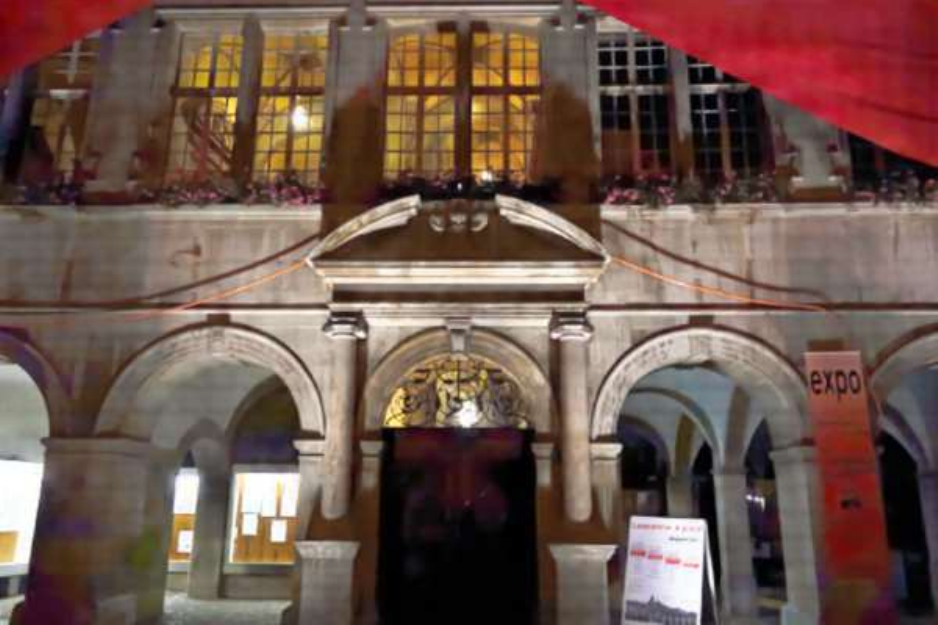}}
    \subfloat{\includegraphics[width = .10 \linewidth]{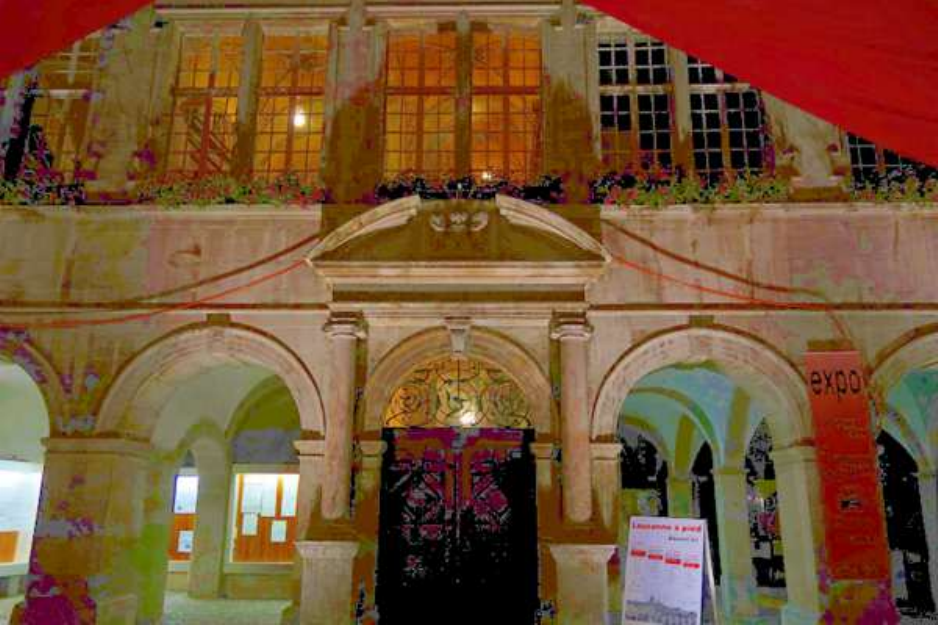}}
    \subfloat{\includegraphics[width = .10 \linewidth]{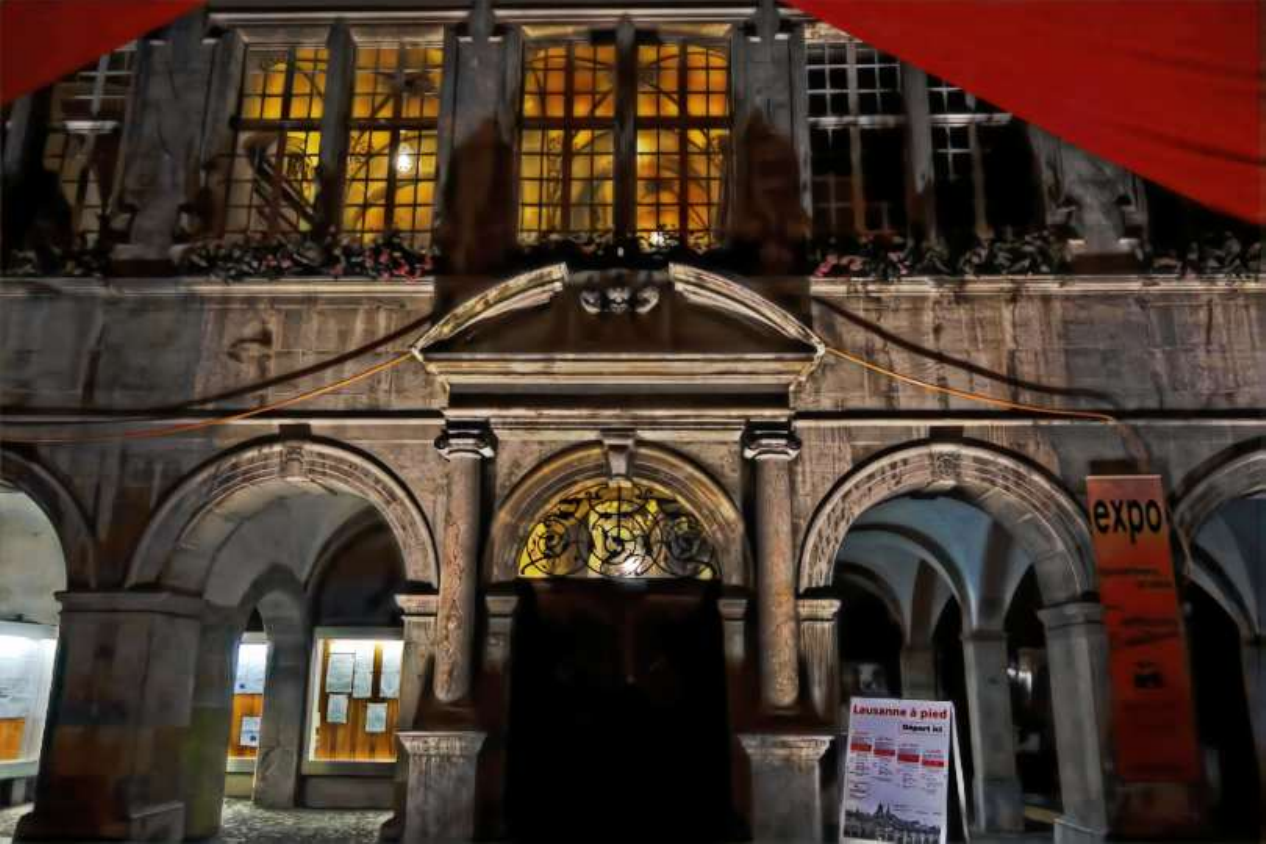}}
    \subfloat{\includegraphics[width = .10 \linewidth]{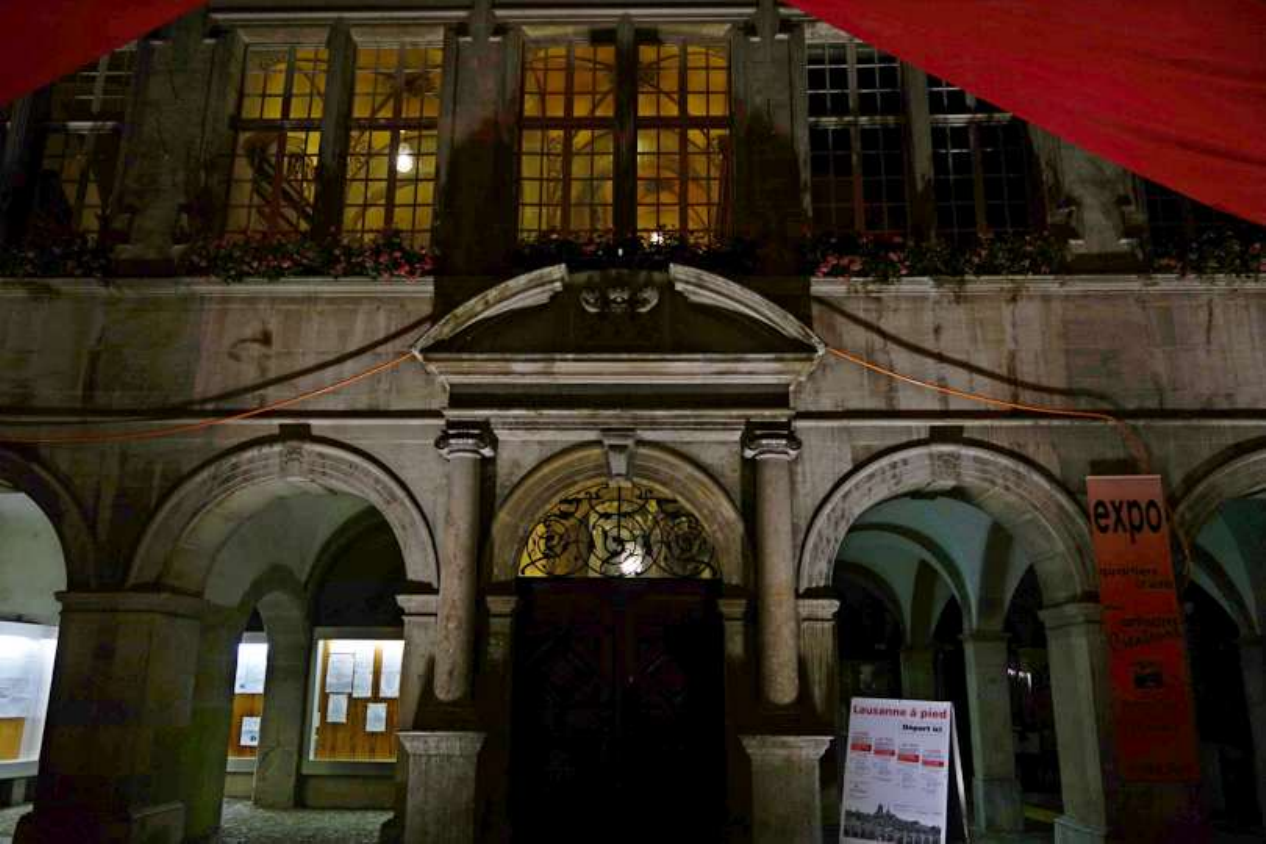}}\\\vspace{-0.16in}
    \subfloat{\includegraphics[width = .10 \linewidth]{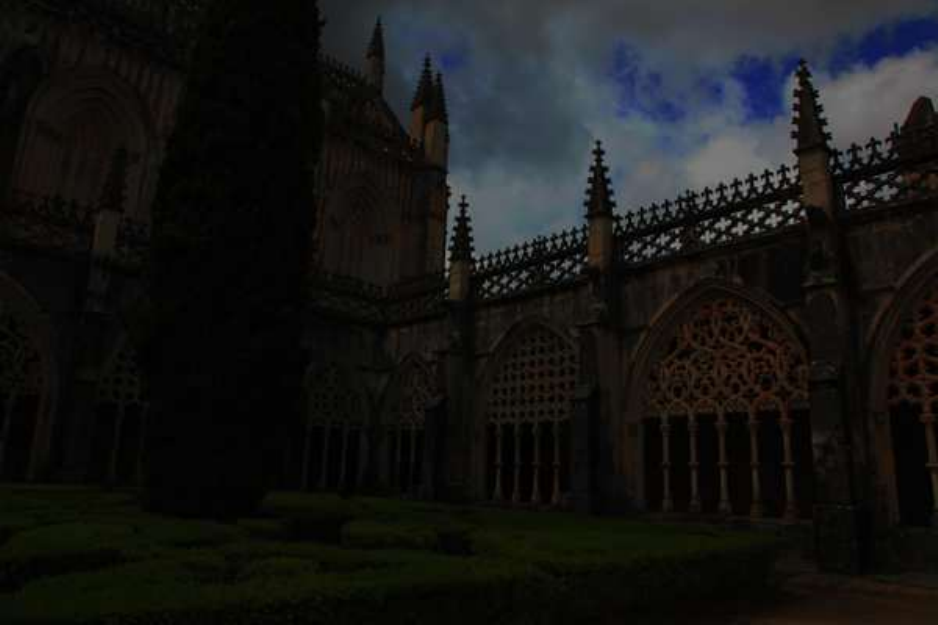}}
    \subfloat{\includegraphics[width = .10 \linewidth]{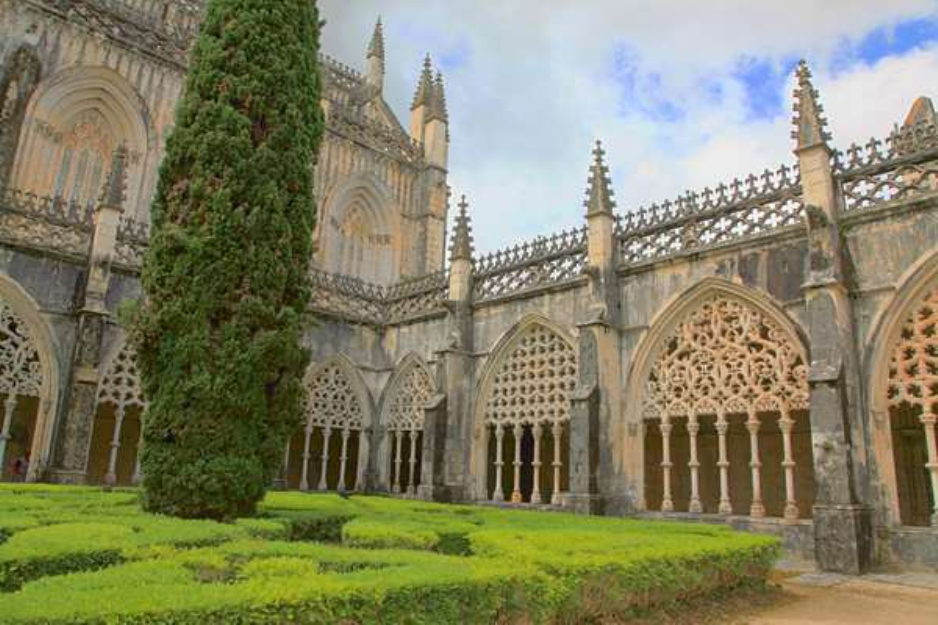}}
    \subfloat{\includegraphics[width = .10 \linewidth]{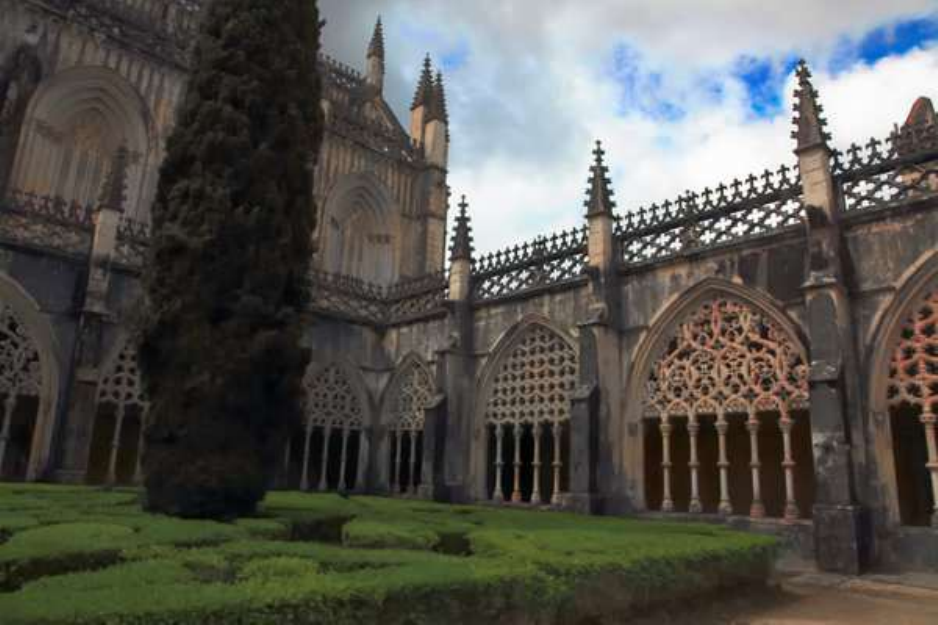}}
    \subfloat{\includegraphics[width = .10 \linewidth]{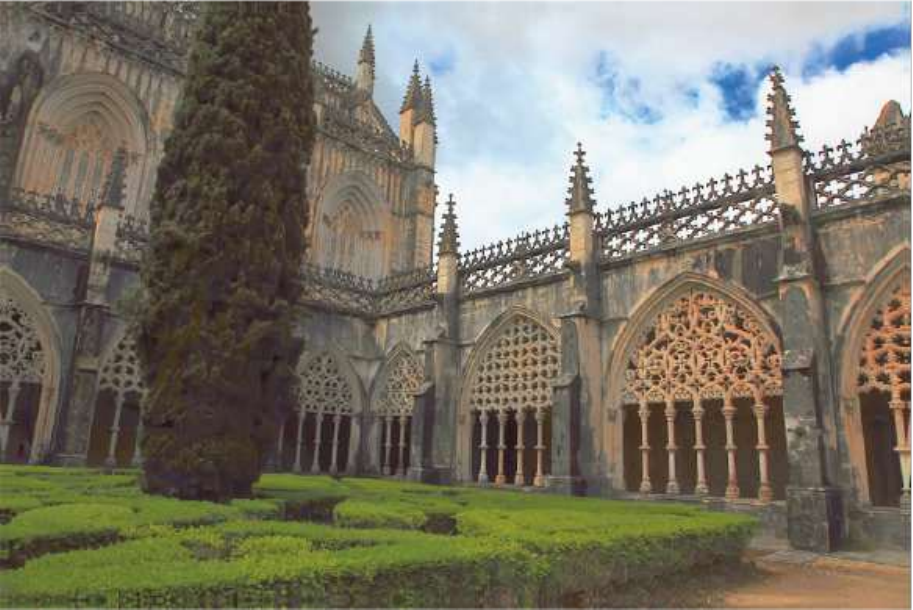}}
    \subfloat{\includegraphics[width = .10 \linewidth]{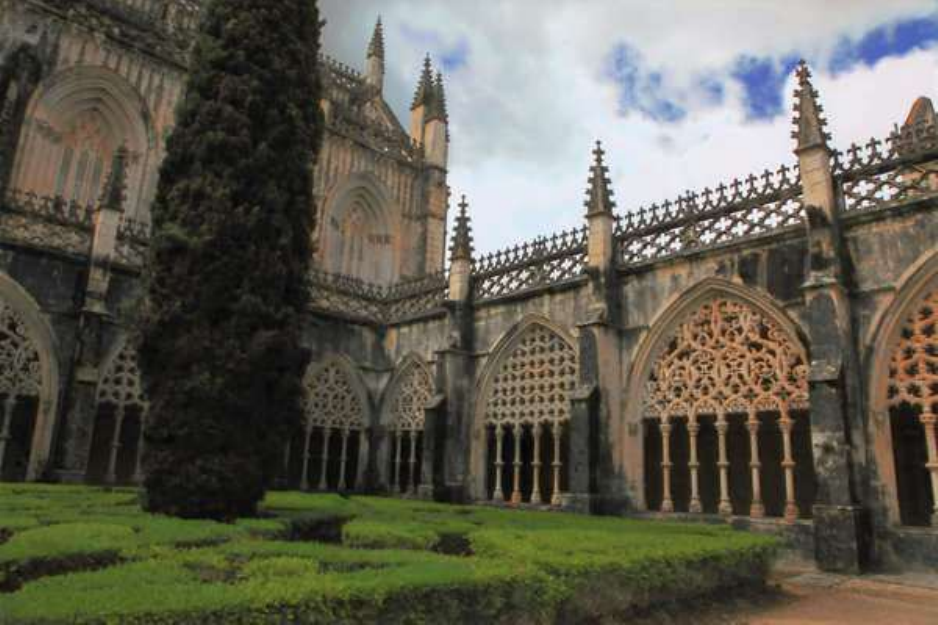}}
    \subfloat{\includegraphics[width = .10 \linewidth]{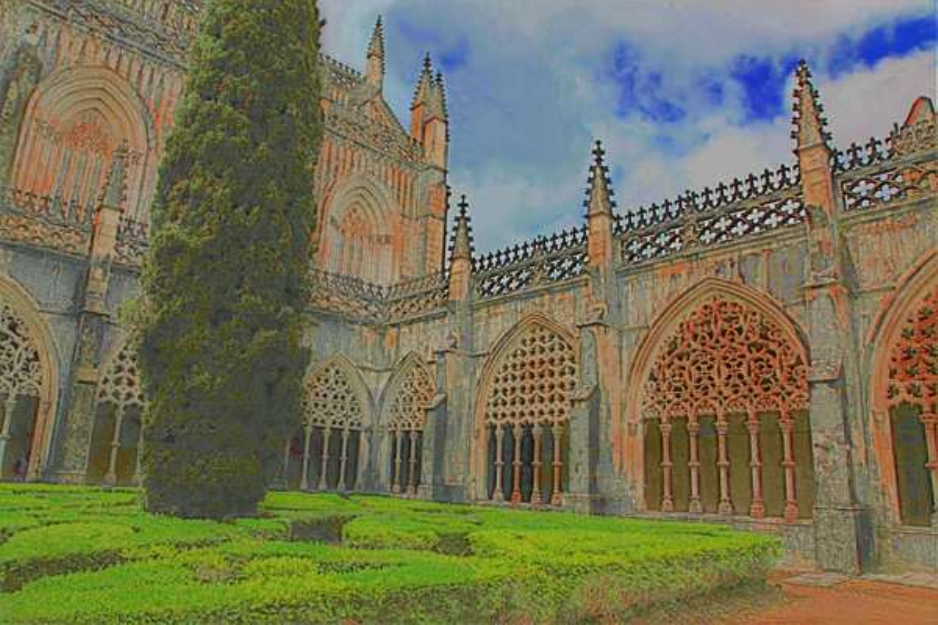}}
    \subfloat{\includegraphics[width = .10 \linewidth]{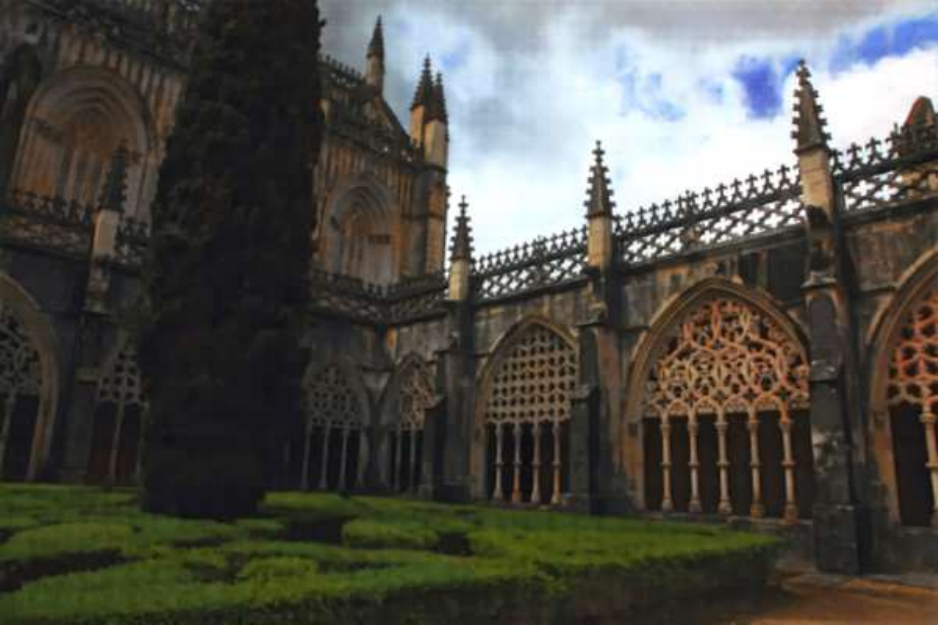}}
    \subfloat{\includegraphics[width = .10 \linewidth]{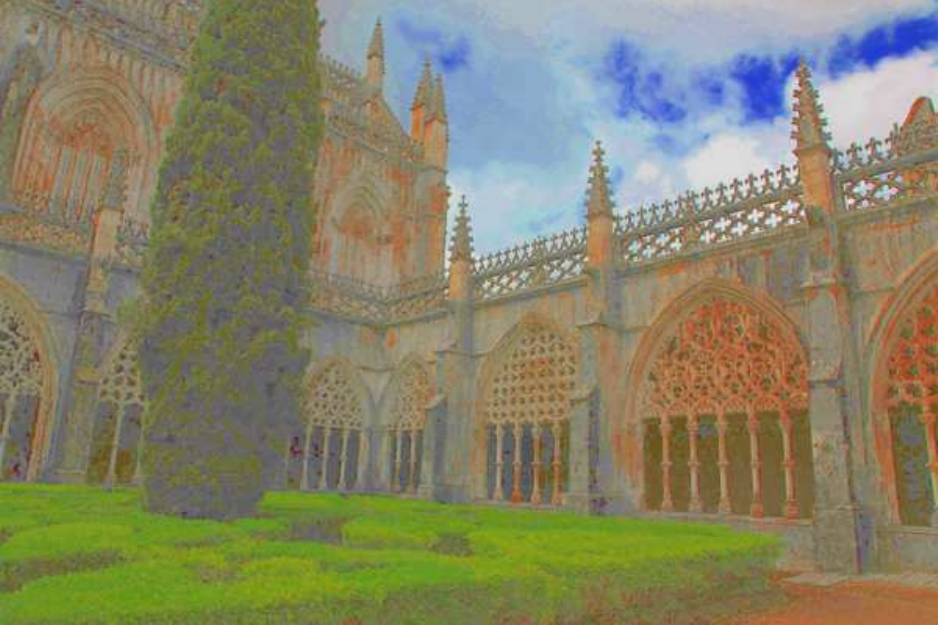}}
    \subfloat{\includegraphics[width = .10 \linewidth]{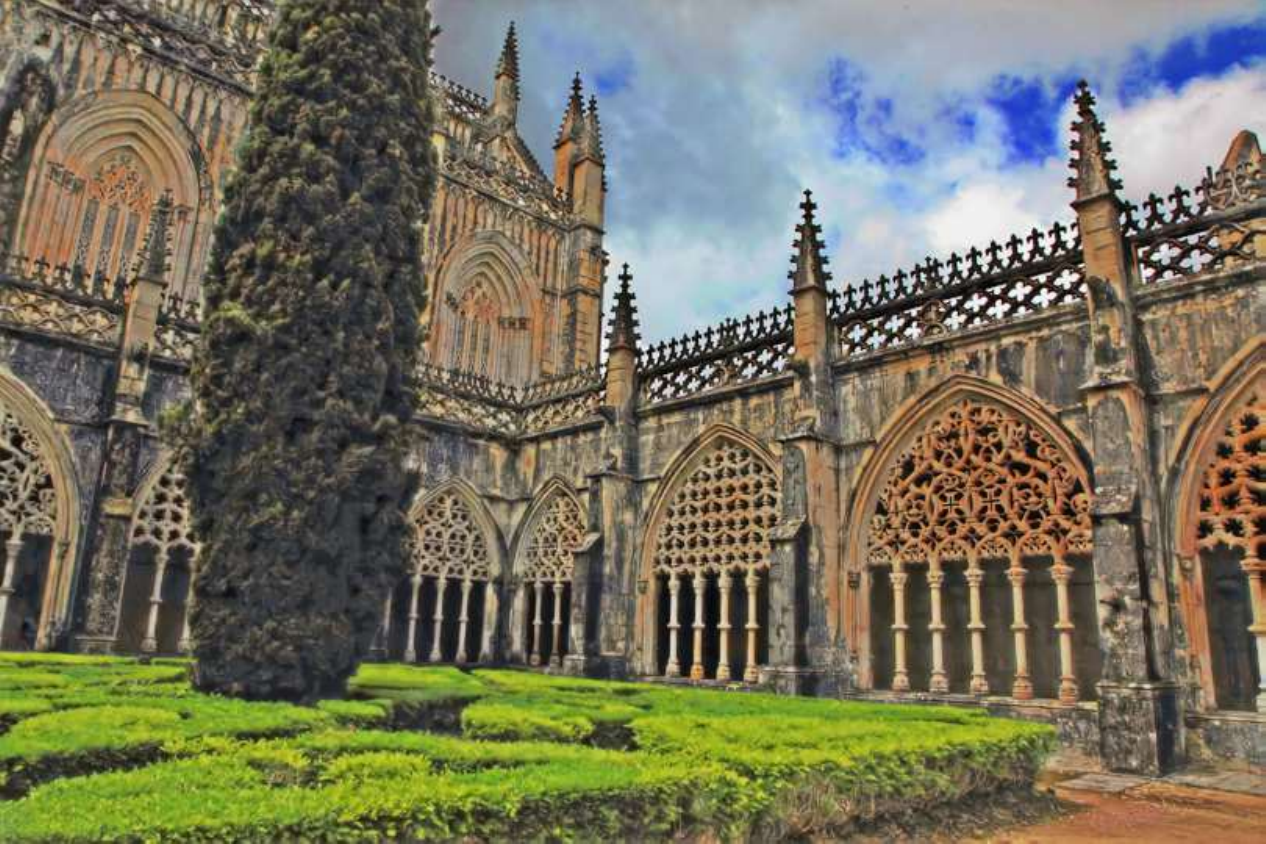}}
    \subfloat{\includegraphics[width = .10 \linewidth]{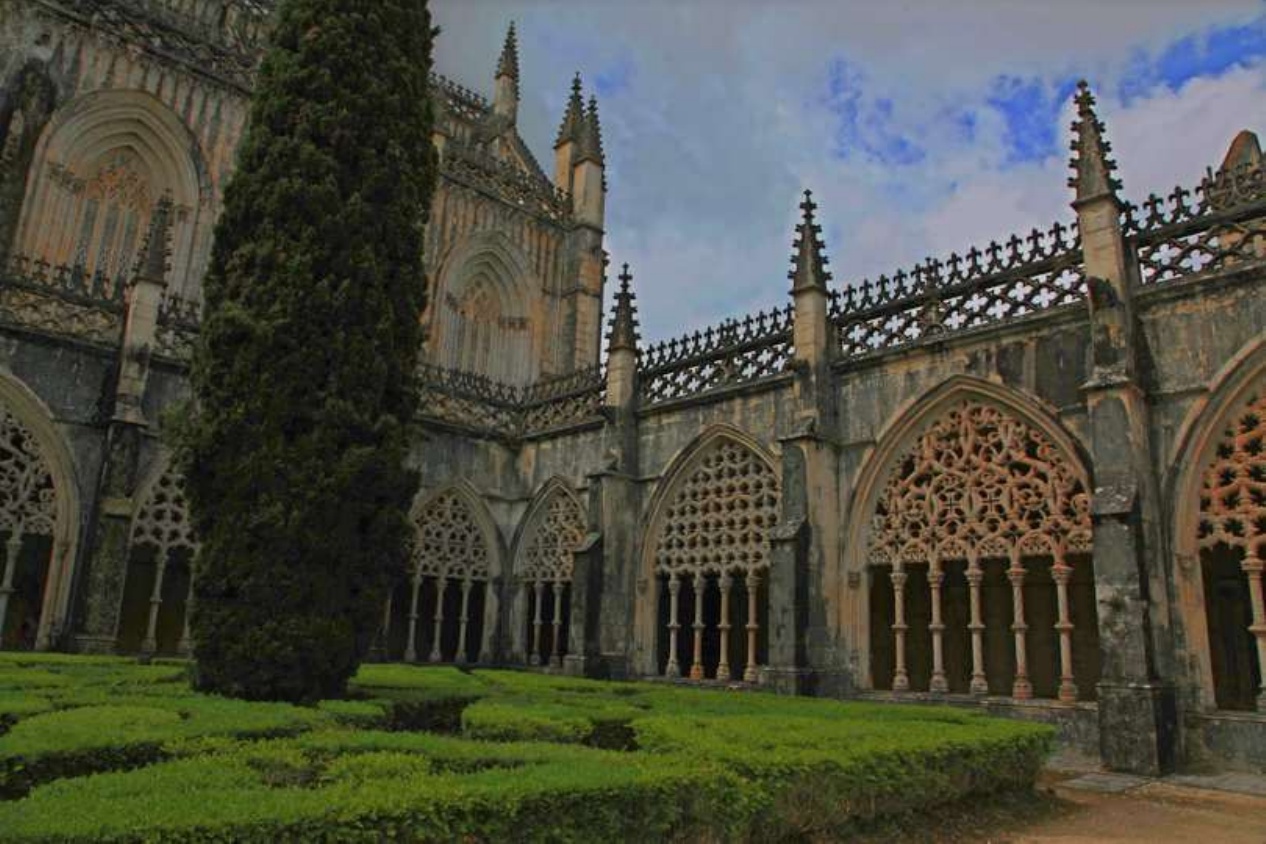}}\\\vspace{-0.1in}
    \subfloat{\includegraphics[width = .10 \linewidth]{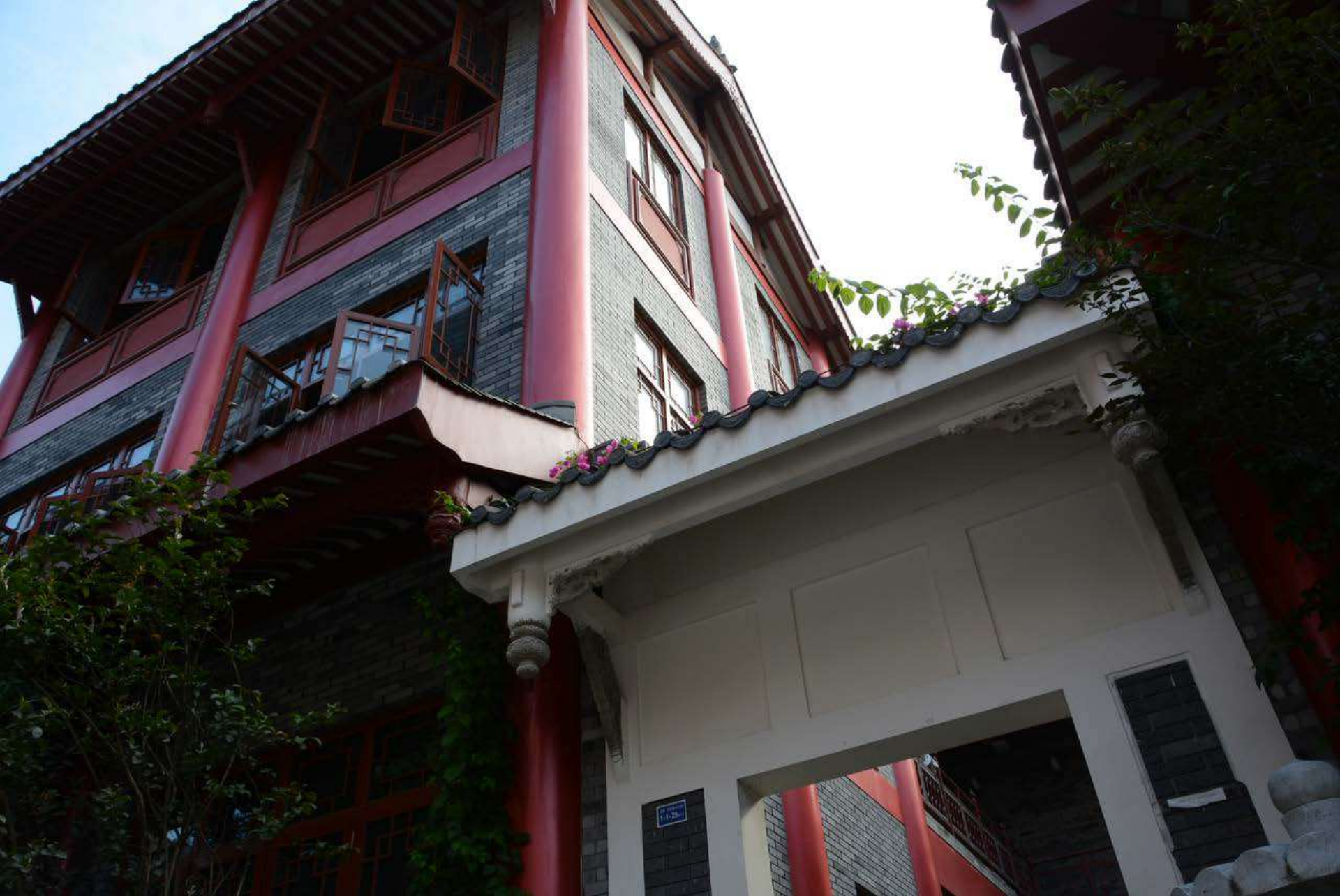}}
    \subfloat{\includegraphics[width = .10 \linewidth]{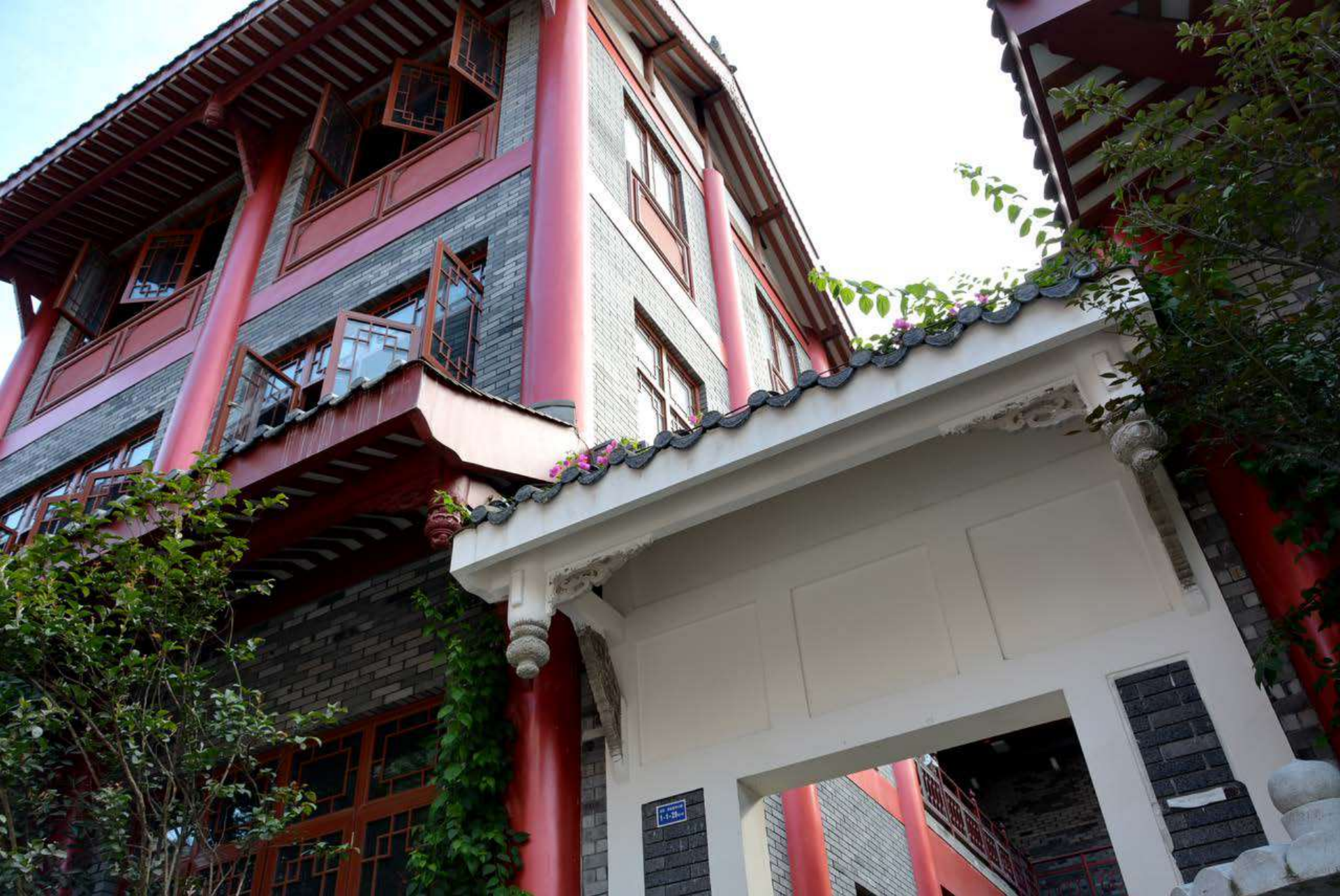}}
    \subfloat{\includegraphics[width = .10 \linewidth]{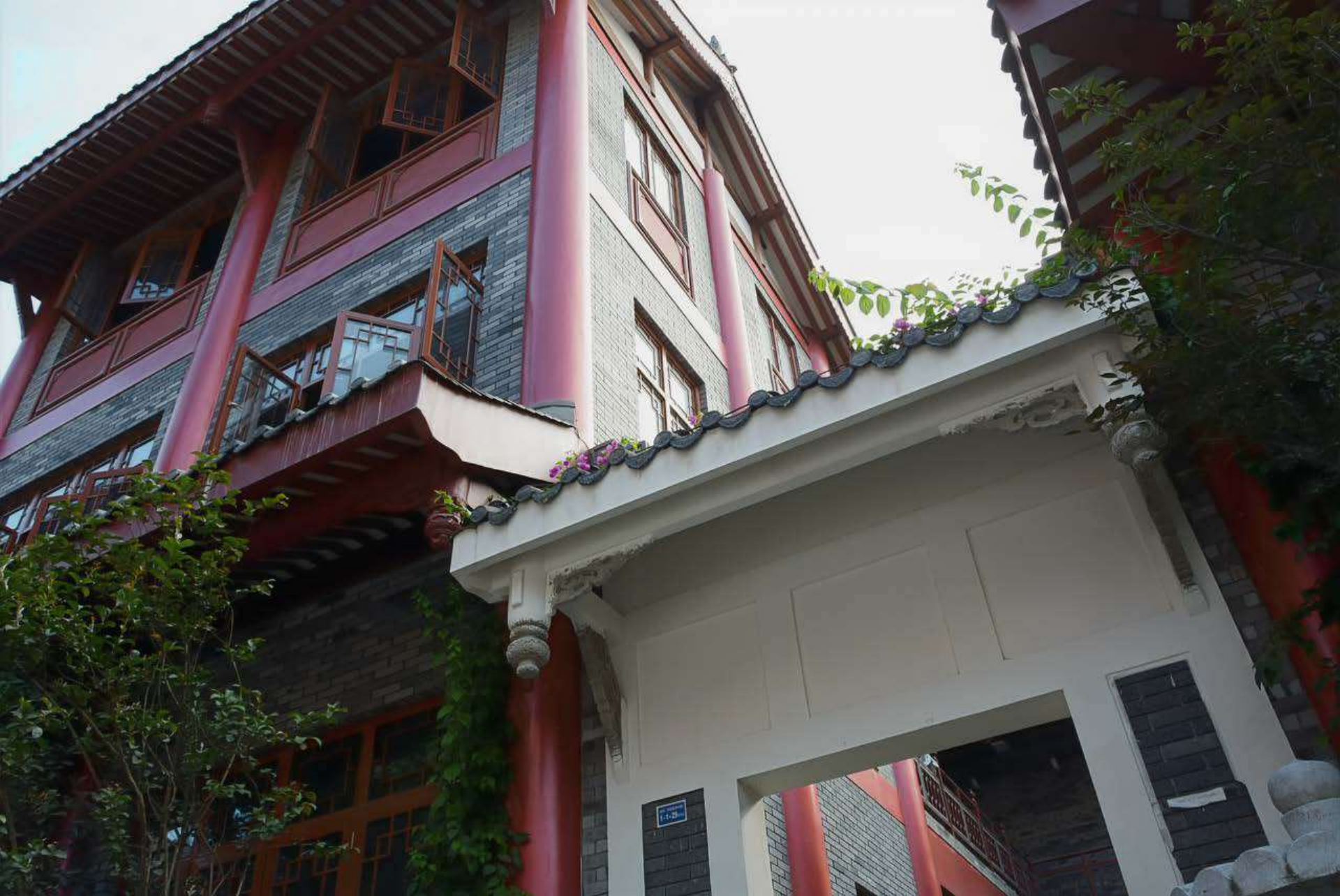}}
    \subfloat{\includegraphics[width = .10 \linewidth]{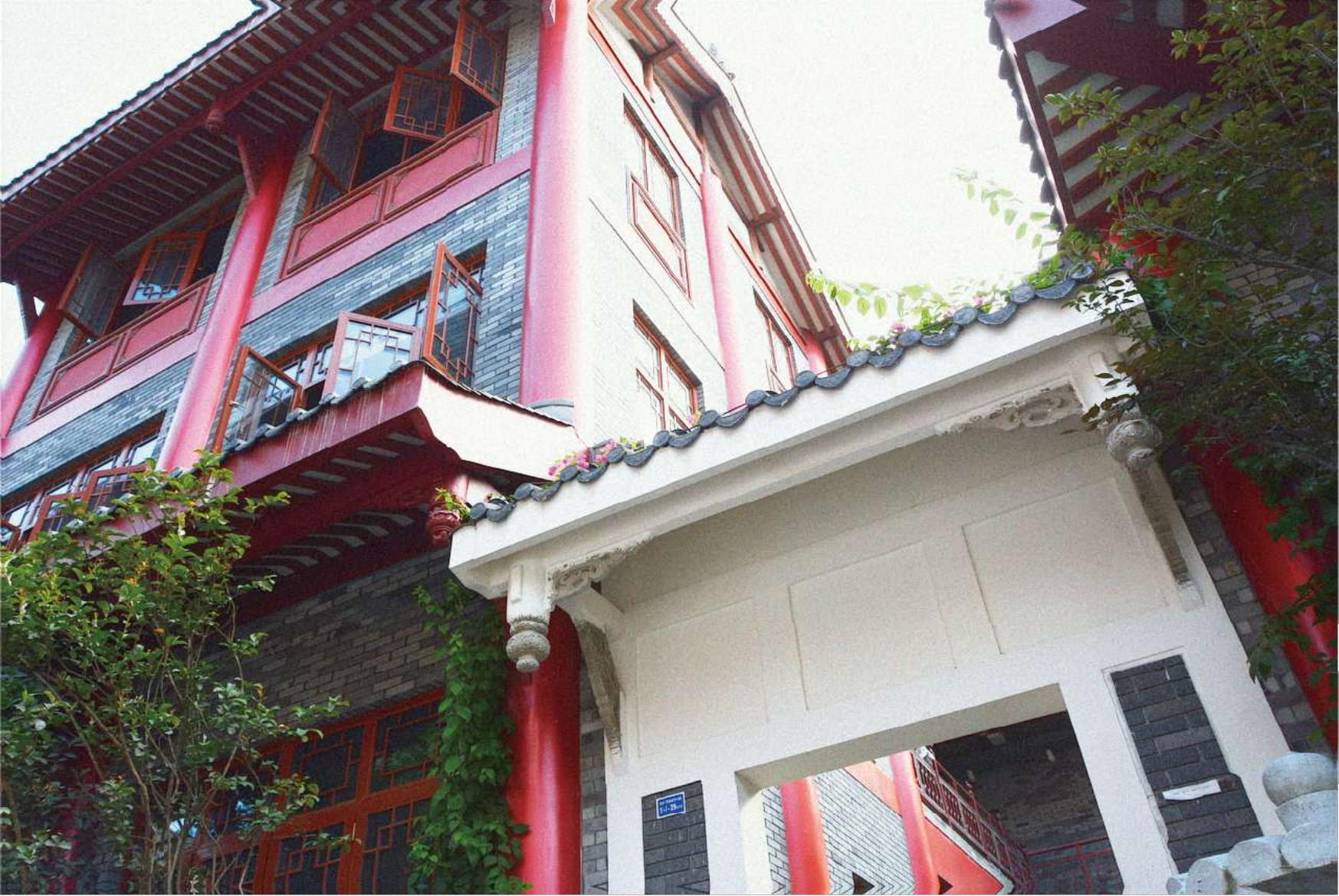}}
    \subfloat{\includegraphics[width = .10 \linewidth]{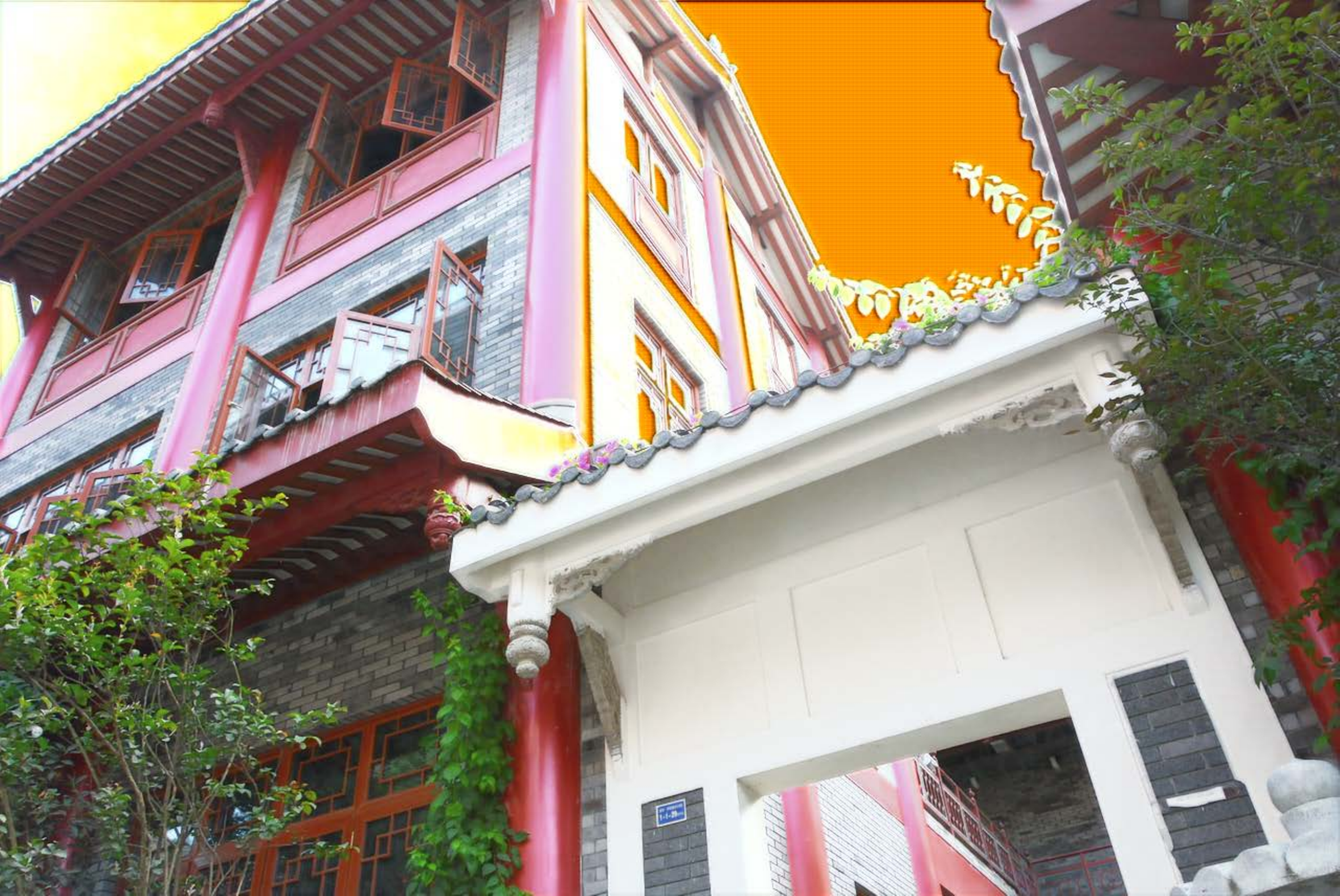}}
    \subfloat{\includegraphics[width = .10 \linewidth]{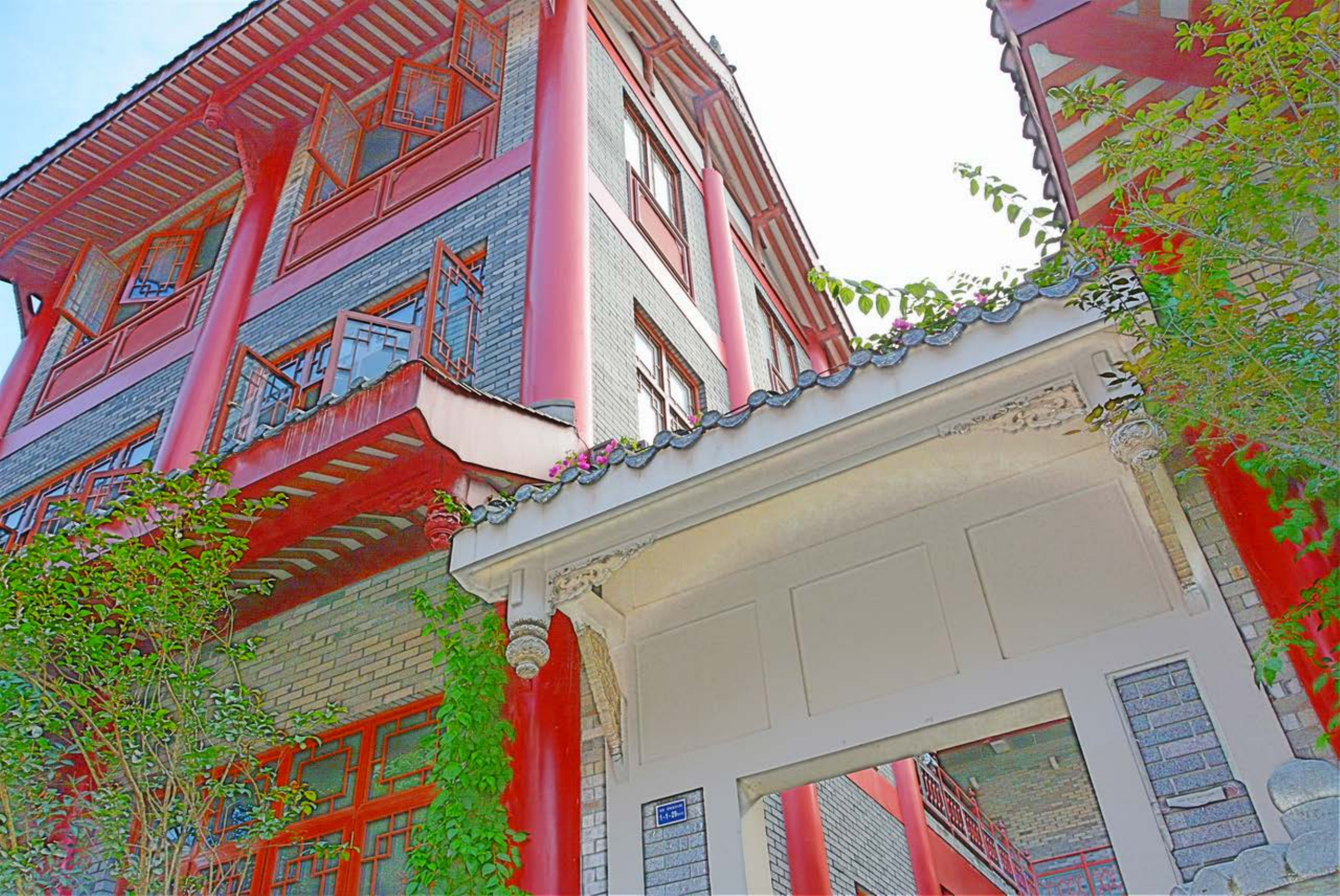}}
    \subfloat{\includegraphics[width = .10 \linewidth]{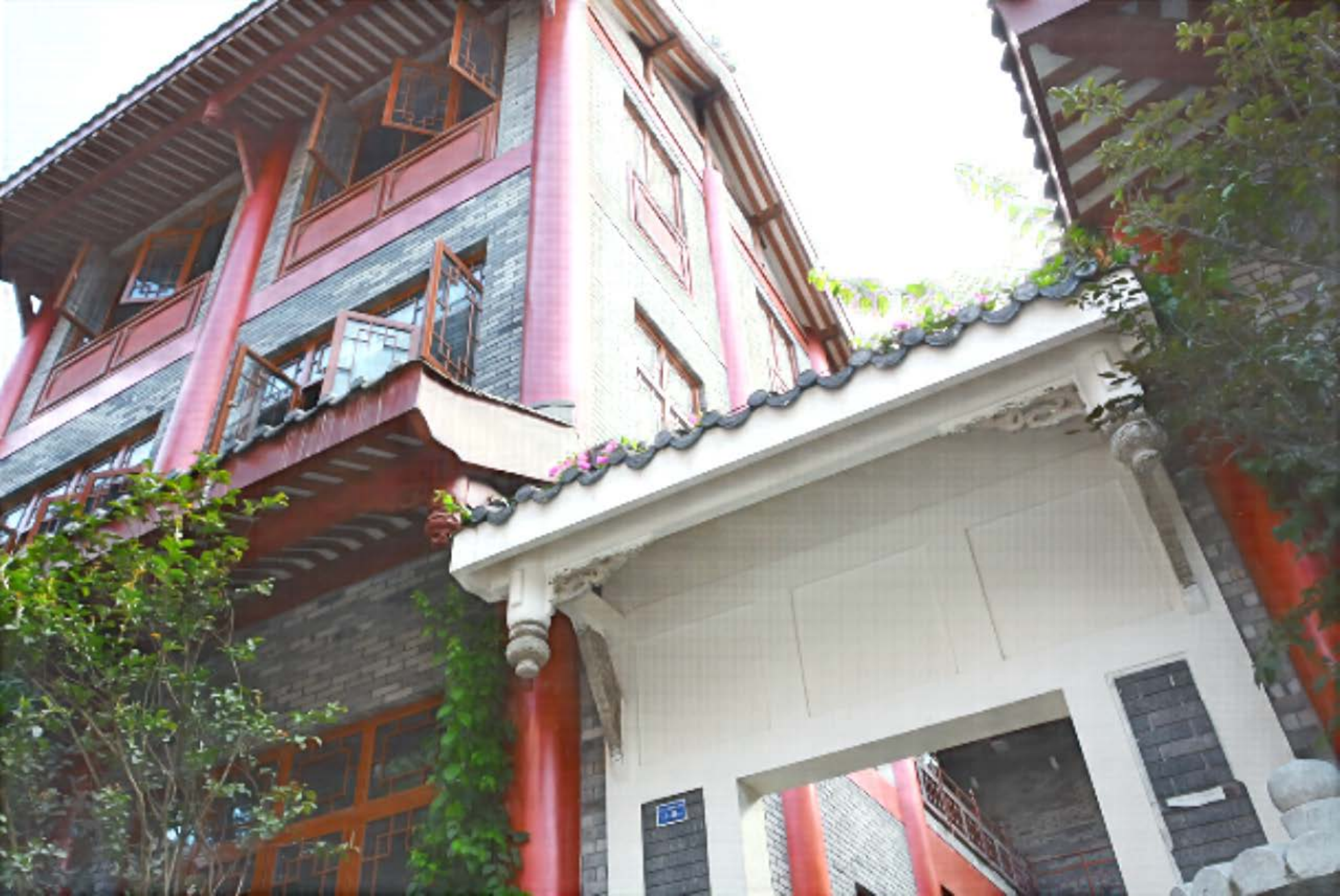}}
    \subfloat{\includegraphics[width = .10 \linewidth]{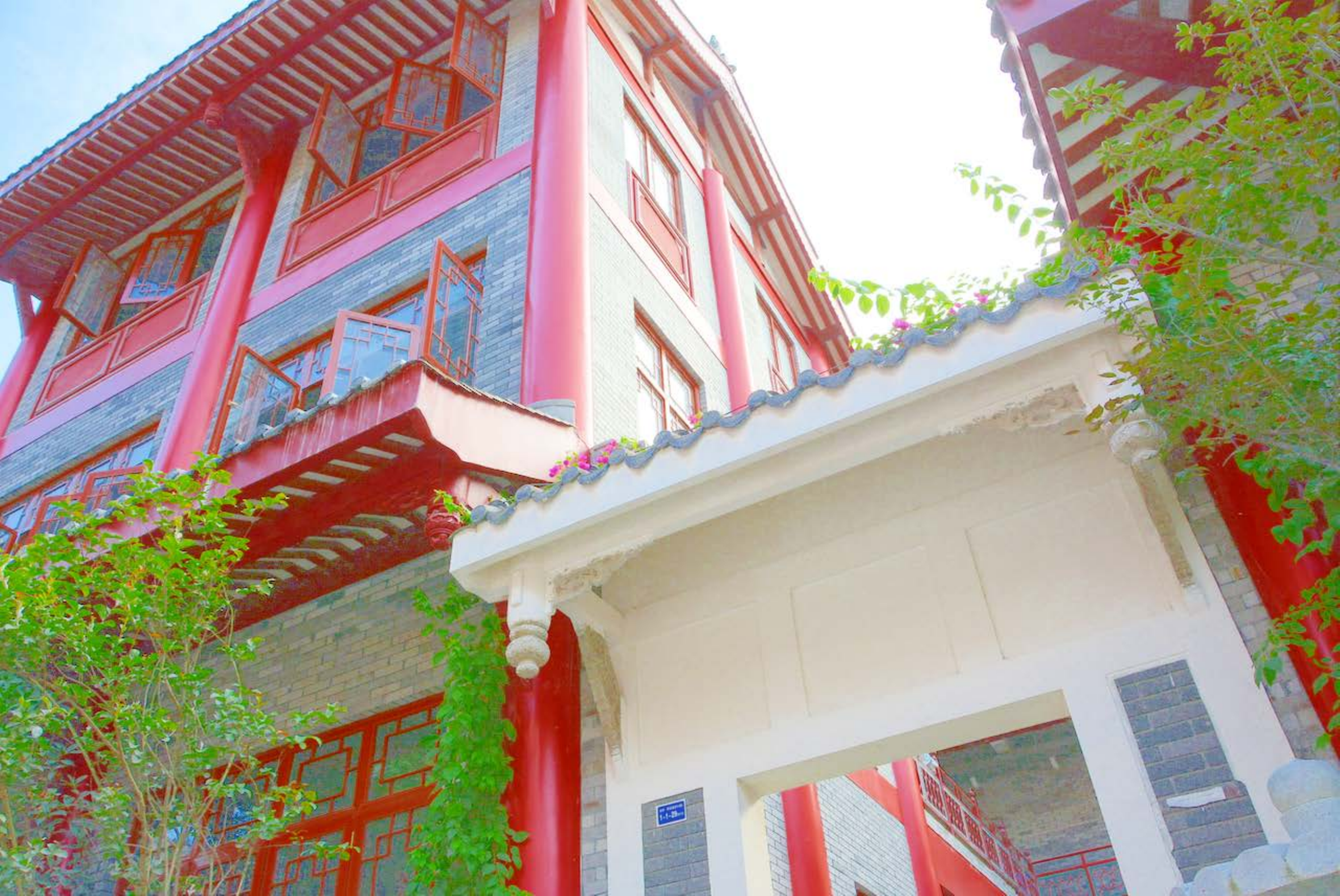}}
    \subfloat{\includegraphics[width = .10 \linewidth]{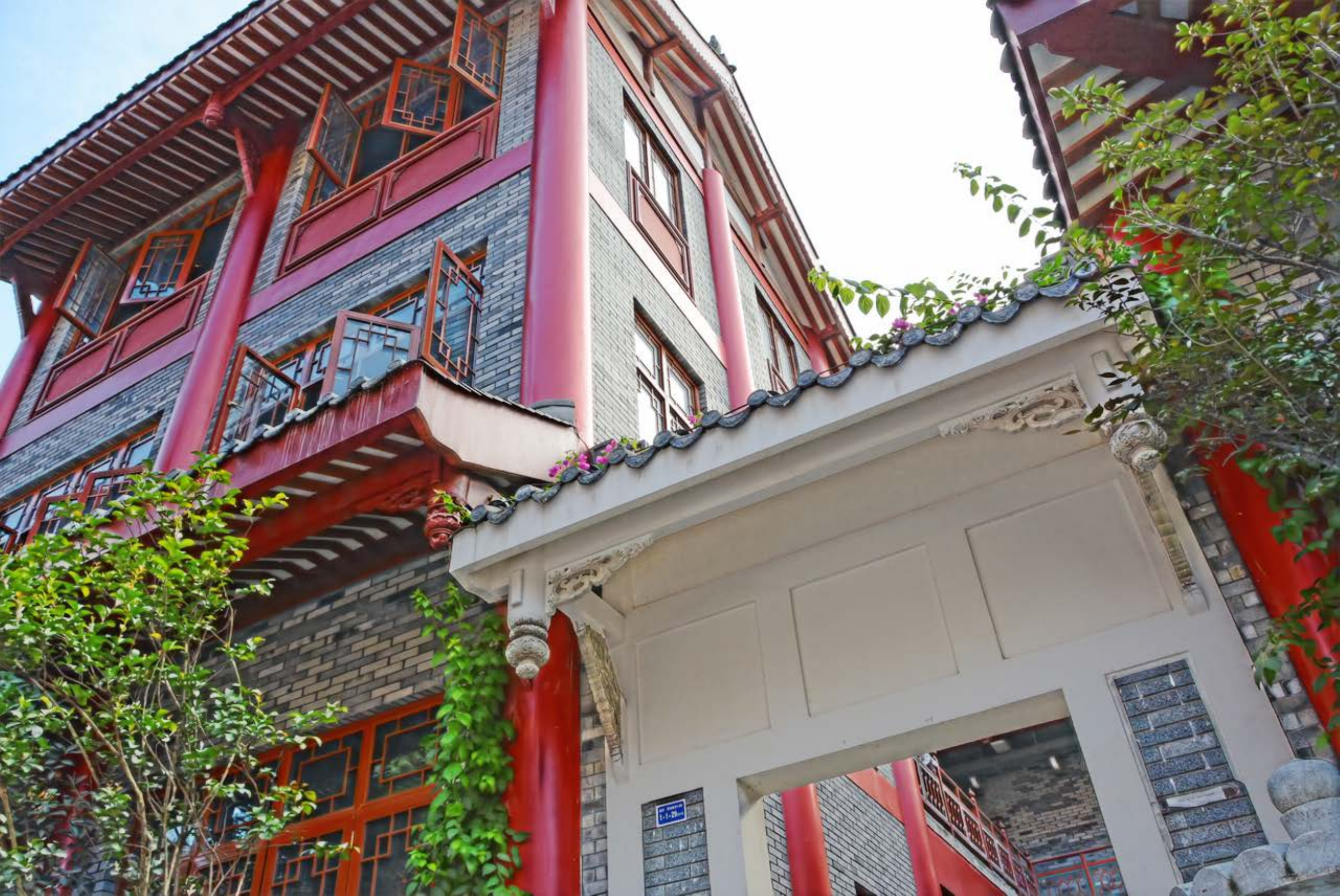}}
    \subfloat{\includegraphics[width = .10 \linewidth]{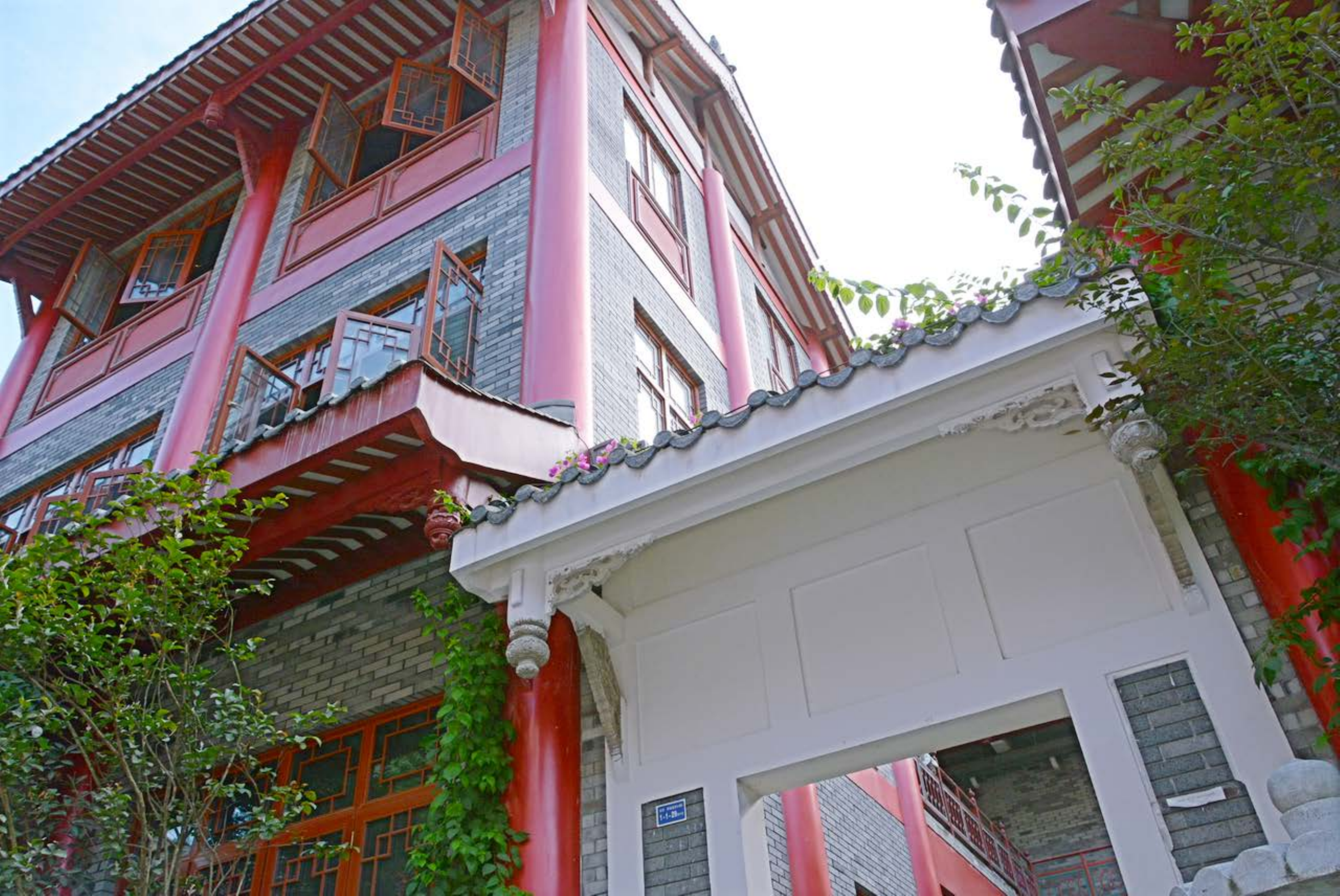}}\\\vspace{-0.16in}
    \subfloat{\includegraphics[width = .10 \linewidth]{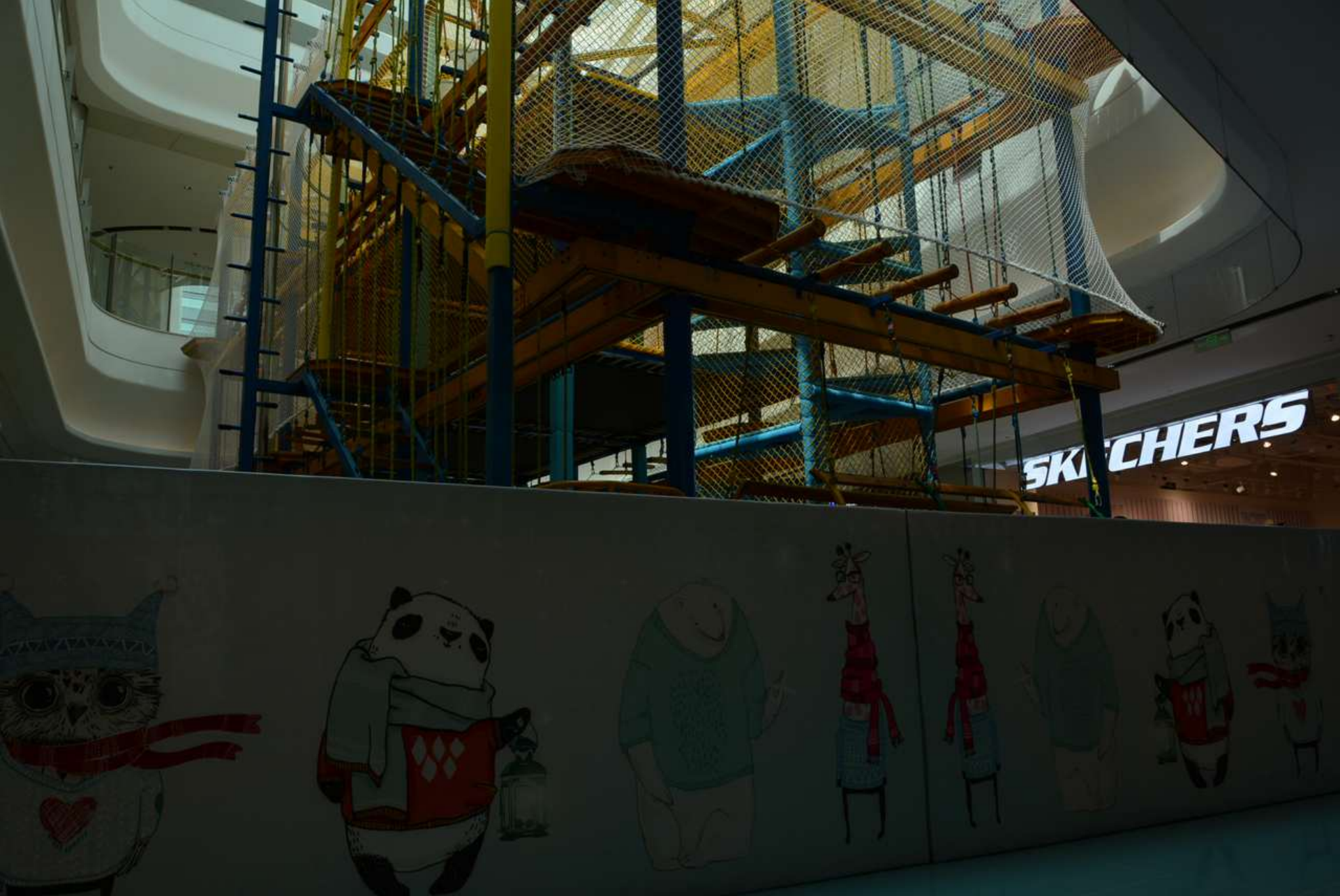}}
    \subfloat{\includegraphics[width = .10 \linewidth]{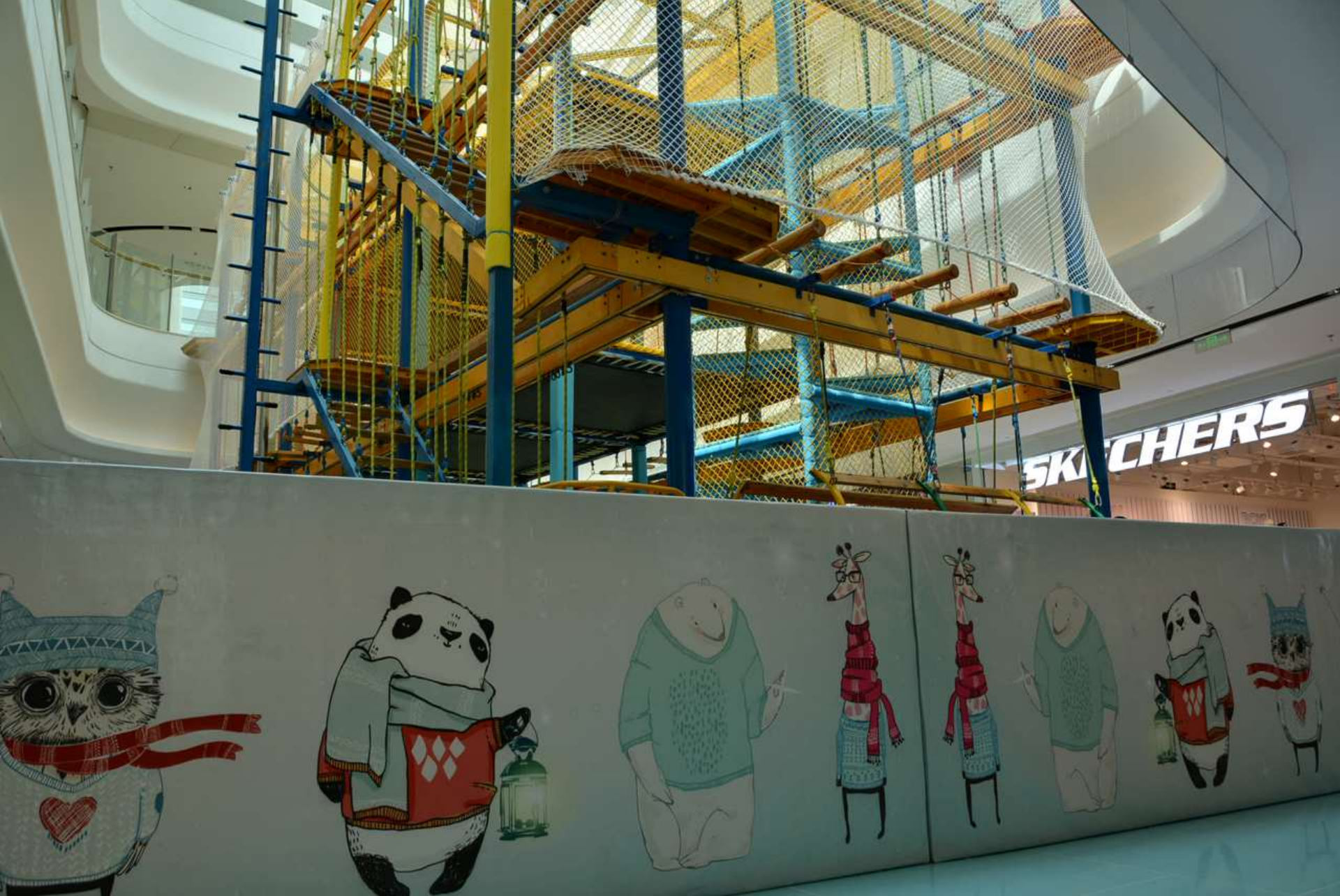}}
    \subfloat{\includegraphics[width = .10 \linewidth]{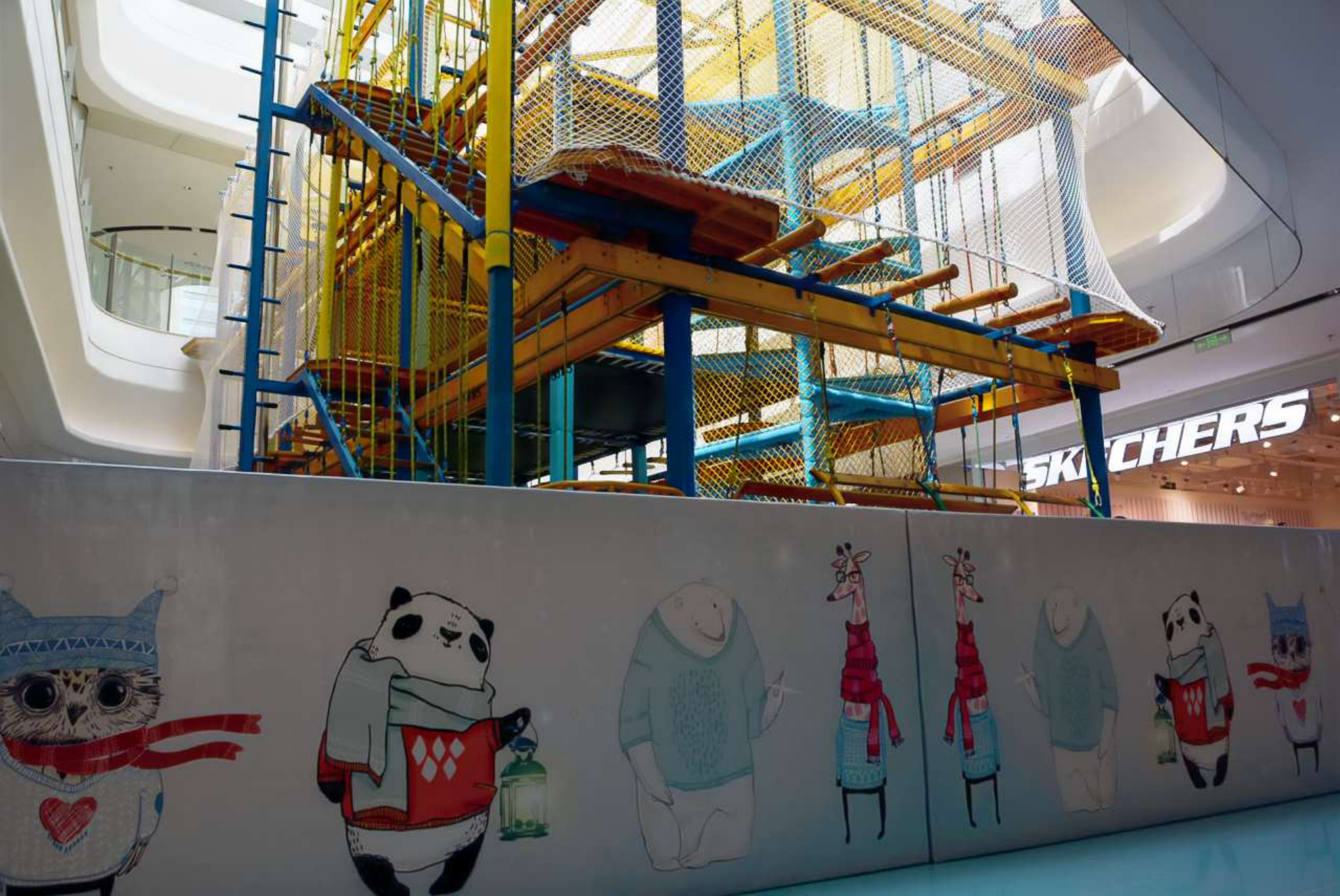}}
    \subfloat{\includegraphics[width = .10 \linewidth]{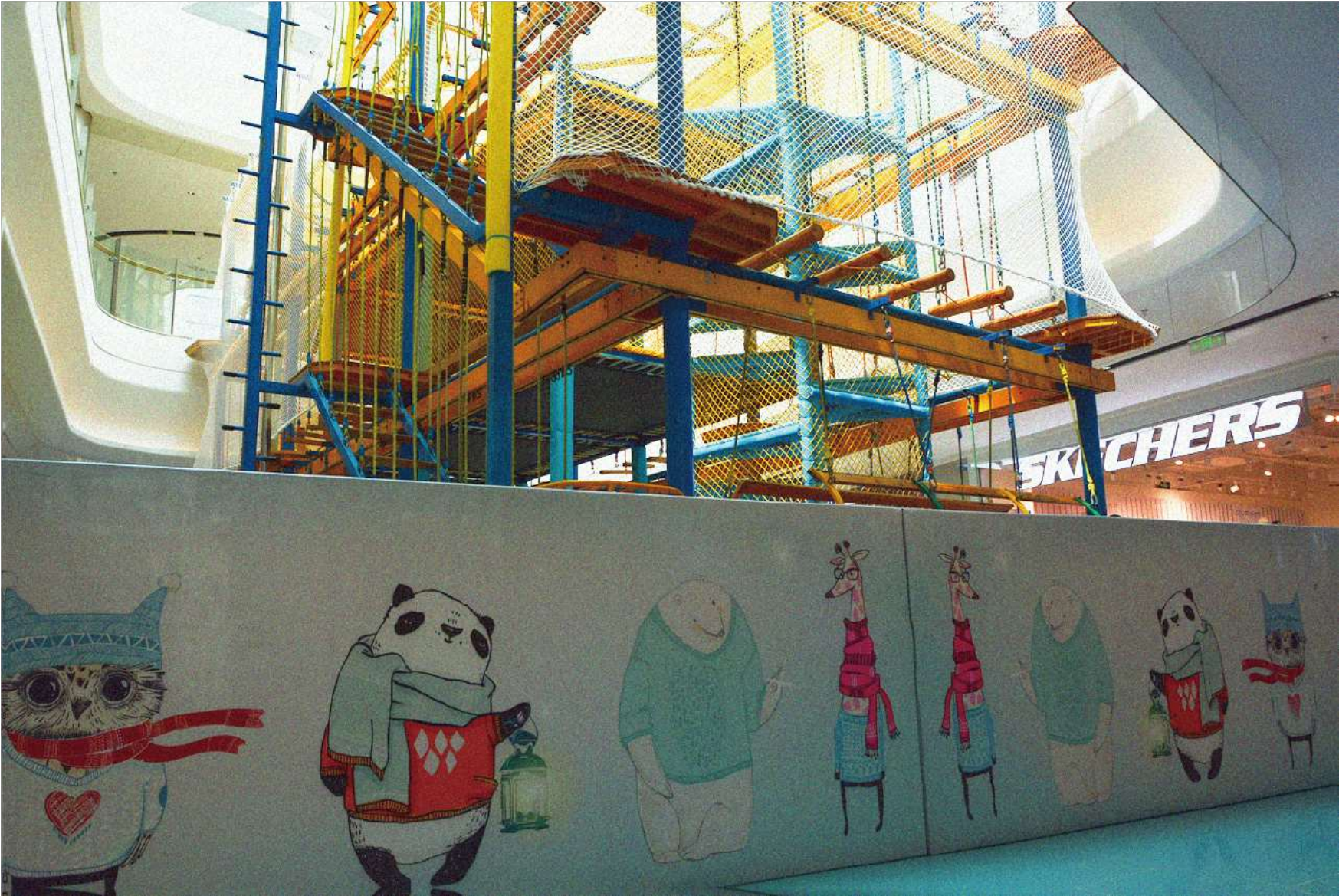}}
    \subfloat{\includegraphics[width = .10 \linewidth]{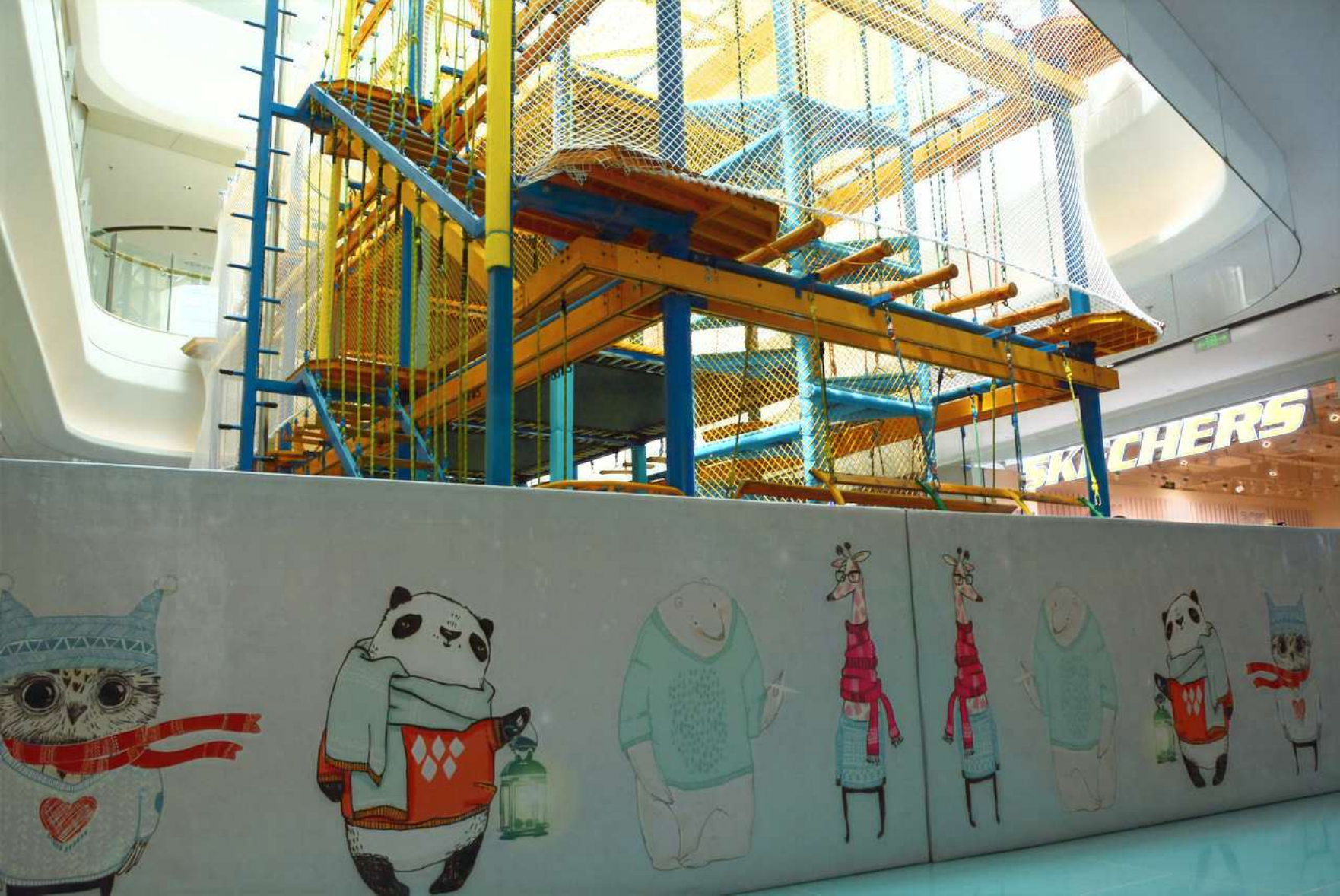}}
    \subfloat{\includegraphics[width = .10 \linewidth]{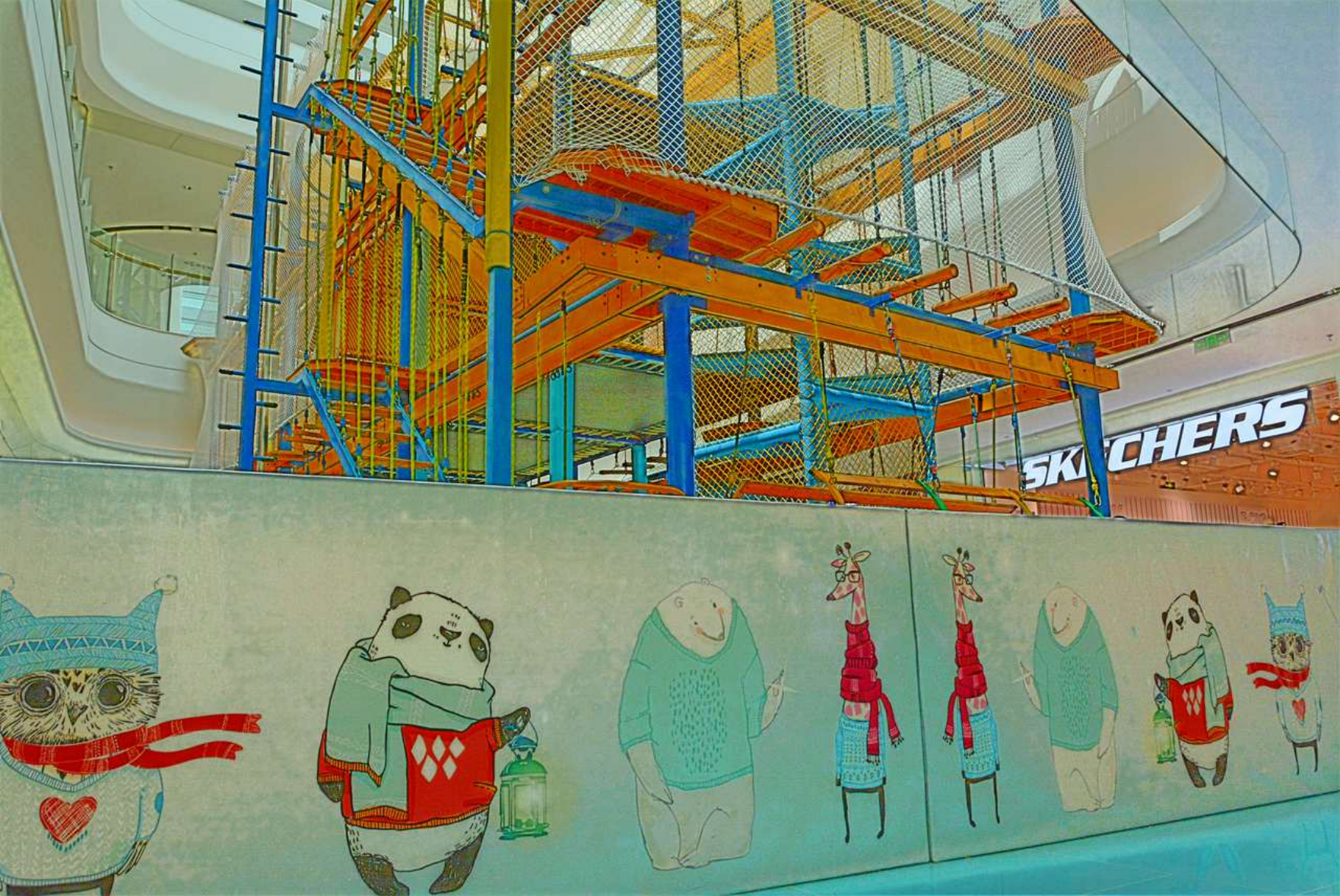}}
    \subfloat{\includegraphics[width = .10 \linewidth]{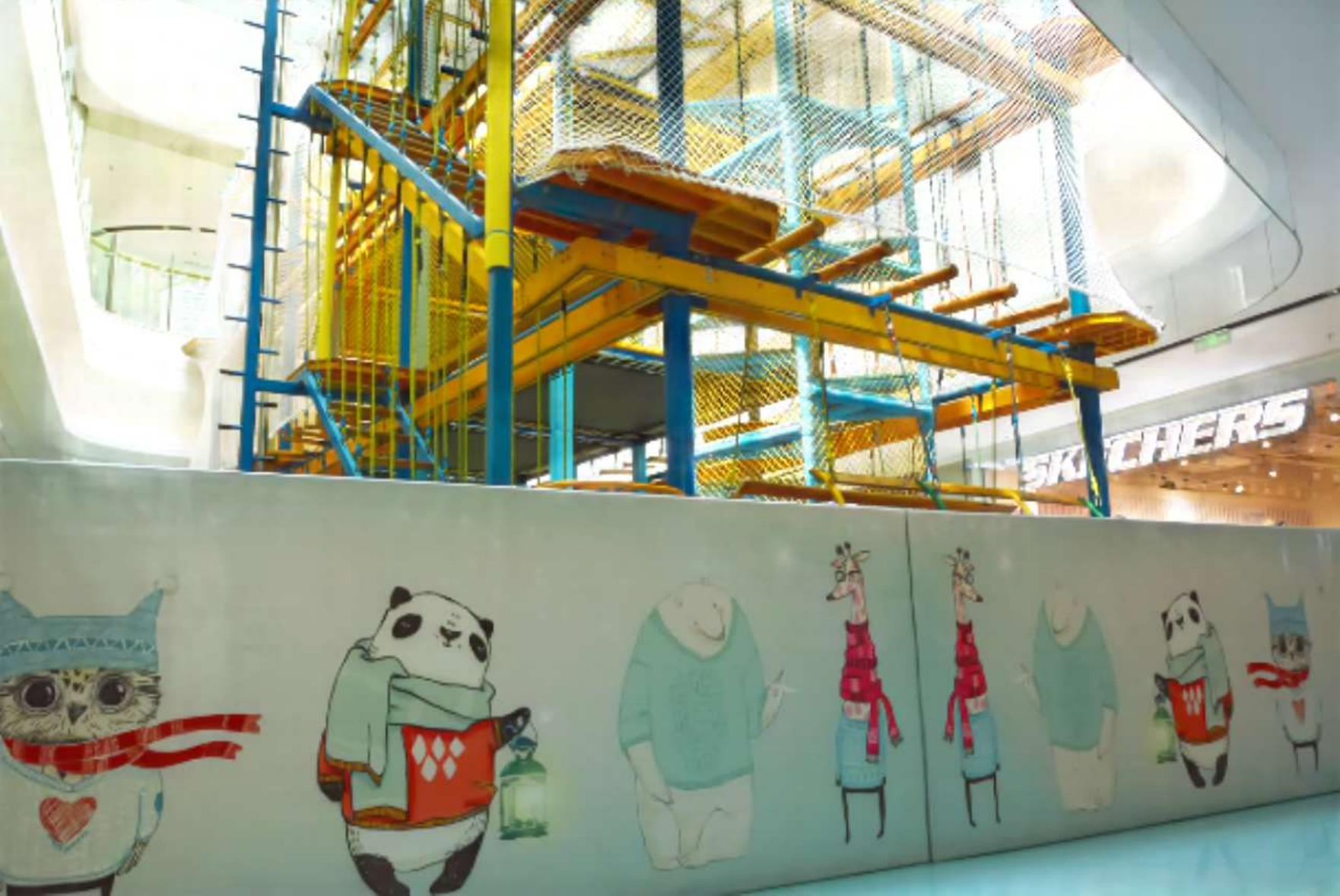}}
    \subfloat{\includegraphics[width = .10 \linewidth]{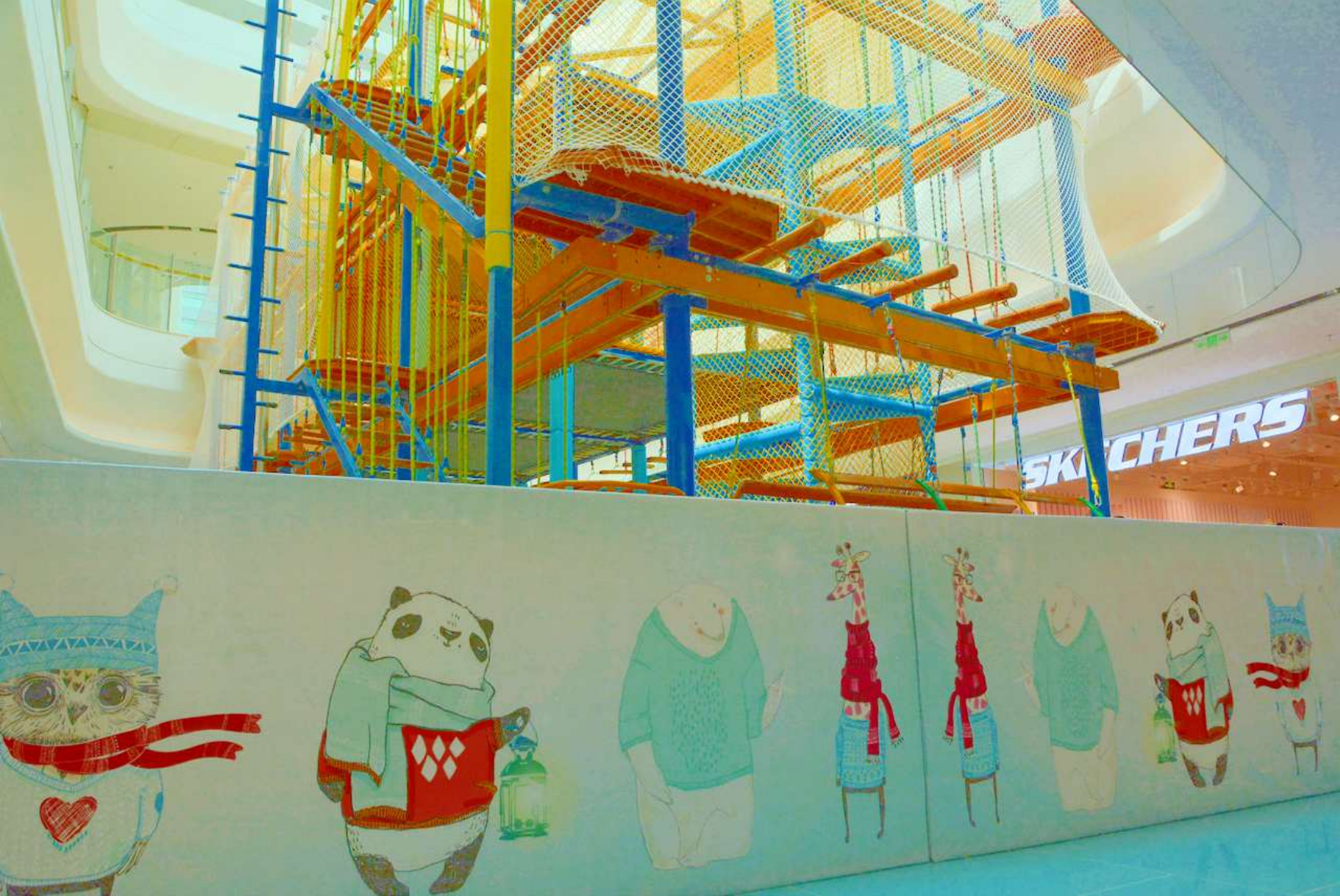}}
    \subfloat{\includegraphics[width = .10 \linewidth]{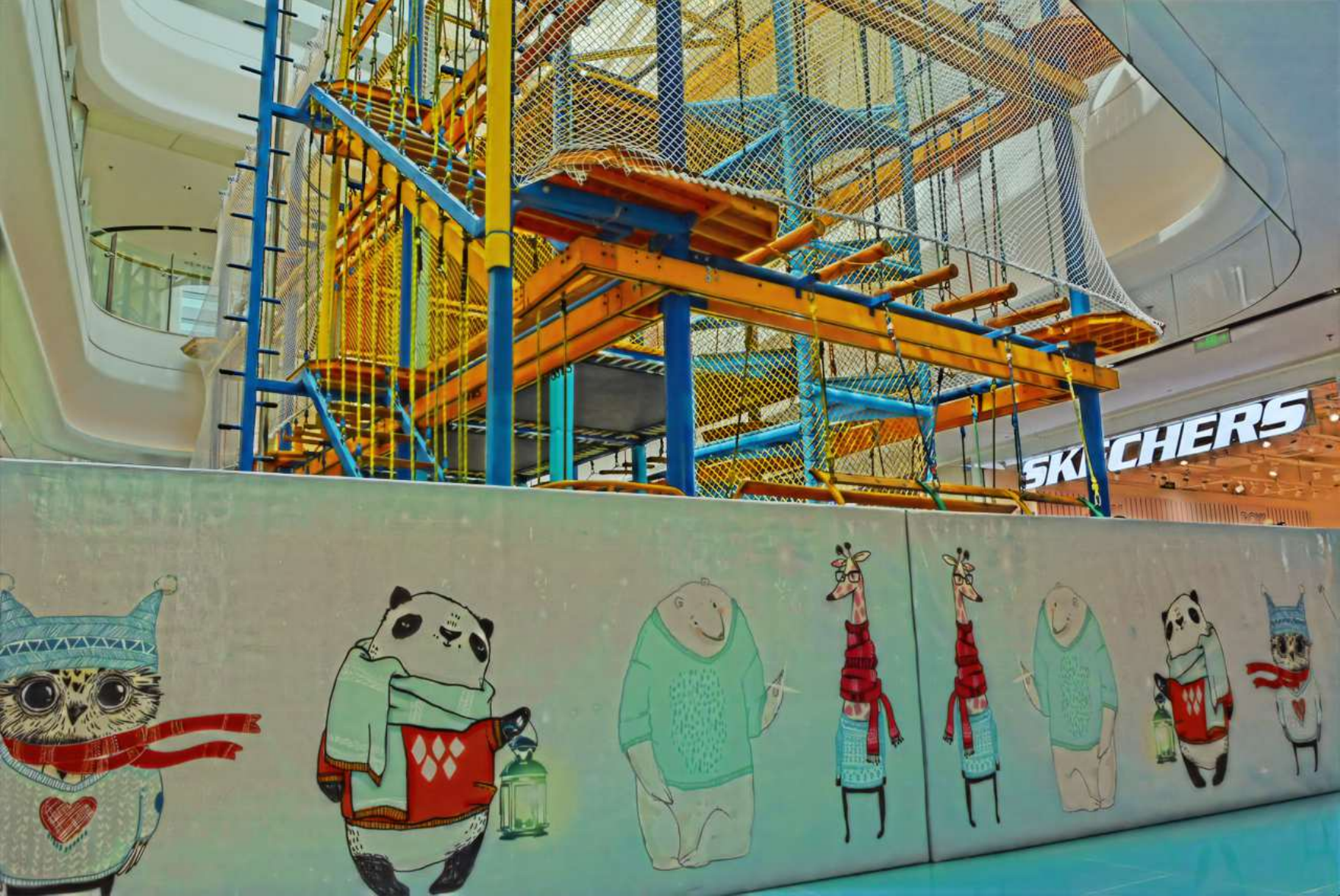}}
    \subfloat{\includegraphics[width = .10 \linewidth]{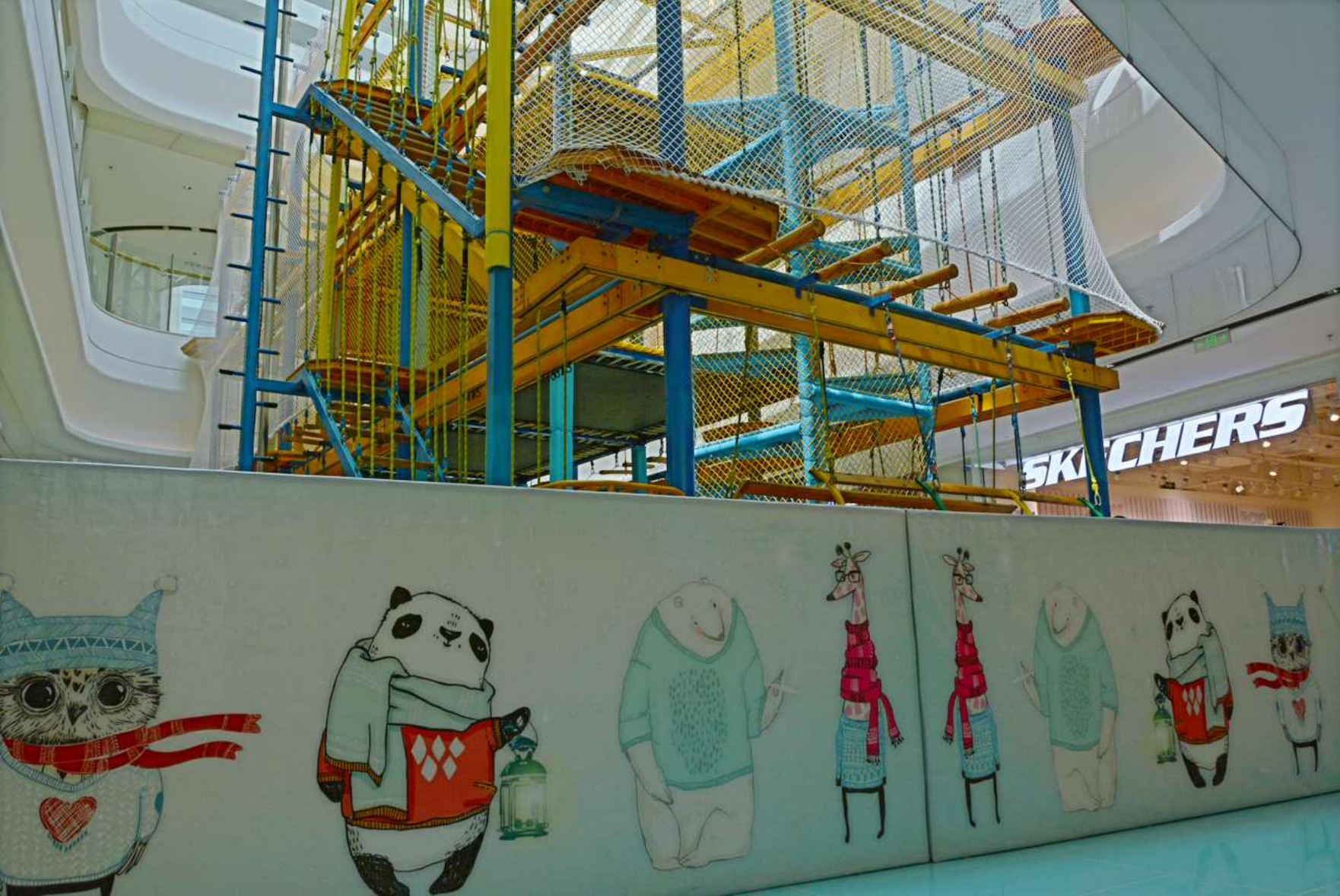}}\\\vspace{-0.16in}
    \subfloat{\includegraphics[width = .10 \linewidth]{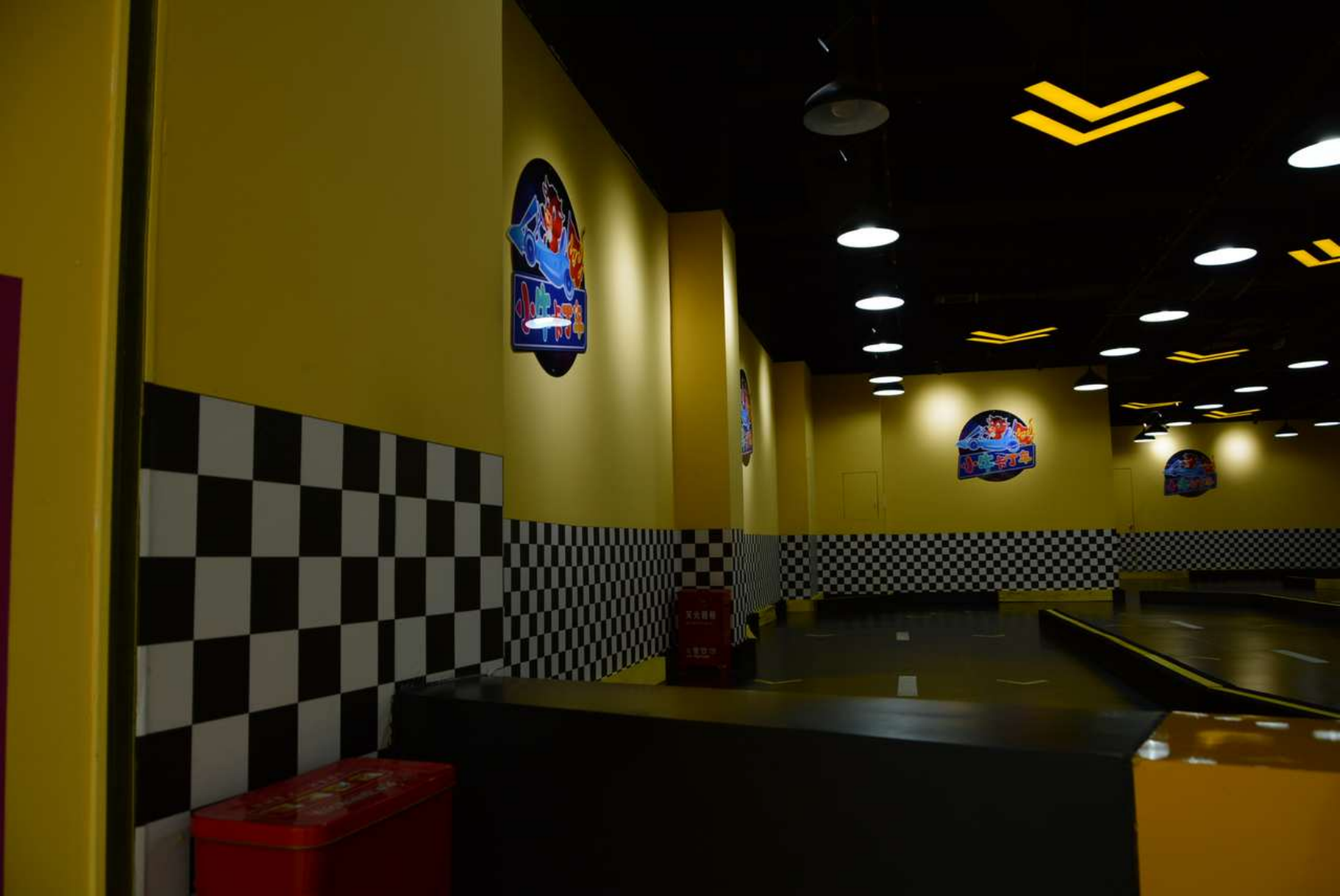}}
    \subfloat{\includegraphics[width = .10 \linewidth]{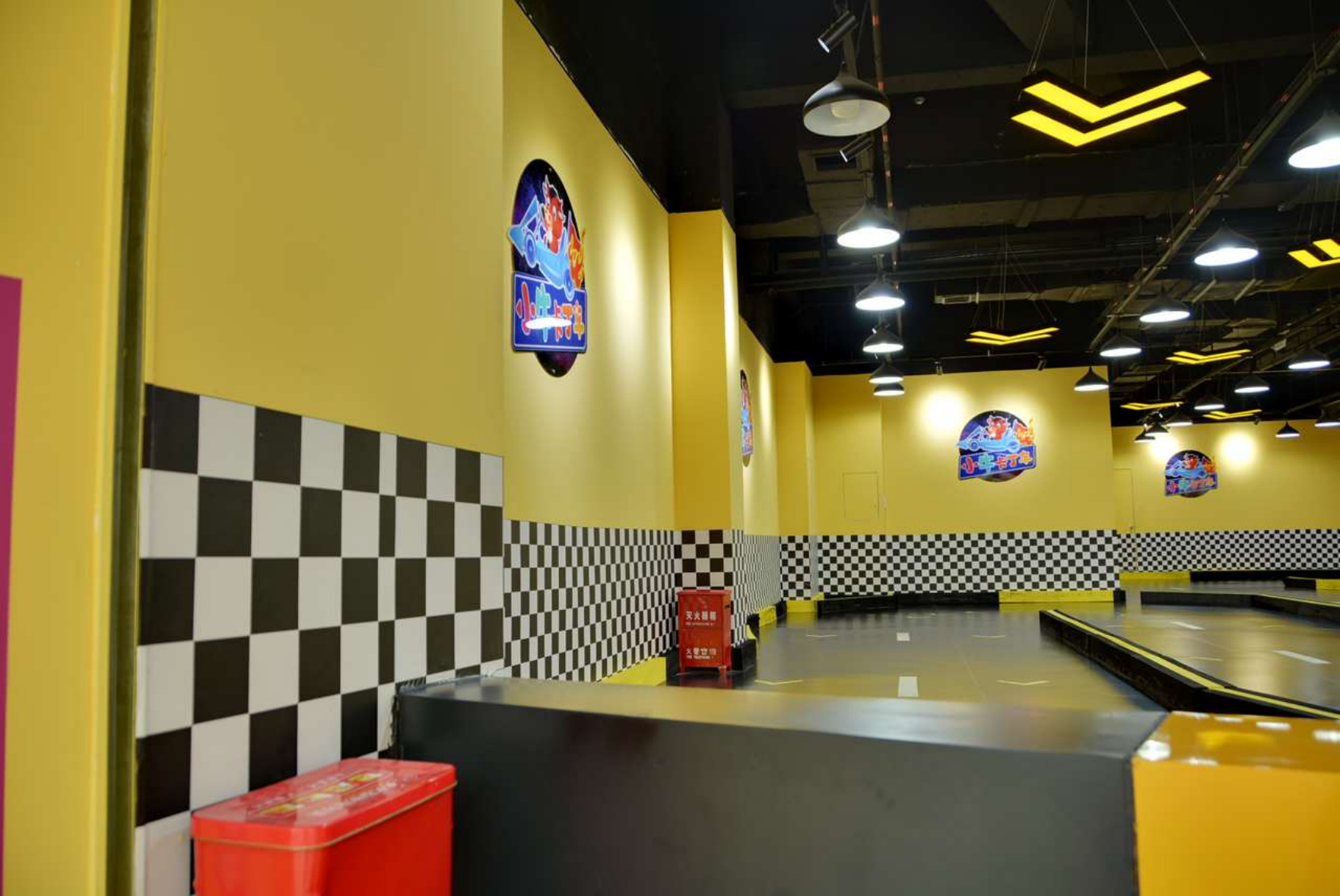}}
    \subfloat{\includegraphics[width = .10 \linewidth]{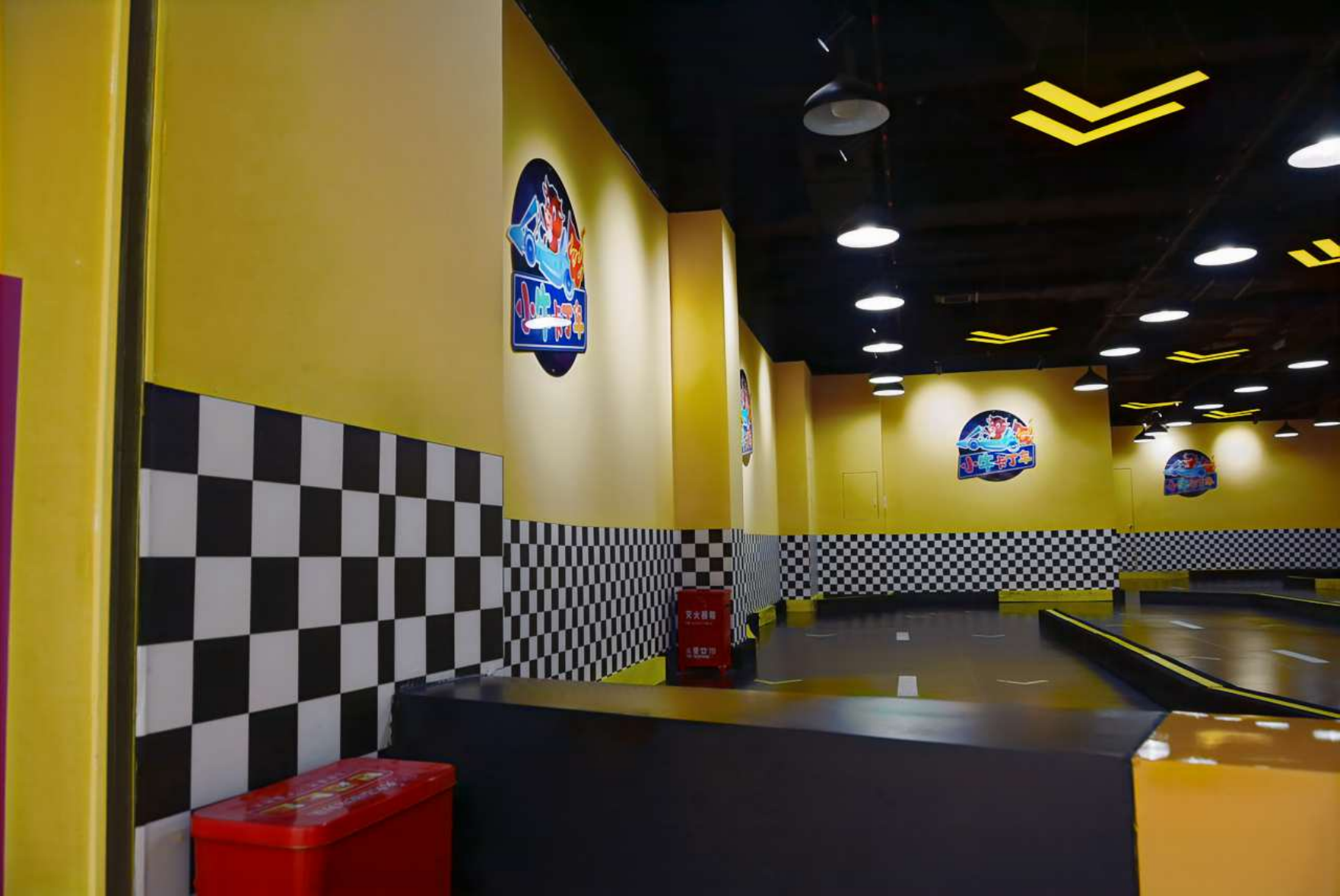}}
    \subfloat{\includegraphics[width = .10 \linewidth]{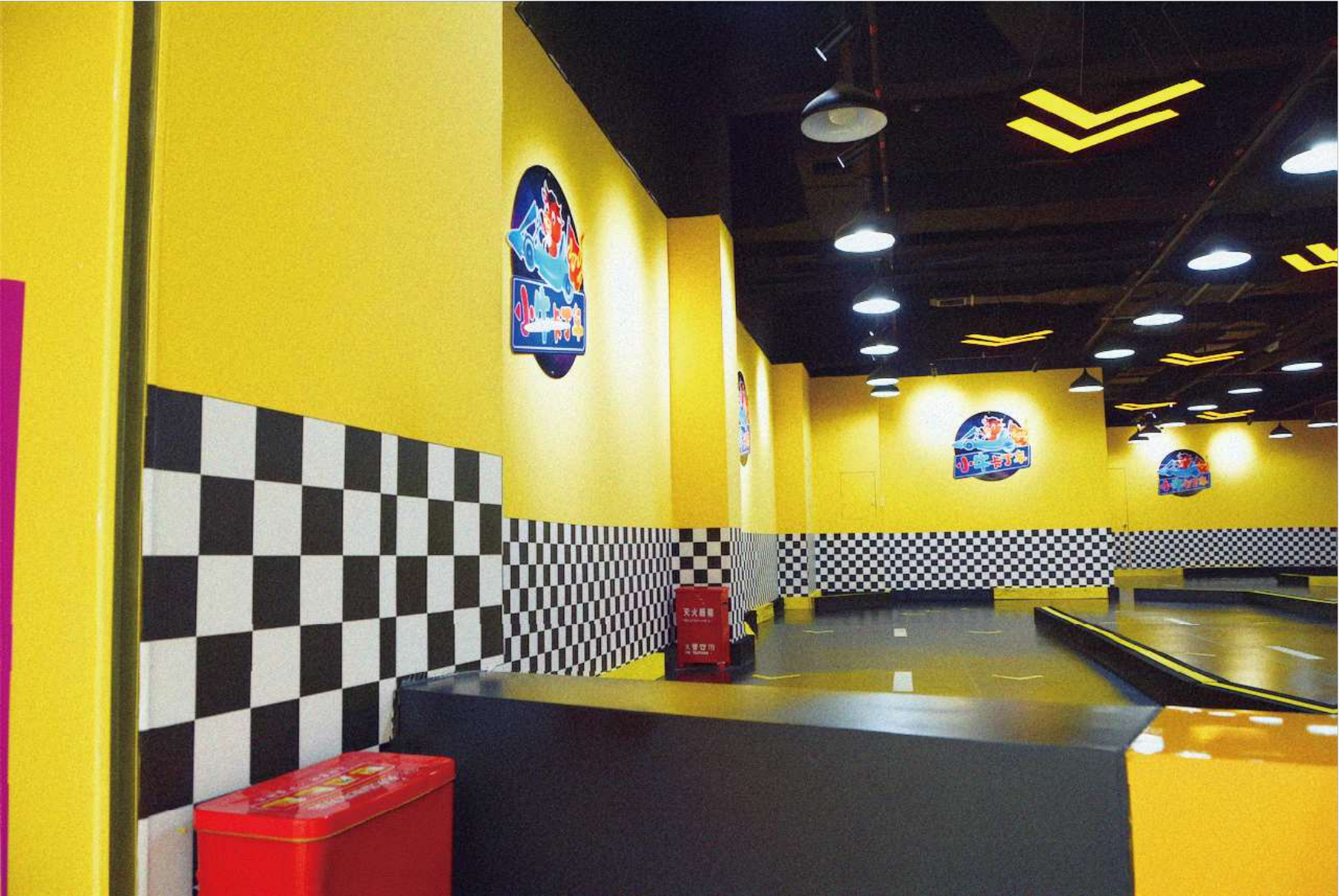}}
    \subfloat{\includegraphics[width = .10 \linewidth]{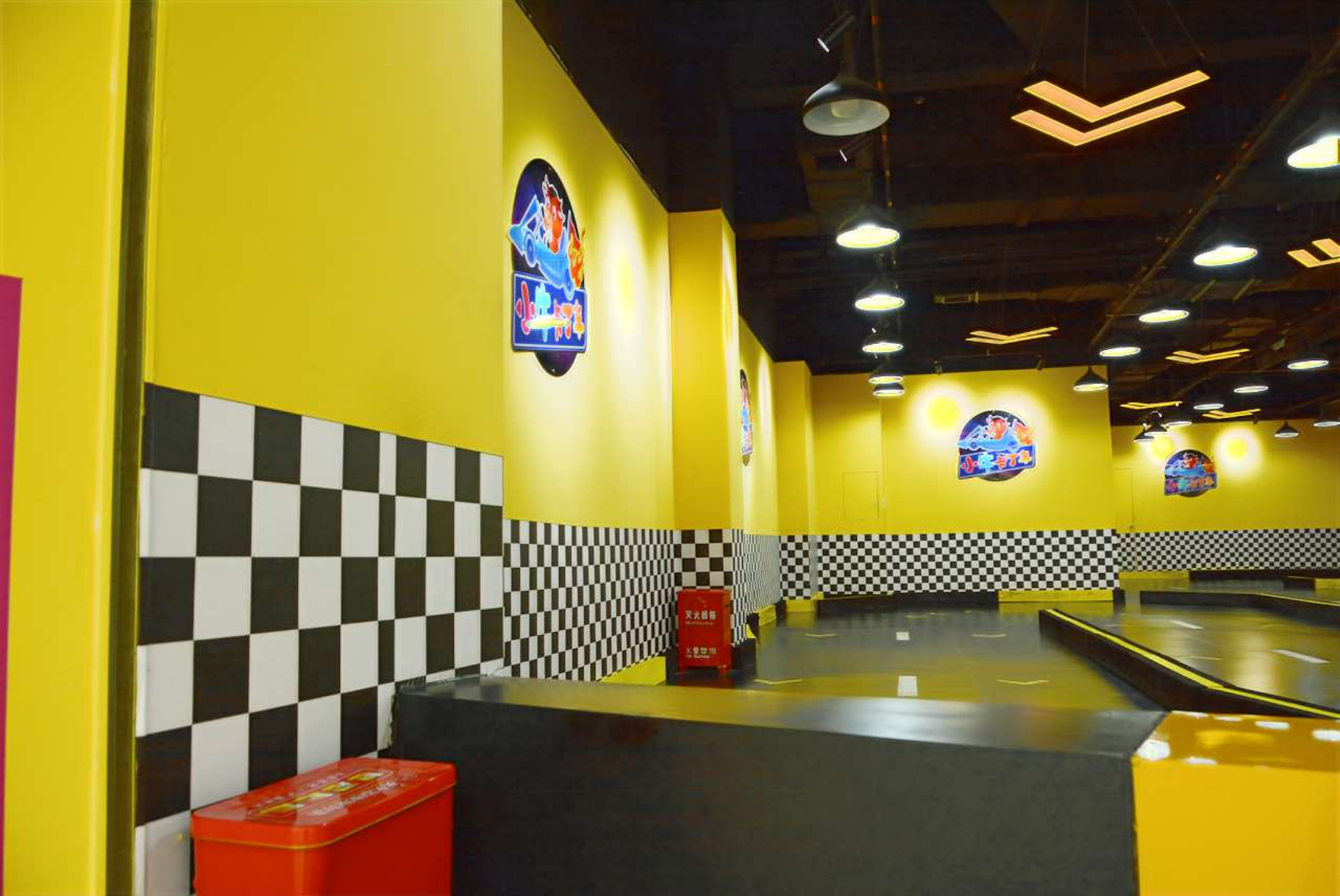}}
    \subfloat{\includegraphics[width = .10 \linewidth]{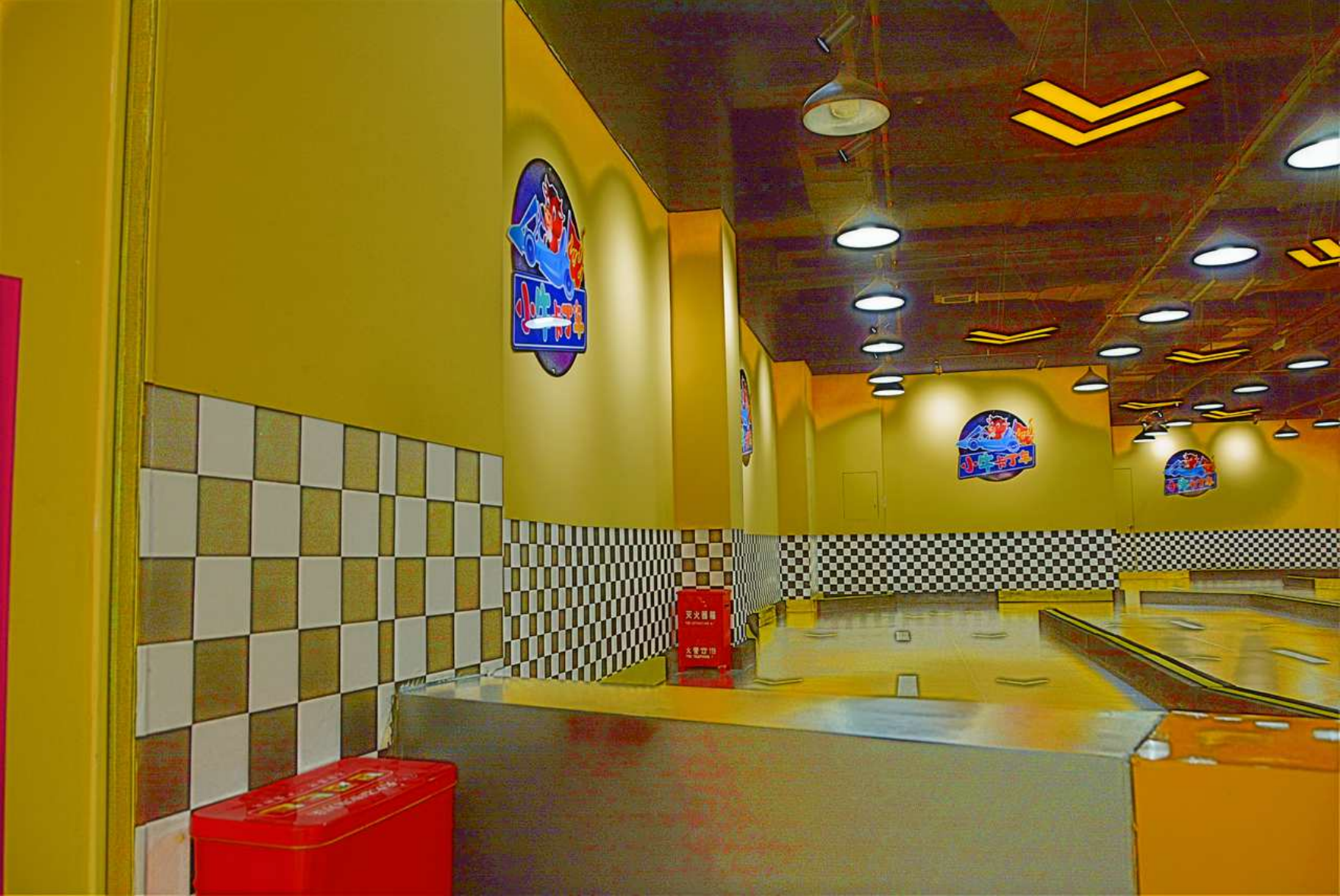}}
    \subfloat{\includegraphics[width = .10 \linewidth]{image/SNR-Net/00010.pdf}}
    \subfloat{\includegraphics[width = .10 \linewidth]{image/LIME/304_lime.pdf}}
    \subfloat{\includegraphics[width = .10 \linewidth]{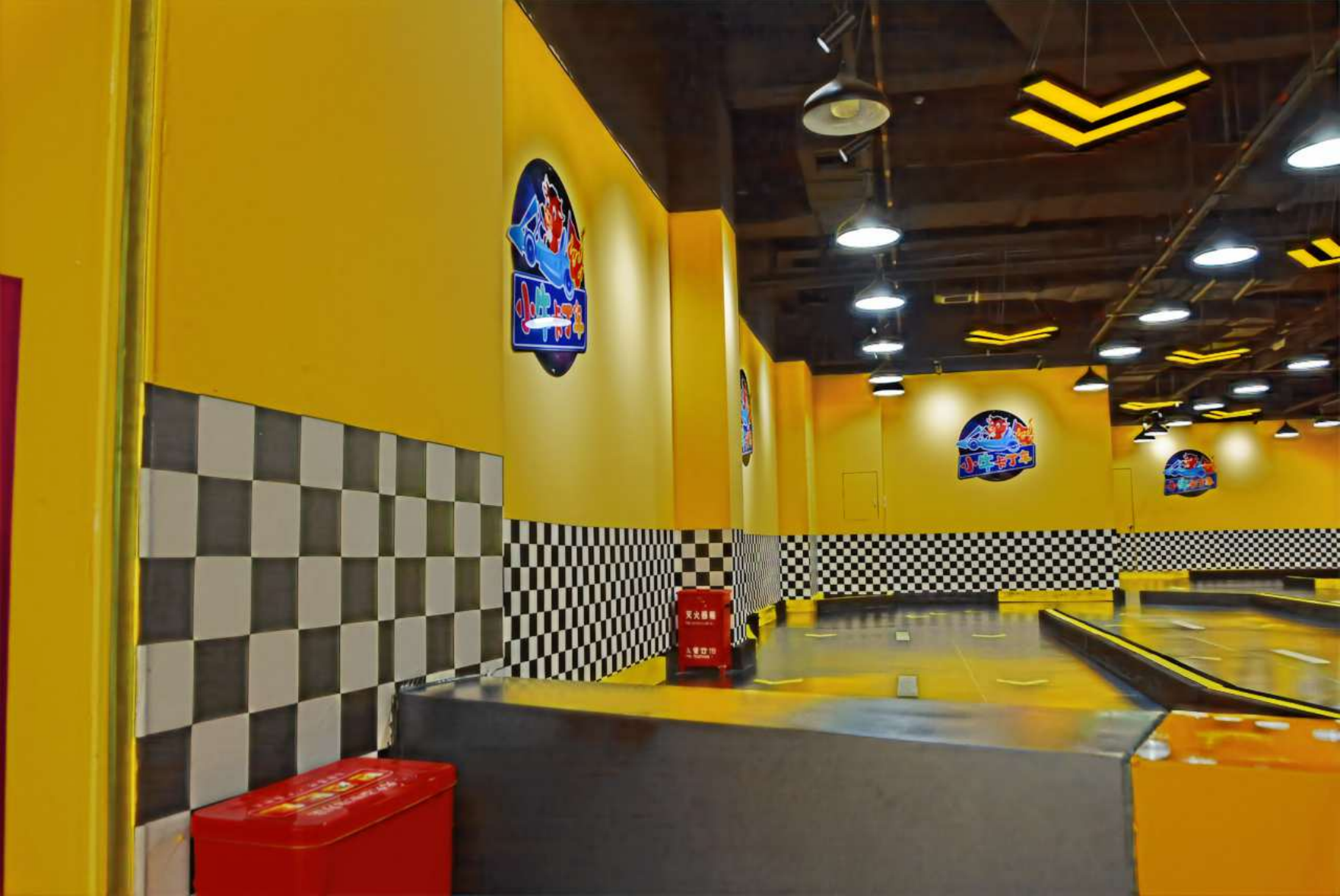}}
    \subfloat{\includegraphics[width = .10 \linewidth]{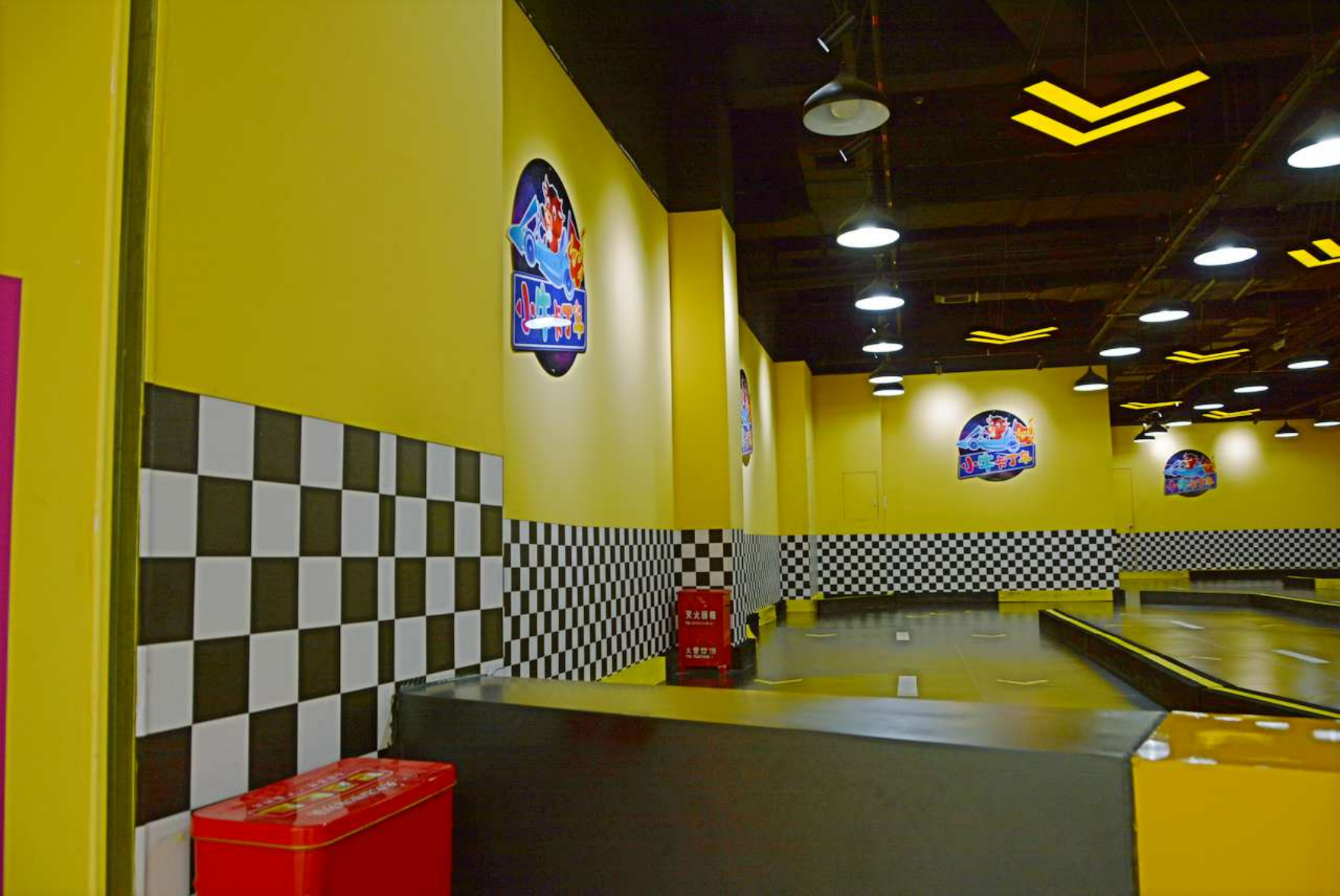}}\\\vspace{-0.1in}
    \subfloat{\includegraphics[width = .10 \linewidth]{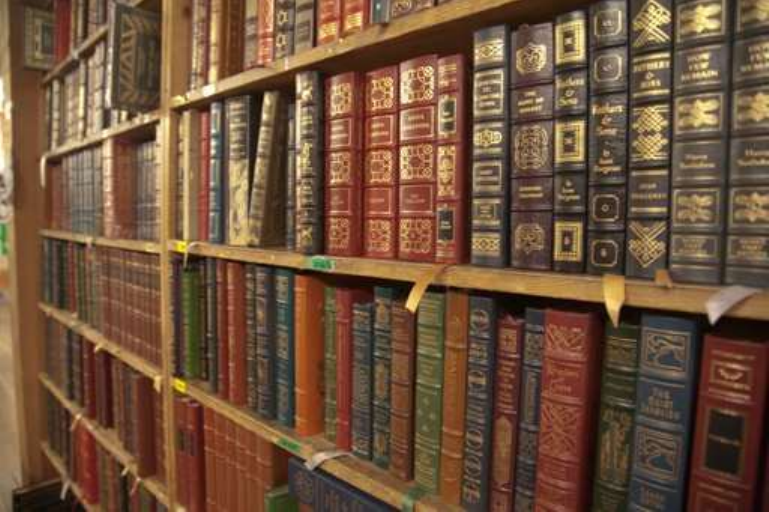}}
    \subfloat{\includegraphics[width = .10 \linewidth]{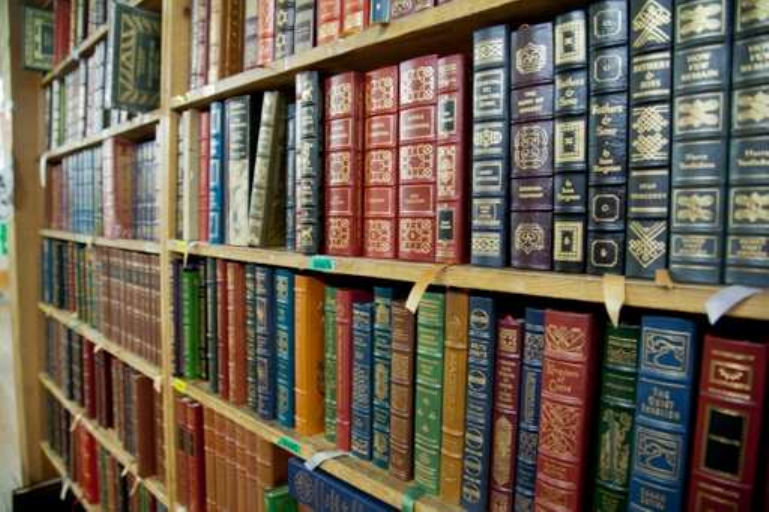}}
    \subfloat{\includegraphics[width = .10 \linewidth]{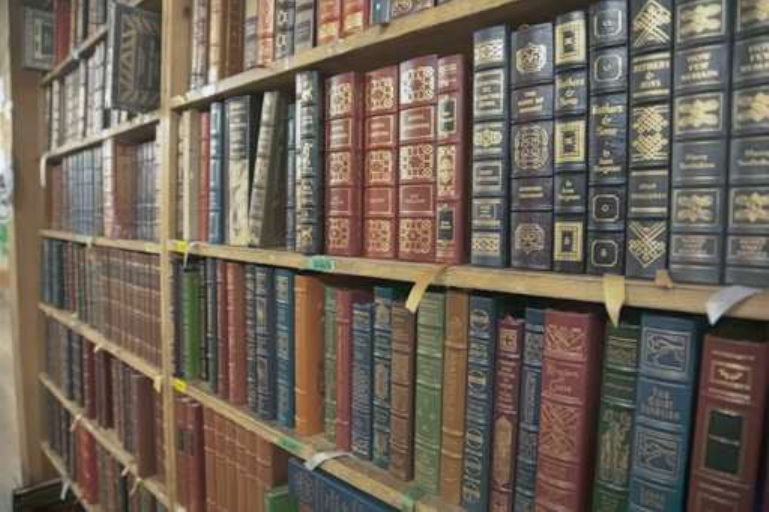}}
    \subfloat{\includegraphics[width = .10 \linewidth]{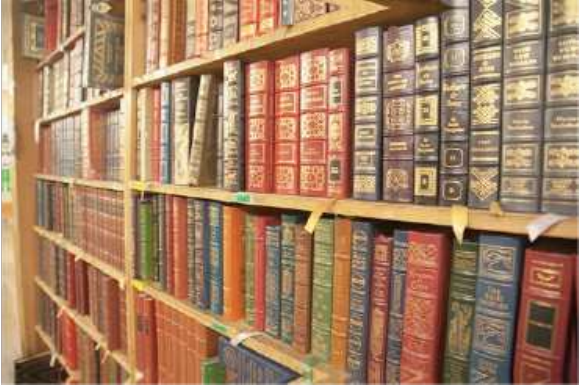}}
    \subfloat{\includegraphics[width = .10 \linewidth]{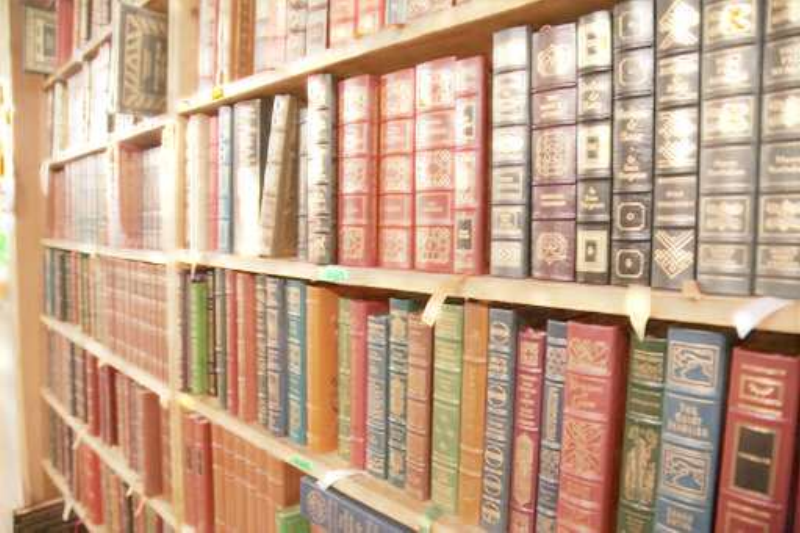}}
    \subfloat{\includegraphics[width = .10 \linewidth]{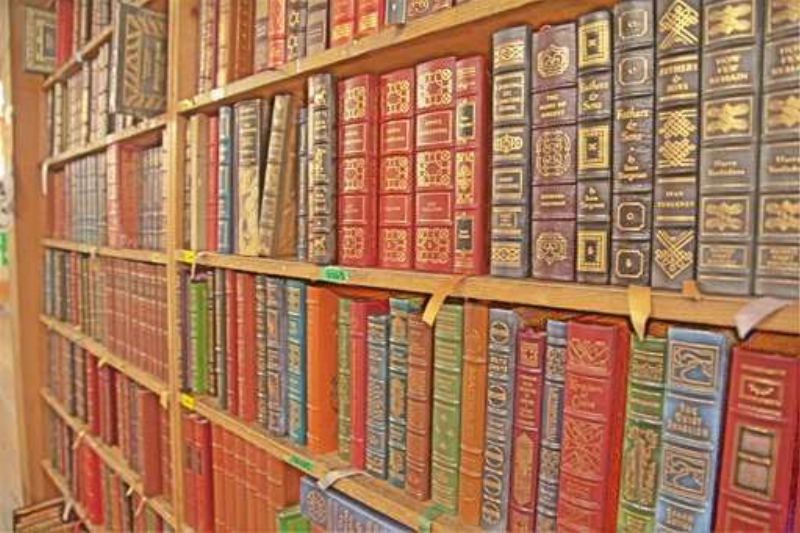}}
    \subfloat{\includegraphics[width = .10 \linewidth]{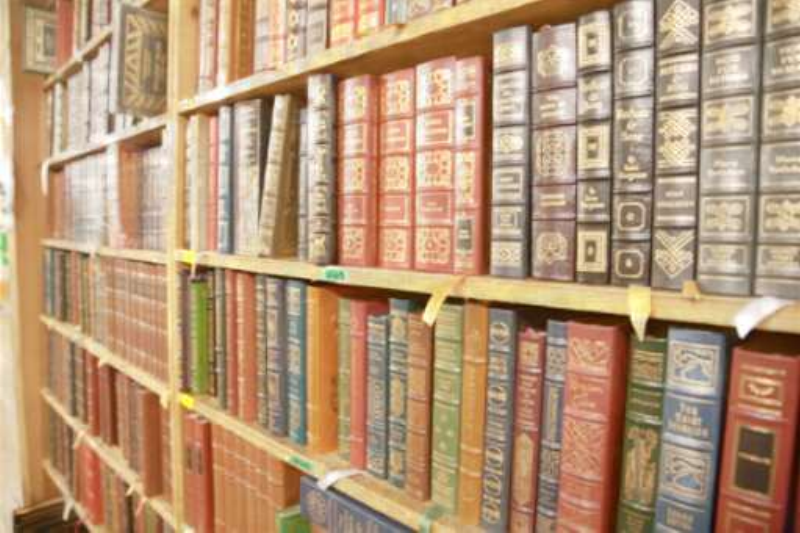}}
    \subfloat{\includegraphics[width = .10 \linewidth]{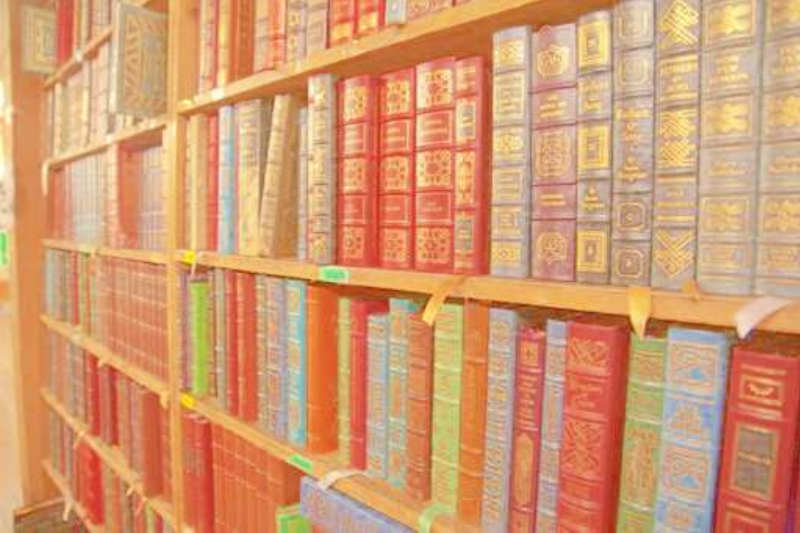}}
    \subfloat{\includegraphics[width = .10 \linewidth]{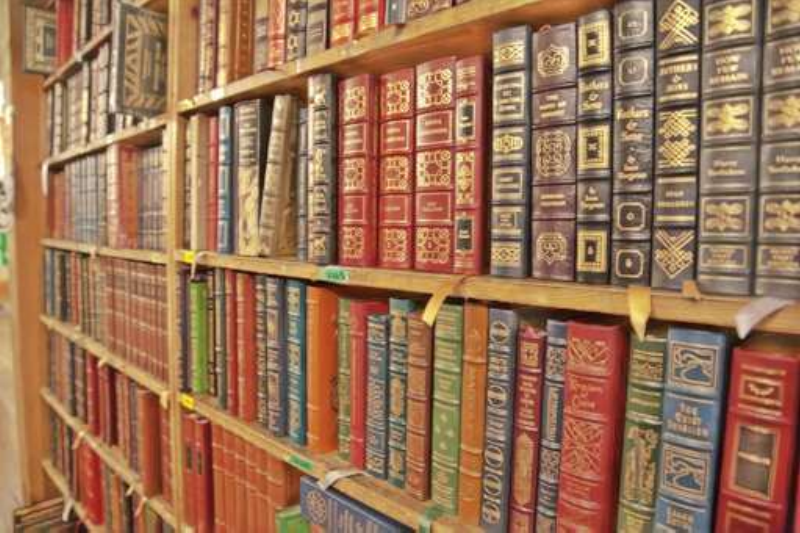}}
    \subfloat{\includegraphics[width = .10 \linewidth]{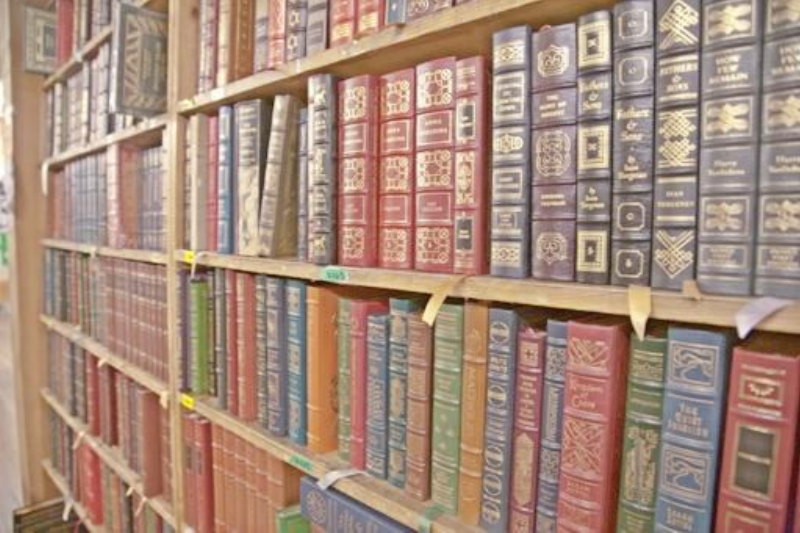}}\\\vspace{-0.16in}
    \subfloat{\includegraphics[width = .10 \linewidth]{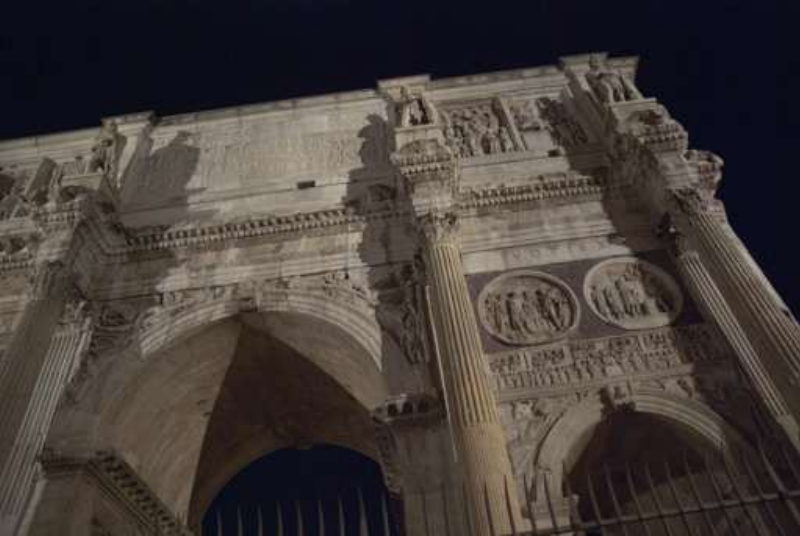}}
    \subfloat{\includegraphics[width = .10 \linewidth]{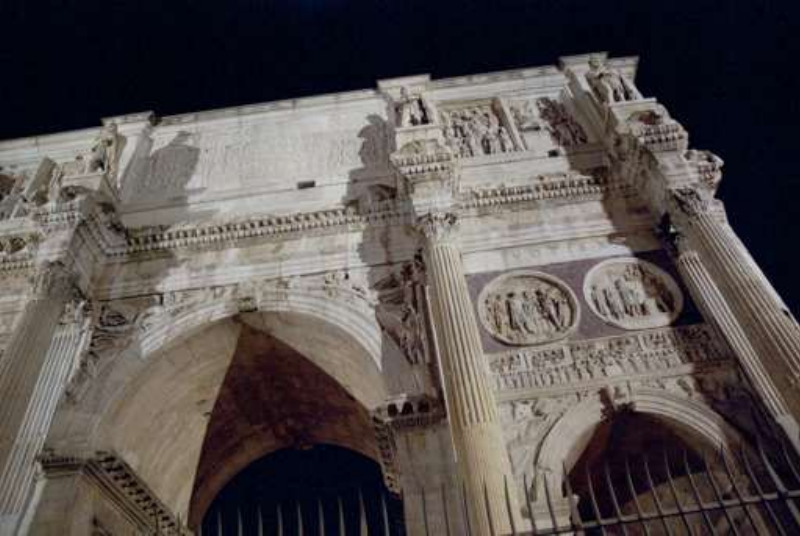}}
    \subfloat{\includegraphics[width = .10 \linewidth]{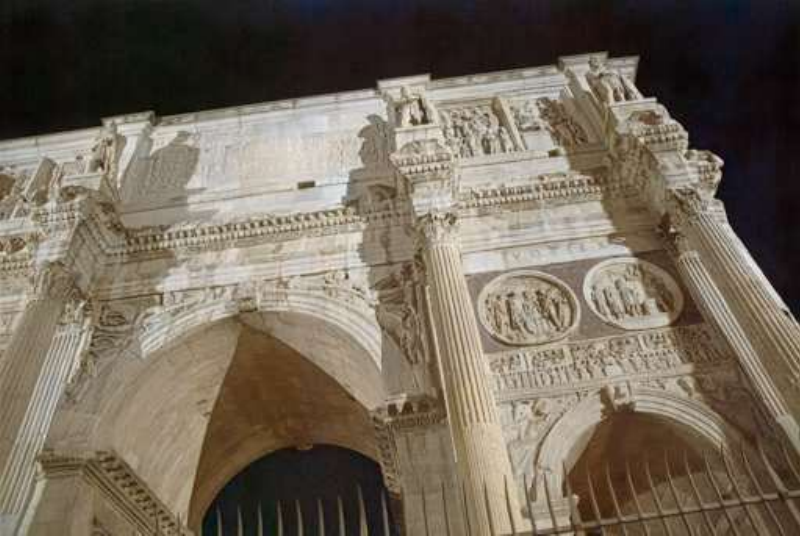}}
    \subfloat{\includegraphics[width = .10 \linewidth]{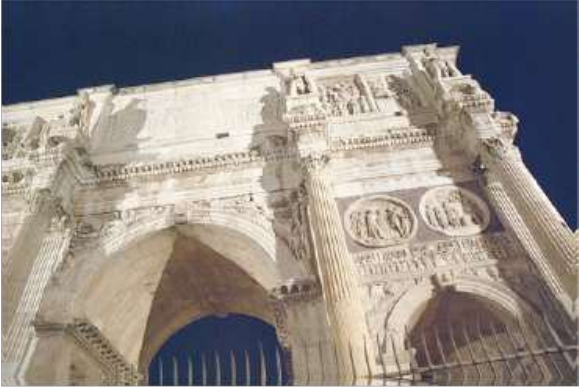}}
    \subfloat{\includegraphics[width = .10 \linewidth]{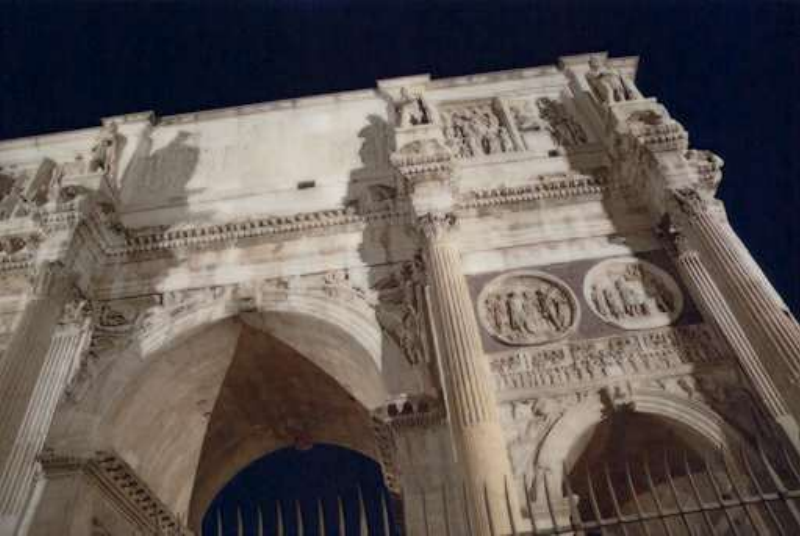}}
    \subfloat{\includegraphics[width = .10 \linewidth]{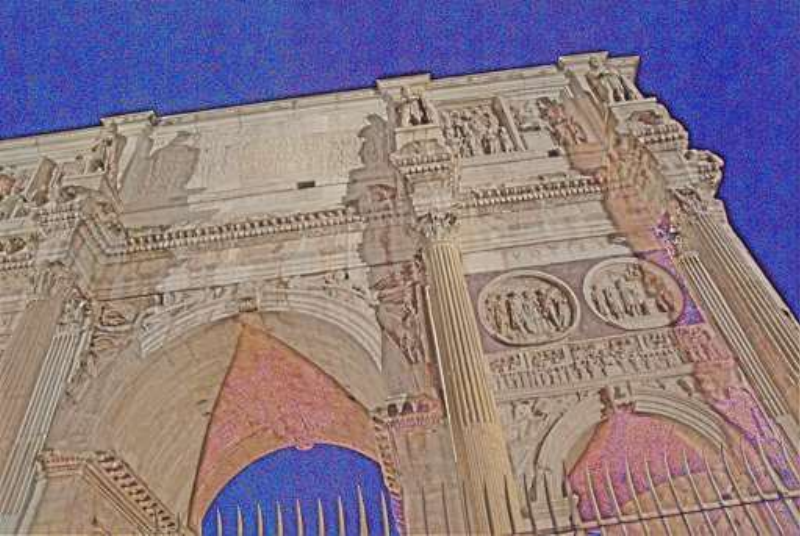}}
    \subfloat{\includegraphics[width = .10 \linewidth]{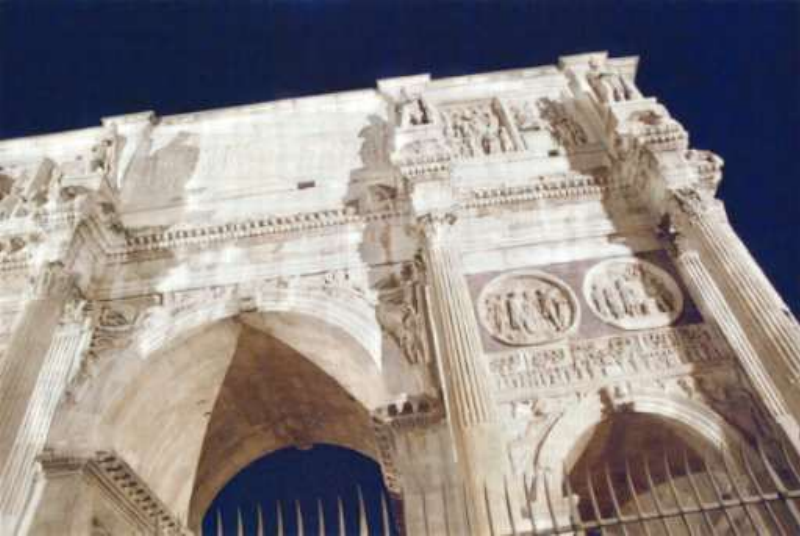}}
    \subfloat{\includegraphics[width = .10 \linewidth]{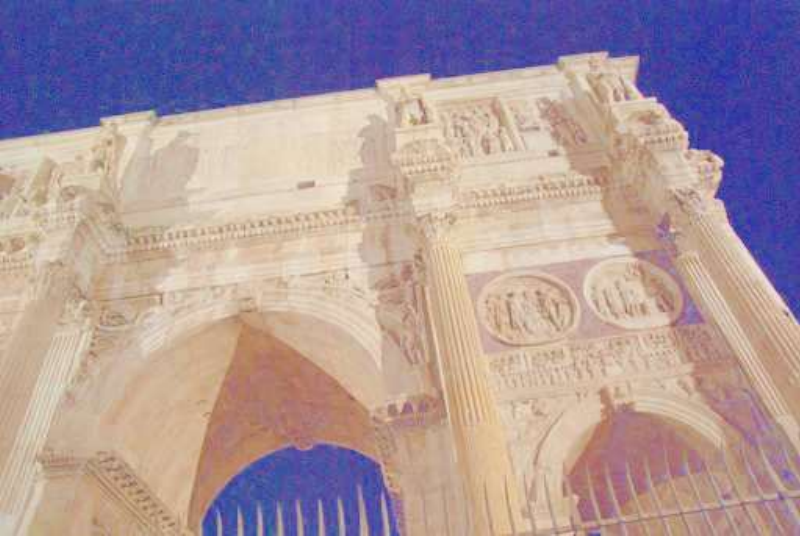}}
    \subfloat{\includegraphics[width = .10 \linewidth]{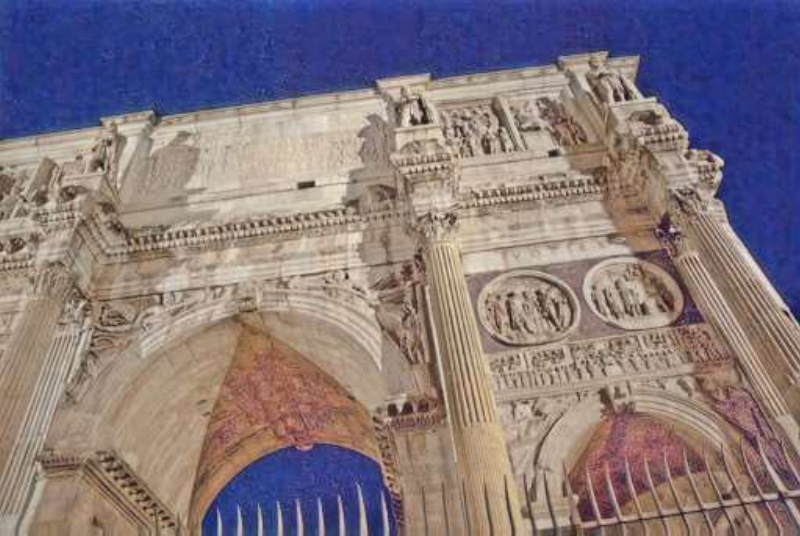}}
    \subfloat{\includegraphics[width = .10 \linewidth]{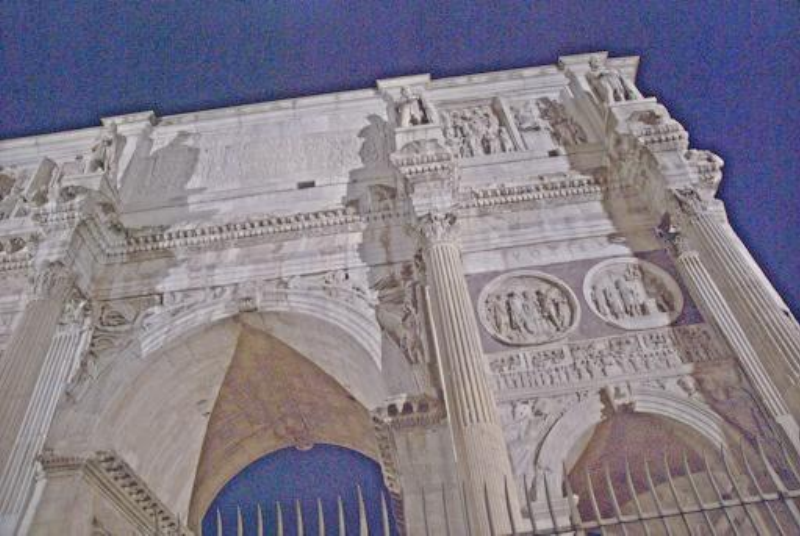}}\\\vspace{-0.16in}
    \setcounter{subfigure}{0}
    \subfloat[Input low light images]{\includegraphics[width = .10 \linewidth]{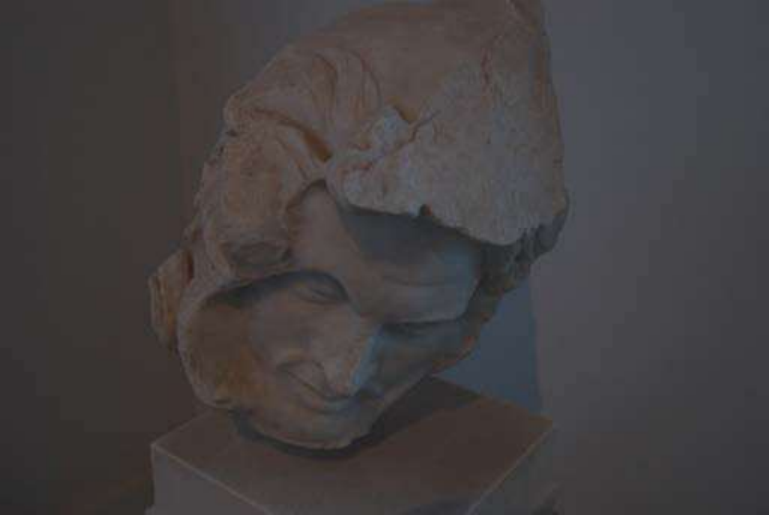}}
    \subfloat[Ground truth]{\includegraphics[width = .10 \linewidth]{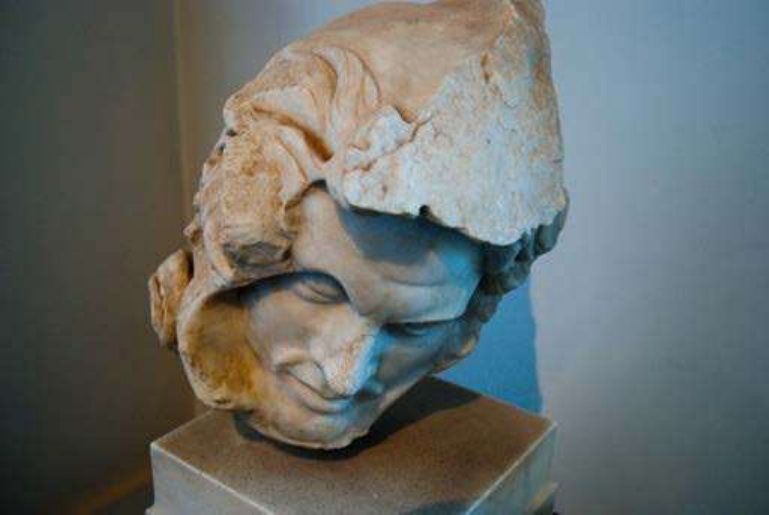}}
    \subfloat[WREN (proposed)]{\includegraphics[width = .10 \linewidth]{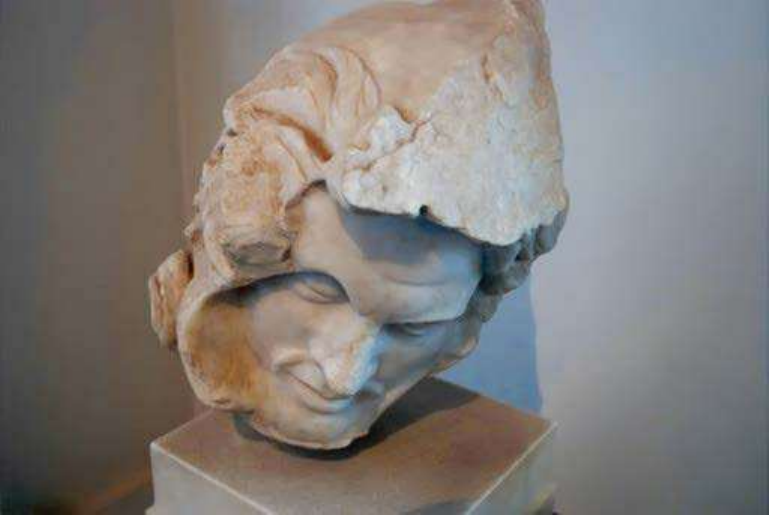}}
    \subfloat[ReDDiT]{\includegraphics[width = .10 \linewidth]{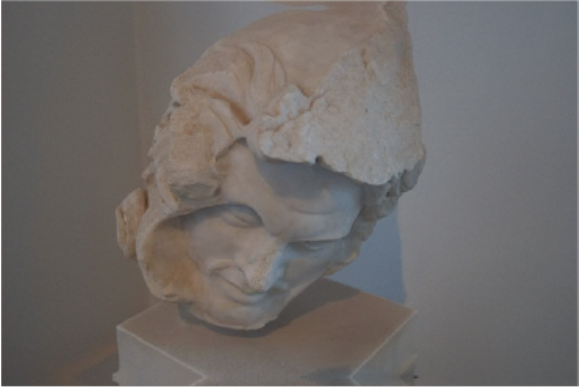}}
    \subfloat[Retinexformer]{\includegraphics[width = .10 \linewidth]{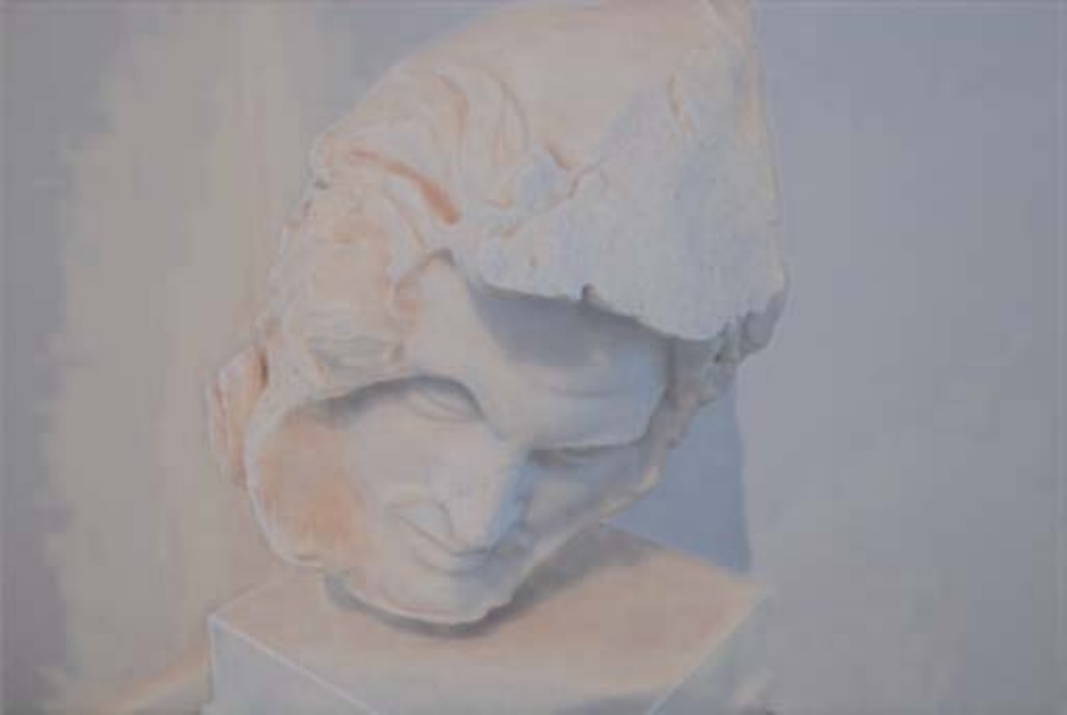}}
    \subfloat[RetinexNet]{\includegraphics[width = .10 \linewidth]{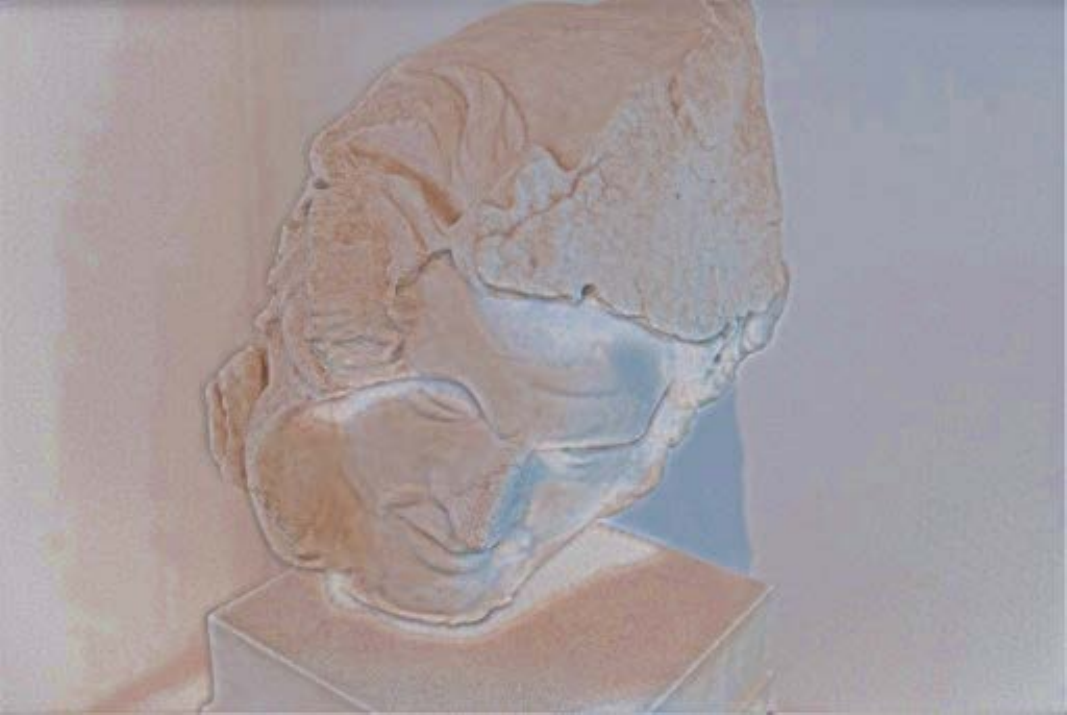}}
    \subfloat[SNR-Net]{\includegraphics[width = .10 \linewidth]{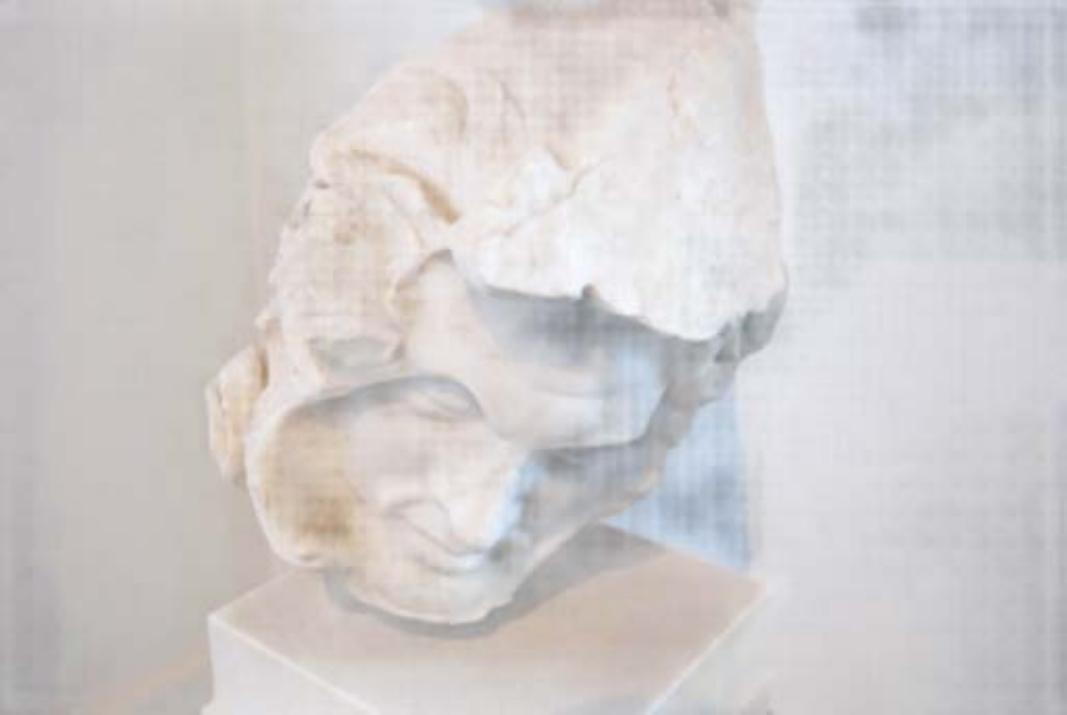}}
    \subfloat[LIME]{\includegraphics[width = .10 \linewidth]{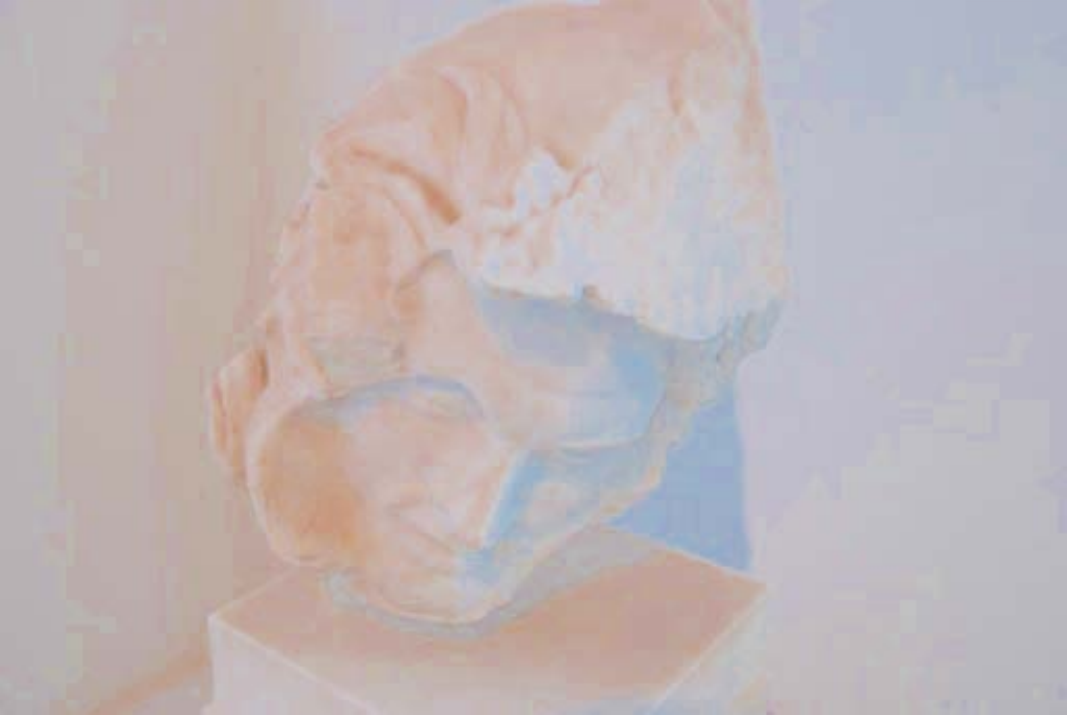}}
    \subfloat[KinD++]{\includegraphics[width = .10 \linewidth]{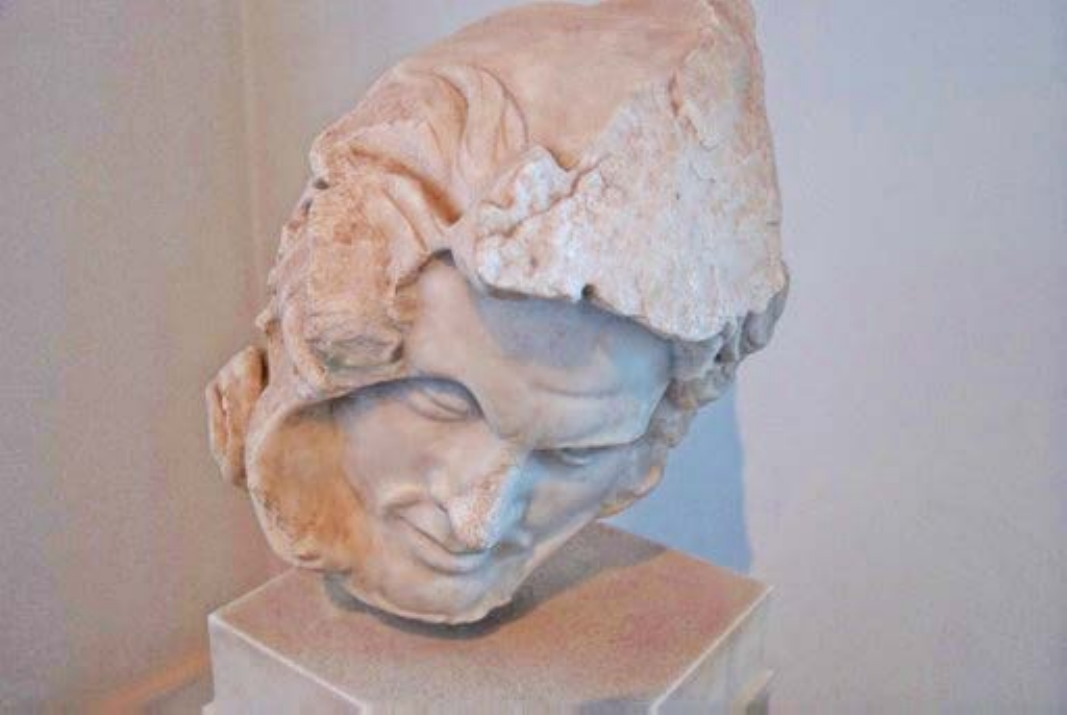}}
    \subfloat[Zero-DCE]{\includegraphics[width = .10 \linewidth]{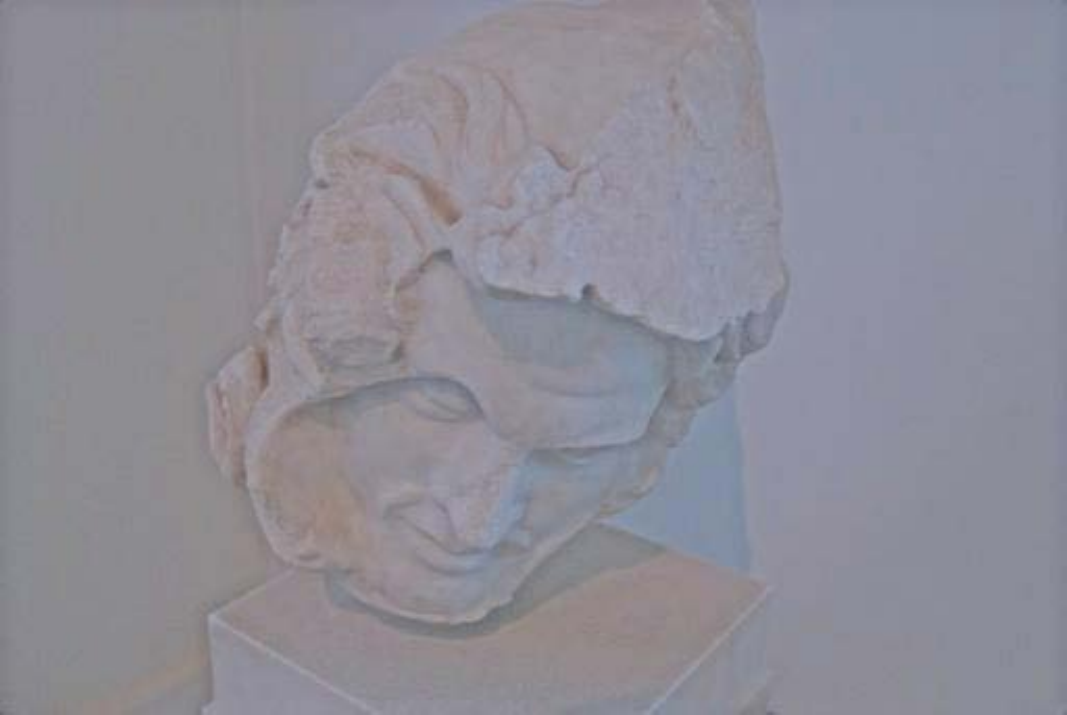}}\\
    \caption{Low light image enhancement results on several datasets. From top to bottom, every three-rows correspond to images in the LOLv1, SICE, LIEQ, NTIRE, and FiveK datasets.}
    \label{fig:LOL}
\end{figure*}

\section{Experiment}
\label{sec:experiment}
In this section, we perform LLIE experiments on multiple datasets and compare our method with existing deep learning and model-based methods.
\subsection{Setup}
We use five datasets: LOLv1 \cite{Chen2018Retinex}, SICE \cite{caiLearningDeepSingle2018a}, LIEQ \cite{zhaiPerceptualQualityAssessment2021}, NTIRE 2024 \cite{Liu2024NTIRE2024LLIE}, and MIT-Adobe FiveK \cite{bychkovskyLearningPhotographicGlobal2011}. 
These datasets contain pairs of low light and normal light images. Note that LOLv1, NTIRE 2024, and MIT-Adobe FiveK only provide training/validation splits, i.e., no official test set.

In the experiment, WREN is trained with LOLv1 (485 images), and evaluation is conducted on multiple test sets across the five datasets.
This demonstrates the robustness against the training datasets, especially for different dynamic ranges among datasets.

For SICE, we use Part~1 as the test set (360 images, excluding extremely large images). For LIEQ, we use all images as the test set (100 images). For the other datasets, we use their validation sets as test sets (LOLv1 has 15 images, NTIRE has 15 images, and FiveK has 500 images).

We compare our method with six baselines: ReDDiT \cite{lanEfficientDiffusionLow2025}, Retinexformer \cite{caiRetinexformerOnestageRetinexbased2023}, RetinexNet \cite{Chen2018Retinex}, SNR-Net \cite{xuSNRAwareLowlightImage2022}, KinD++ \cite{zhangBrighteningLowlightImages2021a}, Zero-DCE \cite{guoZeroReferenceDeepCurve2020a}, and LIME \cite{guoLIMELowLightImage2017c}.
Existing models are re-trained with the codes obtained by the original authors' repository, and evaluated on the same splits similar to WREN.
For WREN, we use the following set of hyperparameters: $\{\alpha,\beta,\gamma,\zeta,\omega,\delta,\eta\}=\{1.0,2.0,2.0,1.0,0.5,0.5,0.5\}$.
The training configurations are summarized in Table~\ref{tab:config}.

The performance is compared with four metrics: peak signal-to-noise ratio (PSNR), SSIM \cite{zhouwangImageQualityAssessment2004}, perceptual similarity (LPIPS) \cite{zhangUnreasonableEffectivenessDeep2018}, and mean absolute error (MAE).

\subsection{Results}
Table~\ref{tab:results} summarizes the numerical results where we calculate the mean of the metrics across all test data in the five datasets.
As shown in the table, WREN achieves the best performance on all metrics.
Even against the recent LLIE method ReDDiT, WREN has $>1$ dB PSNR and $>0.1$ LPIPS gains.
Some methods have very low ($< 0.5$) SSIMs: This implies that those methods are sensitive to the dynamic range changes where their numerical performance varies across to the datasets.

Figure~\ref{fig:LOL} presents enhanced images in the various datasets.
For the images only with dark regions (rows 1 to 9), WREN, ReDDiT, Retinexformer, and SNR-Net are comparable. RetinexNet, LIME, and KinD++ tend to amplify sensor noise because their illumination enhancements also affect noise.
This effect is especially visible in dark regions.
They also yield annoying color changes due to the enhancement of the normal light images.

For the images with a mixture of various lighting regions (rows 10 to 15), the enhancement performances significantly differ while WREN obtains almost satisfactorily results.
It is especially visible in the row 10.
RetinexFormer over-enhances the sky region, resulting in a color shift.
Images obtained by ReDDiT, RetinexNet, SNR-Net, LIME, KinD++, and Zero-DCE are overexposed.
These visualized results indicate that existing methods over-enhance low light images (at least, partly) since they are mainly fit to images only with dark regions. In contrast, WREN successfully enhances low light regions while preserving normal light regions.

\section{Conclusion}
\label{sec:con}
In this paper, we propose WREN, a retinex-guided LLIE method with double U-Net-like sub-networks. Our method consists of two modules: 1) retinex decomposition module that decomposes a low light image into reflectance and illumination maps through one shared encoder with two disjoint decoders, and 2) illumination compensation module that enhances the illumination map by a customized U-Net.
In the experiments, WREN achieves the state-of-the-art performance both in numerical comparisons and visual qualities of enhanced images for multiple datasets.

\bibliographystyle{IEEEtran}
\bibliography{ICASSP_LLIE}

\end{document}